\documentclass{article}
\usepackage{frExamplee}
\usepackage{booktabs}       
\usepackage{amsfonts}       
\usepackage{amsmath}
\usepackage{amssymb}
\usepackage{graphicx}
\usepackage{csquotes}
\usepackage[backend=biber, style=apa,]{biblatex}
\usepackage{setspace}
\usepackage[usenames, dvipsnames]{xcolor}
\usepackage{xspace}
\usepackage{caption}
\usepackage{subcaption}
\usepackage{multirow}
\usepackage{float}
\usepackage{wrapfig}
\usepackage{placeins}
\usepackage{algpseudocode}
\usepackage{xcolor}
\usepackage{algorithm}
\usepackage{algorithmicx}
\usepackage{hyperref}
\usepackage{makecell}

\definecolor{darkyellow}{rgb}{0.85, 0.65, 0.0}

\usepackage{fancyhdr} 
\title{Field Testing of a Stochastic Planner for ASV Navigation Using Satellite Images}

\author{
Philip (Yizhou) Huang\thanks{Corresponding author, worked done during Master's degree at the University of Toronto.} \\
Robotics Institute\\
Carnegie Mellon University\\
Pittsburgh, USA \\
\texttt{philiphuang@cmu.edu} \\
\And
Tony (Qiao) Wang \\
Division of Engineering Science \\
University of Toronto\\
Toronto, Canada \\
\texttt{tonyivt.wang@mail.utoronto.ca} \\
\AND
Florian Shkurti \\
Department of Computer Science\\
University of Toronto\\
Toronto, Canada \\
\texttt{florian@cs.toronto.edu} \\
\And
Timothy D. Barfoot \\
Institute for Aerospace Studies\\
University of Toronto\\
Toronto, Canada \\
\texttt{tim.barfoot@utoronto.ca}
}

\begin{document}

\maketitle

\begin{abstract}
We introduce a multi-sensor navigation system for autonomous surface vessels (ASV) intended for water-quality monitoring in freshwater lakes. Our mission planner uses satellite imagery as a prior map, formulating offline a mission-level policy for global navigation of the ASV and enabling autonomous online execution via local perception and local planning modules.  A significant challenge is posed by the inconsistencies in traversability estimation between satellite images and real lakes, due to environmental effects such as wind, aquatic vegetation, shallow waters, and fluctuating water levels. Hence, we specifically modelled these traversability uncertainties as stochastic edges in a graph and optimized for a mission-level policy that minimizes the expected total travel distance. To execute the policy, we propose a modern local planner architecture that processes sensor inputs and plans paths to execute the high-level policy under uncertain traversability conditions. Our system was tested on three km-scale missions on a Northern Ontario lake, demonstrating that our GPS-, vision-, and sonar-enabled ASV system can effectively execute the mission-level policy and disambiguate the traversability of stochastic edges. Finally, we provide insights gained from practical field experience and offer several future directions to enhance the overall reliability of ASV navigation systems.
 
\end{abstract}

\section{Introduction}

\begin{figure}[t!]
    \centering
    \subfloat[Strong wind]{
        \includegraphics[width=0.45\columnwidth]{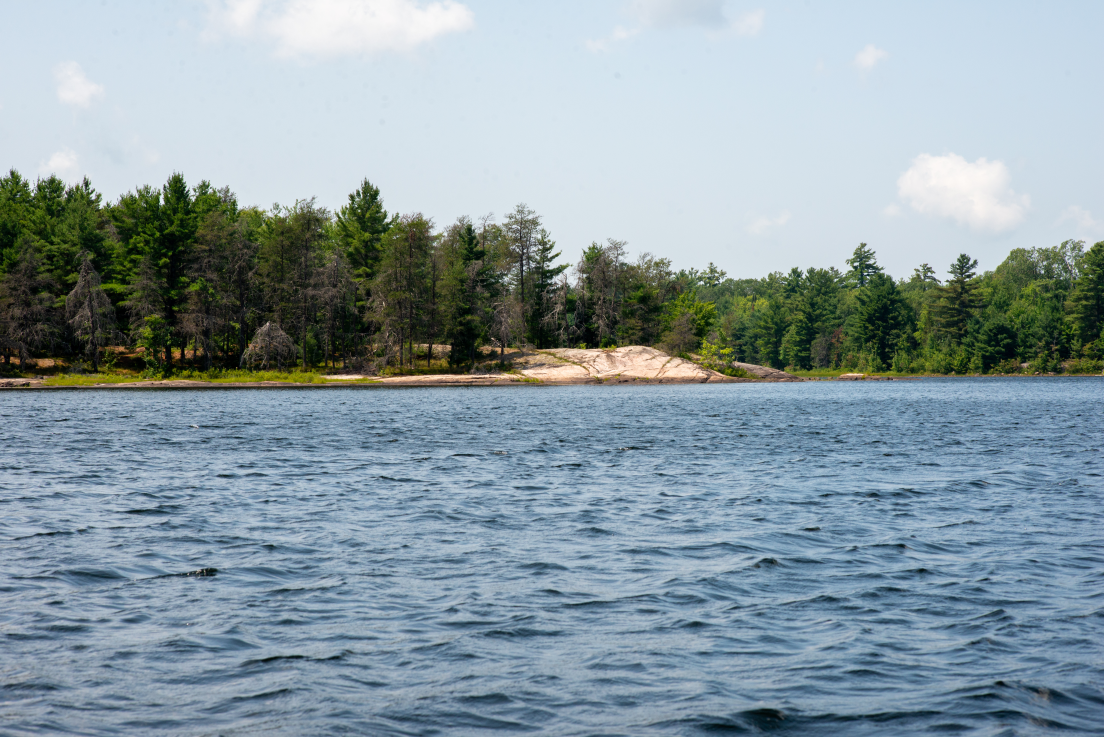}
        \label{fig:windy_water}
    }
    \hfill
    \hfill
    \subfloat[Protruding logs]{
        \includegraphics[width=0.45\columnwidth]{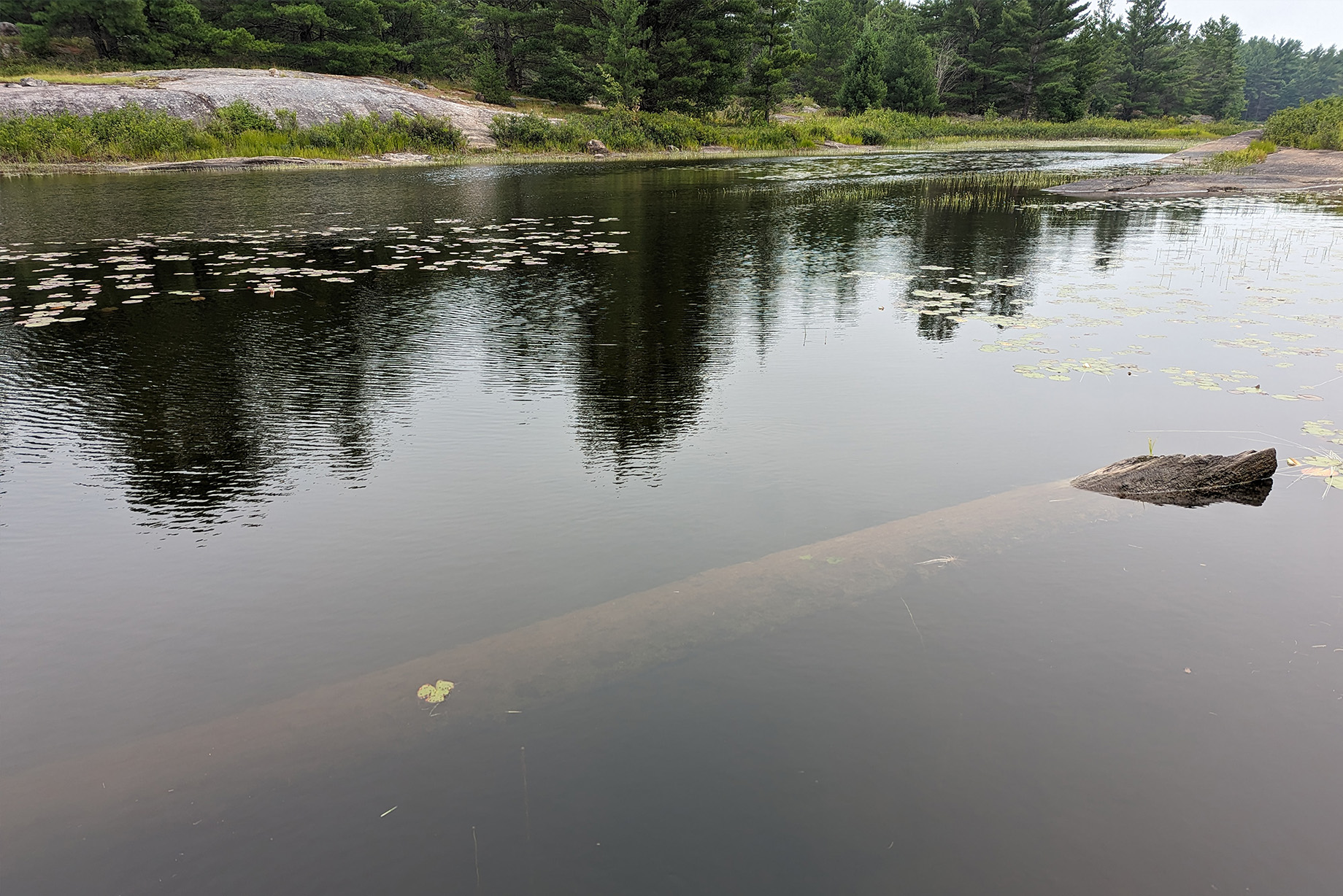}
        \label{fig:extruding_logs}
    }
    \hfill
    \subfloat[Shallow water]{
        \includegraphics[width=0.45\columnwidth]{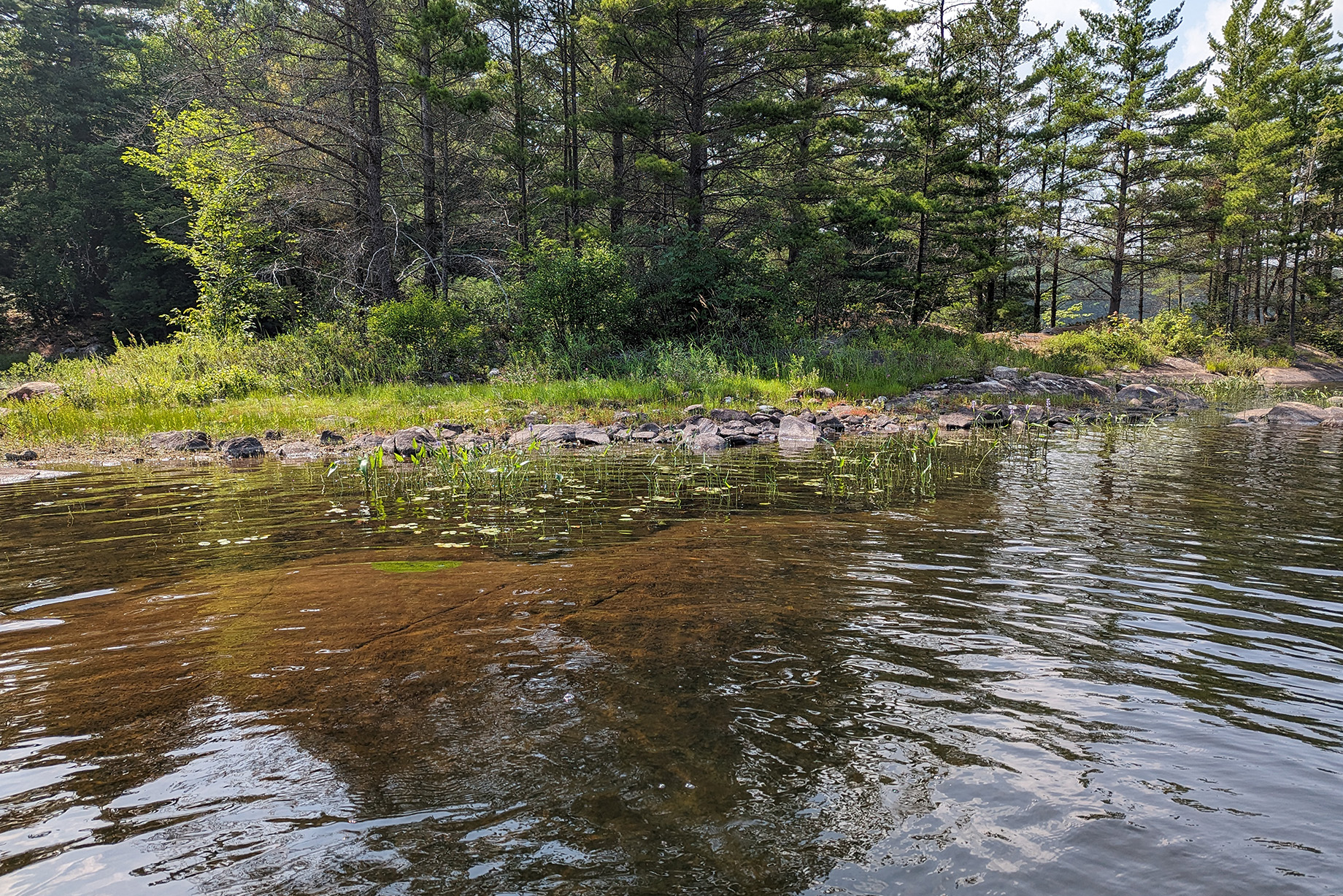}
        \label{fig:shallow_water_and_aquatic_plants}
    }
    \hfill
        \subfloat[Aquatic plants]{
        \includegraphics[width=0.45\columnwidth]{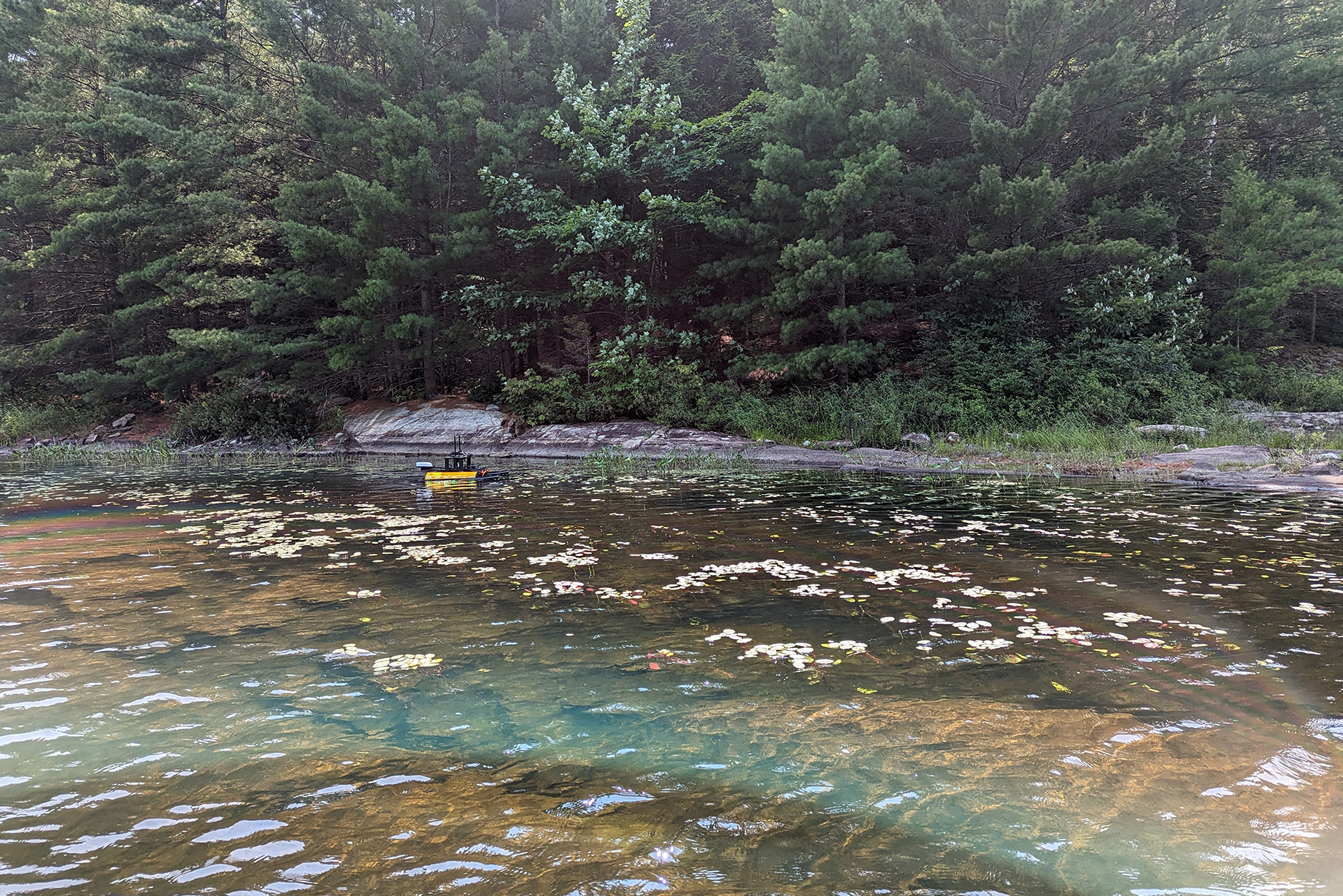}
        \label{fig:lilypads}
    }
    \caption{Real-world challenges that motivate the use of stochastic edges in our planning setup.}
    \label{fig:obstacles}
\end{figure}

Autonomous Surface Vessels (ASVs) have seen increasing attention as a technology to monitor rivers, lakes, coasts, and oceans in recent years \autocite{robot_env2012, Odetti2020-iv, Ferri2015-sb, Madeo2020-yq, Cao2020-fc, Dash2021-uf, Ang2022-ep, MahmoudZadeh2022-su}. A fundamental challenge to the wide adoption of ASVs is the ability to navigate safely and autonomously in uncertain environments, especially for long durations. 
For example, many existing ASV systems require the user to precompute a waypoint sequence. The robot then visits these target locations on a map and attempts to execute the path online \autocite{tang2020practical, Vasilj2017DesignDA}.
However, disturbances such as strong winds, waves, unseen obstacles, aquatic plants that may or may not be traversable, and even simply changing visual appearances in a water environment are challenging for ASV navigation (Fig. \ref{fig:obstacles}). 
Many potential failures in robot perception and control systems may also undermine the mission's overall success. 
Engineering challenges such as power and computational budget also make it challenging to implement many robot autonomy modules and integrate them with onboard sensors.
To ensure the safety of the overall operation, users of the ASV system may also wish to understand its high-level behaviour and any decisions made during the mission.

Our long-term goal is to use an ASV to monitor lake environments and collect water samples for scientists. A requirement for achieving this, and the primary focus of this paper, is to ensure robust global and safe local navigation. To enhance the robustness of the overall system, we identify waterways that are prone to local blockage as stochastic edges and plan 
mission-level policies offline on our high-level map.
Uncertainties that arise during policy execution are handled by the local planner.
One planning framework that is suitable for modelling uncertain paths is the Canadian Traveller Problem (CTP) \autocite{ctp1991}, a variant of the shortest-path planning problem for an uncertain road network. 
The most significant feature in a CTP graph is the stochastic edge, which has a probability of being blocked.
The state of any stochastic edge can be disambiguated by visiting the edge. 
Once the state has been visited and classified as traversable or not, it remains the same.
Separating planning into a high-level mission planner and a local online planner offers several advantages. The high-level planner creates a global policy with contingencies that can be adjusted online for any unmapped obstacles. Offline planning improves interpretability and reduces the computational resources required online. This allows the user to easily inspect the planned paths before deploying the robot. With offline planning, more time can be allocated to find the optimal global paths. The local planner can then focus on accurately tracking the global path and adjusting for any unmapped obstacles and environmental uncertainties.

In our prior work \autocite{boat2022}, we proposed a navigation framework --- the Partial Covering Canadian Traveller Problem (PCCTP) --- to solve a mission-planning problem in an uncertain environment.
The framework used a stochastic graph derived from coarse satellite images to plan an adaptive policy that visits all reachable target locations. 
Stochasticity in the graph represents possible events where a water passage between two points is blocked due to changing water levels, strong wind, and other unmapped obstacles.
The optimal policy is computed offline with a best-first tree-search algorithm.
We evaluated our solution method on 1052 Canadian lakes selected from the \textit{CanVec Series} Ontario dataset \autocite{canvec} and showed it can reduce the total distance to visit all targets and return. In our past field tests, we found that completing the global mission fully autonomously, even for a 5-node policy, was very challenging. A total of seven manual interventions were required for reasons other than battery replacement in the two old field trials conducted. The failure to detect unmapped local obstacles directly led to collisions. We observed that the previous local planner experienced edge cases where it could not find a valid path around local obstacles while tracking the global plan. In addition, the local navigation system had many intermittent errors that temporarily stopped the robot due to false positives from obstacle detection. These past field experiences highlight the need to improve the previous system and conduct more field tests in new environments.

This article extends our previous work as described by \textcite{boat2022} in two ways.  
First, we made significant improvements to our local planner responsible for tracking the global path and handling any locally occurring uncertainties such as obstacles. 
Our ASV system estimates the waterline using a learned network and a stereo camera and detects underwater obstacles using a mechanically scanning sonar. 
We fuse both sensors into an occupancy grid map, facilitating a sampling-based local motion planner to compute a pathway to track the global path while avoiding local obstacles. 
As in our previous research, we use a timer to distinguish stochastic edges and select appropriate policy branches based on the traversability assessment of the stochastic edges.
Secondly, we have validated the overall system on three distinct missions, two of which are new. Our field trials show that our ASV reliably and autonomously executes precomputed policies from the mission planner under varying operating conditions and amid unmapped obstacles, even when the local planner does not perfectly map the local environment or optimally steer the ASV. 
We have also tested the local planner through an ablation study to identify bottlenecks in localization, mapping, and sensor fusion in the field.
Our lessons learned from our field tests are detailed, and we believe this work will serve as a beneficial reference for any future ASV systems developed for environmental monitoring.

\begin{figure}
\centering
\includegraphics[width=\textwidth]{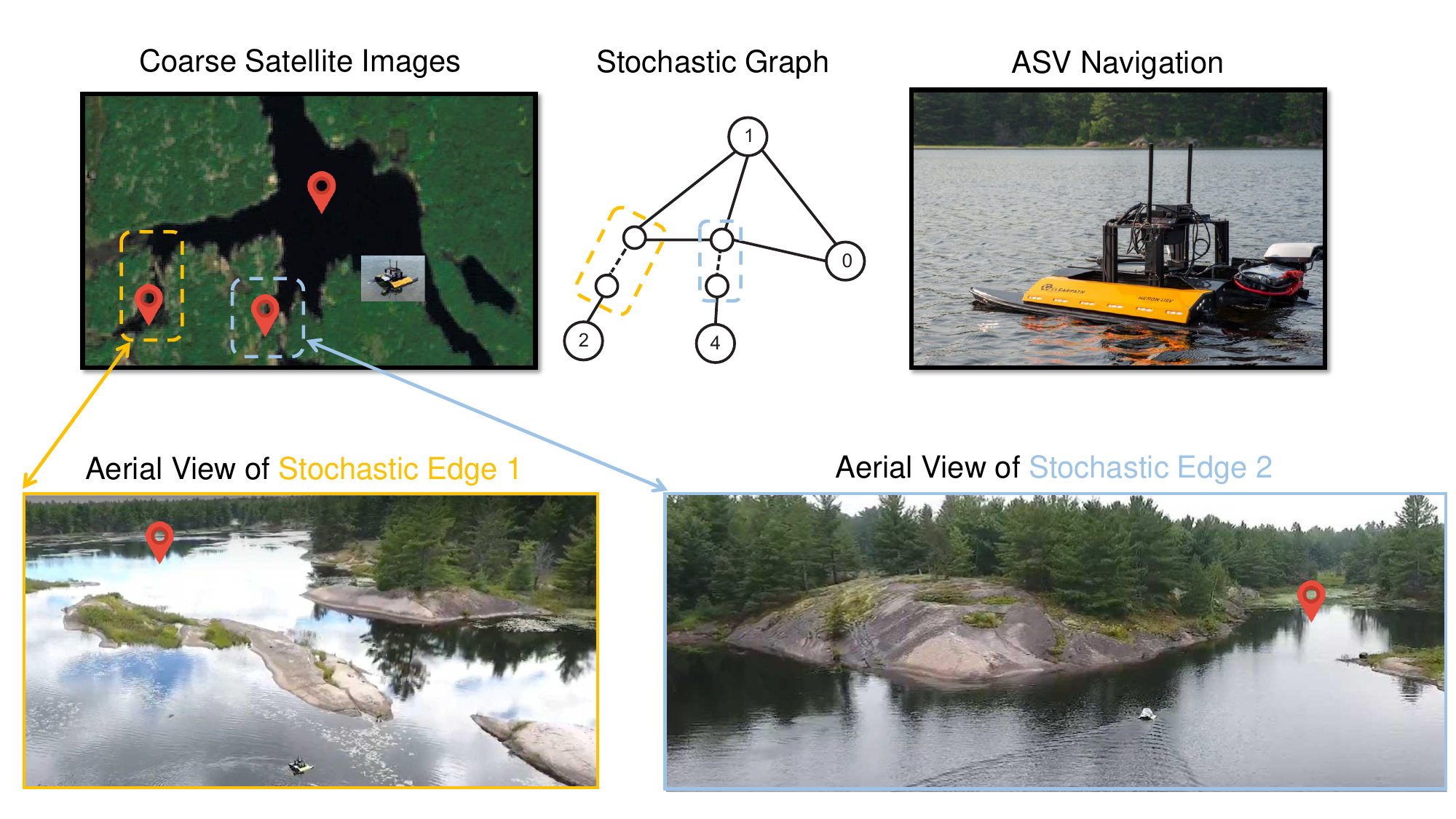}
\caption{A high-level overview of our navigation framework for water sampling. Given a set of user-selected target locations (\textcolor{red}{red} icons), our algorithm identifies stochastic edges from coarse satellite images and plans a mission-level policy for ASV navigation. Aerial views of two stochastic edges from real-world experiments are shown here.}
\label{fig:teaser}
\end{figure}

\section{Related Works}
Autonomous ASV navigation for environmental monitoring requires domain knowledge from multiple fields, such as perception, planning, and overall systems engineering. In this section, we present a brief survey of all these related fields and discuss the relationship to our methods and any remaining challenges. 

\textbf{Satellite Imagery Mapping} First, mission planning in robotics often requires a global, high-level map of the operating environment. Remote sensing is a popular technique to build maps and monitor changes in water bodies around the world because of its efficiency \autocite{WaterRemoteSensing, Yang2017-vb}.
The \textit{JRC} Global Surface Water dataset \autocite{JRCwater} maps changes in water coverage from 1984 to 2015 at a 30 m by 30 m resolution, produced using \textit{Landsat} satellite imagery. 
Since water has a lower reflectance in the infrared channel, an effective method is to calculate water indices, such as Normalized Difference Water Index (NDWI) \autocite{ndwi} or MNDWI \autocite{mndwi}, from two or more optical bands (e.g., green and near-infrared).
However, extracting water data using a threshold in water indices can be nontrivial due to variations introduced by clouds, seasonal changes, and sensor-related issues. 
To address this, \textcite{Li2012-bn} and \textcite{Feyisa2014-io} have developed techniques to select water-extraction thresholds adaptively.
Our approach aggregates water indices from historical satellite images to estimate probabilities of water coverage (see Sec. \ref{sec:graph_estimation}).
Overall, we argue that it is beneficial to build stochastic models of surface water bodies due to their dynamic nature and imperfect knowledge derived from satellite images. 

\textbf{Global Mission Planning} The other significant pillar of building an ASV navigation system is mission planning. First formulated in the 1930s, the Travelling Salesman Problem (TSP) \autocite{TSP_survey} studies how to find the shortest path in a graph that visits every node once and returns to the starting node. 
Modern TSP solvers such as the \textit{Google} OR-tools \autocite{ortools} can produce high-quality approximate solutions for graphs with about 20 nodes in a fraction of a second. 
Other variants have also been studied in the optimization community, such as the Travelling Repairman Problem \autocite{mlp_1986} that minimizes the total amount of time each node waits before the repairman arrives, and the Vehicle Routing Problem \autocite{Toth2002-zq} for multiple vehicles. 
In many cases, the problem graphs are built from real-world road networks, and the edges are assumed to be always traversable. 
In CTP \autocite{ctp1991}, however, edges can be blocked with some probability.
The goal is to compute a policy that has the shortest expected path to travel from a start node to a single goal node. 
CTP can also be formulated as a Markov Decision Process \autocite{bellman1957markovian} and solved optimally with dynamic programming \autocite{Polychronopoulos_undated-rw} or heuristic search \autocite{ctpao*}.
The robotics community has also studied ways in which the CTP framework can be best used in path planning \autocite{Ferguson2004-cf, ctp-guo2019}. 
Our problem setting, the Partial Covering Canadian Traveller Problem (PCCTP), lies at the intersection of TSP and CTP, where the goal is to visit a partial set of nodes on a graph with stochastic edges. 
A similar formulation, known as the Covering Canadian Traveller Problem (CCTP) \autocite{cctp}, presents a heuristic, online algorithm named Cyclic Routing (CR) to visit every node in a complete $n$-node graph with at most $n-2$ stochastic edges.
A key distinction between CCTP and our setting is that CCTP assumes all nodes are reachable, whereas, in PCCTP the robot may give up on unreachable nodes located behind an untraversable edge.

In our work, the user specifies the target locations of the planner, and the ASV can visit these predefined locations and collect water samples for off-site analysis. Modern ASVs can also be equipped with scientific instruments, such as the YSI Sonde used in \autocite{Jeong2020-xs}, for in situ analysis of a spatial area. In this case, robotics path planning can also consider scientific values in addition to efficiency, traversability, and time constraints. Complete coverage planning first determines all areas that can be traversed and then selects either a set motion primitive \autocite{Garneau2013-yc} or a lawnmower pattern \autocite{Karapetyan2018-ic} to encompass a surveying region. The survey region can also be decomposed into many distinct, non-overlapping cells, and the visitation order can be optimized with the TSP algorithm \autocite{Choset2001-nk}. As an alternative, informative path planning adaptively plans the robot's path and goal based on real-time sensor data to maximize scientific value \autocite{Bai2021-su}. A common approach builds a probabilistic model of the environment online with Bayesian models \autocite{Marchant2012-di, Chen-RSS-22} and then identifies the next target location that maximizes the information gain \autocite{Bai2016-uz, Marchant2014-pi}. Informative path planning can also be performed in continuous space with a sampling-based planner, which finds the best path in the sampled tree or roadmap that maximizes information gain subject to a budget constraint \autocite{Hollinger2014-ri}. These techniques have also been applied specifically to ASVs \autocite{Ten_Kathen2023-bq,Kathen2021-pv,Peralta2023-ac,Flaspohler2018-as,Manjanna2017-bm} and even underwater robots \autocite{Kemna2017-da,Girdhar2014-in} for marine environmental monitoring.

\textbf{ASV Systems} In recent years, more ASV systems and algorithms for making autonomous decisions to monitor environments have been built. 
\textcite{Schiaretti2017-tl} classify the autonomy level for ASVs into 10 levels based on control systems, decision-making, and exception handling. 
Many works consider the mechanical, electrical, and control subsystems of their ASV designs \autocite{Ang2022-ep, Madeo2020-yq, Ferri2015-sb}. 
\textcite{Jeong2020-xs} optimized the ASV design to minimize the interference on sensor readings caused by the propulsion system and hull design. 
\textcite{Dash2021-uf} validated the use and accuracy of deploying ASVs for water-quality modelling by comparing the data collected from ASVs with independent sensors, and \textcite{Roznere2021-hv} confirmed that robotic water quality measurements were robust to sensor response time and robot motions. 
More examples of vertically integrated autonomous water-quality monitoring systems using ASVs are presented by \textcite{Chang2021-jz}, \textcite{Cao2020-fc}, and \textcite{Balbuena2017-uj}.
The JetYak platform, introduced in \textcite{Kimball2014-kk}, is a small and inexpensive ASV built for navigating and surveying in shallow or hazardous environments such as glaciers or unexplored ordnance areas. In \textcite{Nicholson2018-el}, the platform has also been retrofitted for large spatial-scale mapping of dissolved carbon dioxide and methane in a marine environment.
Also modelled after JetYak, \textcite{Moulton2018-pa} proposed a more modular and flexible ASV design and discussed many valuable lessons learned to build a fleet of ASVs and their field deployments.
In contrast, our main contribution is a robust mission-planning framework that is complementary to existing designs of ASV systems.

\textbf{Local Motion Planning} Path planning for navigation and obstacle avoidance is a comprehensive field that has been extensively studied \autocite{sanchez2021path}.
The primary purpose of the local planner in this project is to successfully identify and follow a safe path that tracks the global path while averting locally detected obstacles in real-time.
Sampling-based motion planners such as RRT* \autocite{karaman2011sampling} and BIT* \autocite{Gammell_2015} are favourable, owing to their probabilistically complete nature and proven asymptotic optimality given the right heuristics. 
Our local motion planner is based on \textcite{sehn2023beaten}, a variant of the sampling-based planner designed to follow a reference path.  
Using a new edge-cost metric and planning in the curvilinear space, their proposed planner can incrementally update its path to avoid new or moving obstacles without replanning from the beginning while minimizing deviation to the global reference path.
Search-based algorithms, such as D* lite \autocite{koenig2002d} and Field D* \autocite{ferguson2007field}, commonly used in mobile robots and autonomous vehicles, operate on a discretized 2D grid and employ a heuristic to progressively locate a path from the robot's present location to the intended destination. Subsequently, the optimal solution from the path planning is submitted to a low-level controller tasked with calculating the necessary velocities or thrusts in mobile robotics systems. 
Parallel to the planning and control framework, other models such as direct tracking with a constrained model-predictive controller \autocite{ji2016path} and training policies for path tracking through reinforcement learning \autocite{Shan2020ARL} have emerged as new areas of research in recent years.

\textbf{Perception} Lastly, our navigation framework requires local perception modules to clarify uncertainties in our map and avoid obstacles.
Vision-based obstacle detection and waterline segmentation have also received renewed attention in the marine robotics community. 
Recent contributions have largely focused on detecting or segmenting obstacles from RGB images using neural networks \autocite{ewasr, Steccanella2020-he, Qiao2022-mj, Yang2019-yw, Lee2018-gc}. 
A substantial amount of research has been dedicated to identifying waterlines \autocite{Zhou2022-nl, Steccanella2019-rs, Steccanella2020-he, Yin2022-vs} since knowing the whereabouts of navigable waterways can often be sufficient for navigation.
Several annotated datasets collected in different water environments, such as inland waterways \autocite{Cheng2021-mb} and coastal waters \autocite{mastr1325, Bovcon2021-vj} have been published by researchers.
Foundational models for image segmentation, such as `Segment Anything' \autocite{sam}, have also gathered increasing attention due to their incredible zero-shot generalization ability and are being used in tracking \autocite{maalouf2023follow} or remote sensing tasks \autocite{chen2023rsprompter}.
Sonar is another popular sensor that measures distance and detects objects on or under water surfaces using sound waves.
\textcite{Heidarsson2011-jg} pioneered the use of a mechanical scanning sonar for ASV obstacle detection and avoidance and demonstrated that obstacles generated from sonar could serve as labels for aerial images \autocite{Heidarsson2011-ag}.
\textcite{Karoui2015-ft} focused on detecting and tracking sea-surface objects and wakes from a forward-looking sonar image. 
Occupancy-grid mapping, a classic probabilistic technique for mapping the local environment, was used to fuse measurements from sonars and stereo cameras on a mobile ground robot \autocite{elfes1989using}.
For our perception pipeline, we combine the latest advances in computer vision, large datasets from the field, and traditional filtering techniques to make the system robust in real-world operating conditions.
Despite advances, accurate sensor fusion of above-water stereo cameras and underwater sonar for precise mapping on an ASV remains a formidable research challenge.

\vspace{-0.2cm}
\section{Global Mission Planner}
In this section, we will describe the mathematical formulation of the planning problem and present a detailed breakdown of our algorithm. Most of the content in this section, including the problem formulation and the PCCTP-AO* algorithm, has been introduced in our previous work \autocite{boat2022}.

\vspace{-0.2cm}
\subsection{The Problem Formulation}
\label{subsection: problem}
We are interested in planning on a graph representation of a lake where parts of the water are stochastic (i.e., uncertain traversability). 
Constructing such a graph using all pixels of satellite images is impractical since images are very high-dimensional. 
Thus, we extend previous works from CTP \autocite{ctp1991, cctp, ctp-guo2019} 
and distill satellite images into a high-level graph $G$ where some stochastic edges $e$ may be untraversable with probability $p$. 
The state of a stochastic edge can be disambiguated only when the robot traverses the edge in question. 
The robot begins at the starting node $s$ and is tasked to visit all reachable targets $J$ specified by the user (e.g., scientists) before returning to the starting node. 
If some target nodes are unreachable because some stochastic edges block them from the starting node, the robot may give up on these sampling targets.
We call this problem the Partial Covering Canadian Traveller Problem (PCCTP).
Fig. \ref{fig:example_graph} is a simplified graph representation of a lake with two stochastic edges.
The state of the robot is defined as a collection of the following: a list of target nodes that it has visited, the current node it is at, and its knowledge about the stochastic edges. 
A policy sets the next node to visit, given the current state of the robot. 
The objective is to find the optimal policy $\pi^*$ that minimizes the expected cost to cover all reachable targets. 
In the example problem (Fig. \ref{fig:example_graph}), the robot can either disambiguate the left or right stochastic edge to reach the sampling location.
Formally, we define the following terms:
\begin{figure}[t!]
    \centering
    \begin{subfigure}{0.425\textwidth}
        \includegraphics[width=\textwidth]{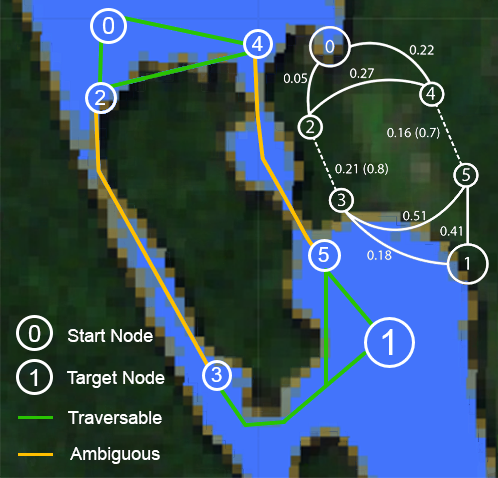}
        \caption{Toy Example Graph}
    \end{subfigure}
    \begin{subfigure}{0.53\textwidth}
        \includegraphics[width=\textwidth]{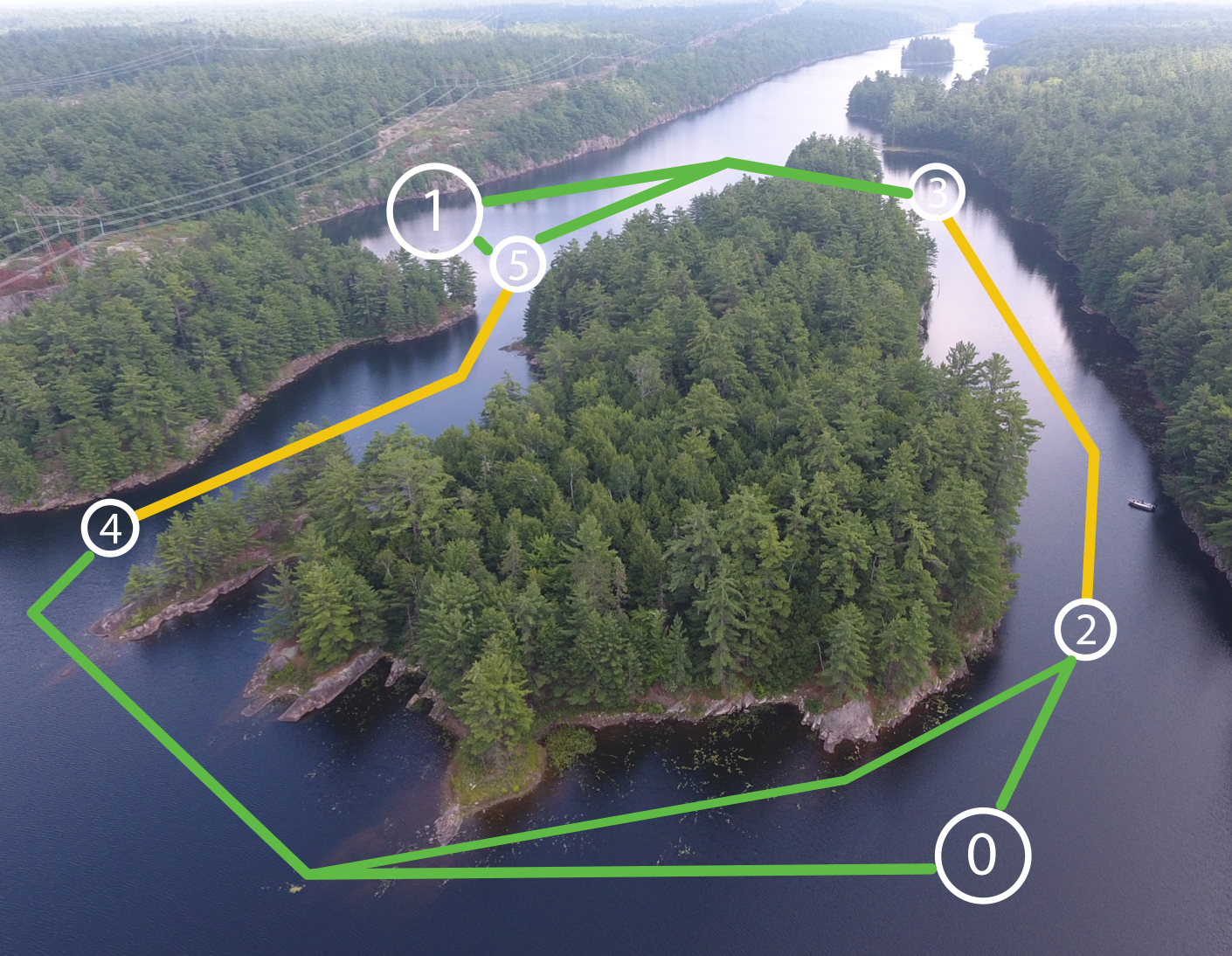}
        \caption{Aerial View (Graph Hand-Sketched)}
    \end{subfigure}
    \caption{A toy example graph shown on the water mask generated from \textit{Sentinel-2} satellite images, with the corresponding graph on an aerial view image shown on the right.
    The planned paths between nodes are simplified for ease of understanding. 
    The number beside each edge of the high-level graph is the path length in km, and the number in brackets is the blocking probability, which is computed using the probability of water coverage in each pixel (represented by its shade of \textcolor{YellowOrange}{orange}) on the path.
    Note that traversable and ambiguous edges are the state before any action.}
    \label{fig:example_graph}
\end{figure}
\begin{itemize}
    \item $G = (V, E)$ is an undirected graph.
    \item $c : E \rightarrow \mathbb{R}_{\geq 0}$ is the cost function for an edge, which is the length of the shortest waterway between two points.
    \item $p : E \rightarrow [0, 1]$ is the blocking probability function. 
    An edge with 0 blocking probability is deterministic; otherwise, it is stochastic.
    \item $k$ is the number of stochastic edges.
    \item $s \in V$ is the start and return node.
    \item $J \subseteq V$ is the subset of target nodes to visit.
    There are $|J| \leq |V| $ goal nodes.
    \item $I = \{ \text{A}, \text{T}, \text{U} \}^k$ is an information vector that represents the robot's knowledge of the status of all $k$ stochastic edges.
    A, T, and U stand for ambiguous, traversable, and untraversable, respectively.
    \item $S \subseteq J$ is the subset of target nodes that the robot has visited.
    \item $a$ is the current node the robot is at.
    \item $x = (a, S, I)$ is the state of the robot. $a$ is the current node, $S$ is the set of visited targets, and $I$ is the current information vector.
    \item $\pi^*$ is the optimal policy that minimizes the cost $\mathbb{E}_{w \sim p(w)} \left[\phi \left(\pi \right)\right]$, where $\phi$ is cost functional of the policy $\pi$ and $w$ is a possible world of stochastic graph, where each stochastic edge is assigned a traversability state. 
\end{itemize}

\begin{figure}[t!]
    \centering
    \includegraphics[width=\columnwidth]{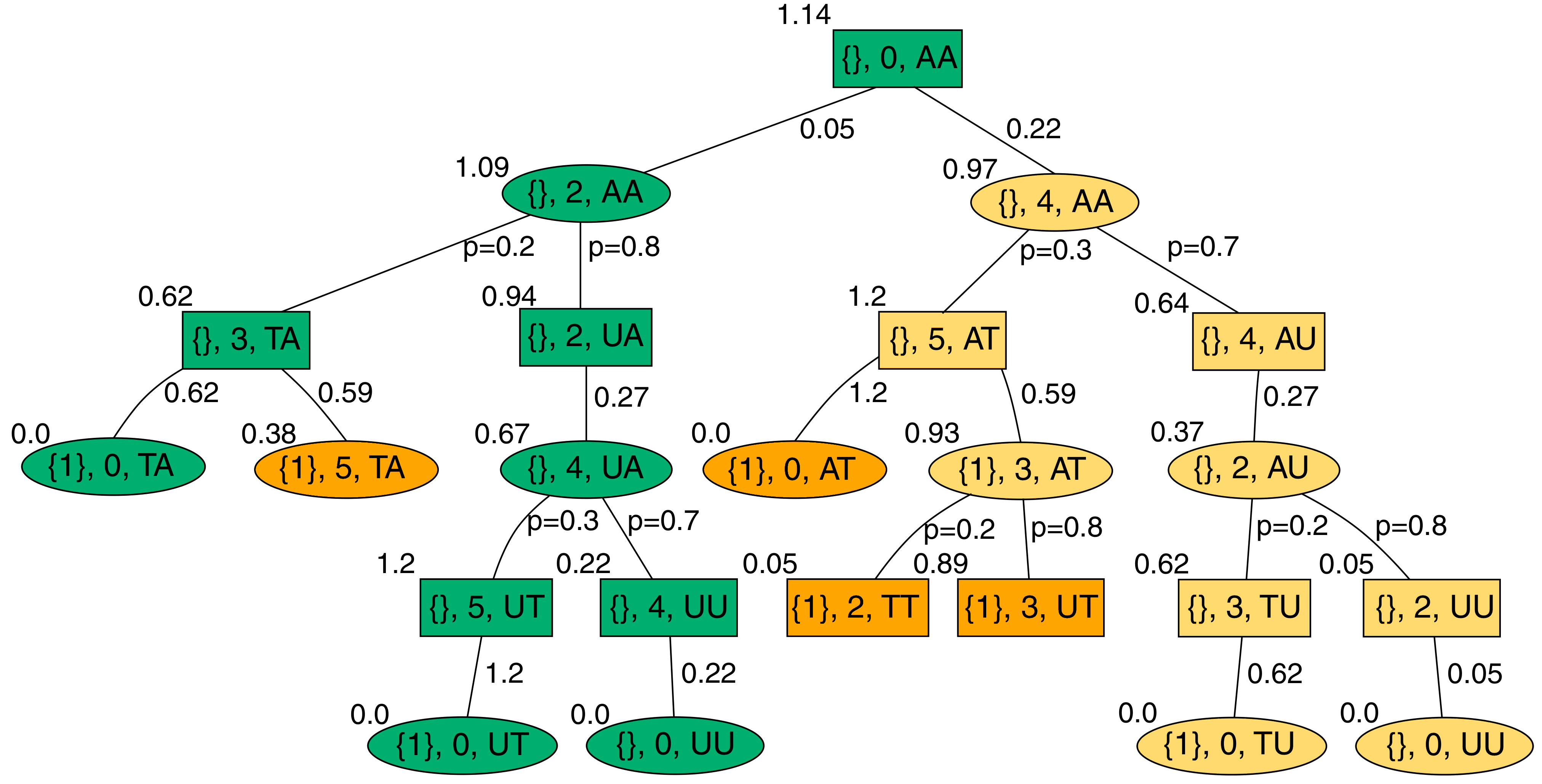}
    \caption{The final AO tree after running PCCTP-AO* on the example in Fig. \ref{fig:example_graph}. 
    The label inside each node is the current state of the robot.
    OR nodes are rectangles, and AND nodes are ellipses. 
    Nodes that are part of the final policy are \textcolor{ForestGreen}{green}, extra expanded nodes are \textcolor{darkyellow}{yellow}, and leaf nodes terminated early are \textcolor{Orange}{orange}. Some \textcolor{Orange}{orange} nodes that are terminated early are left out in this figure for simplicity.    
    This figure is reproduced from Fig. 3 in \textcite{boat2022}.}
    \label{fig:ao_tree}
    \vspace{-0.2cm}
\end{figure}

\subsection{Exactly Solving PCCTP with AO*}

We extend the AO* search algorithm \autocite{ctpao*} used in CTP to find exact solutions to our problem. 
AO* is a heuristic, best-first search algorithm that iteratively builds an AO tree to explore the state space until the optimal solution is found. 
In this section, we will first explain how to use an AO tree to represent a PCCTP instance, then break down how to use AO* to construct the AO tree containing the optimal policy.

\textbf{AO Tree Representation of PCCTP} The construction of the AO tree is a mapping of all possible actions the robot can take and all possible disambiguation outcomes at every stochastic edge. 
Following \textcite{ctpao*}, an AO tree is a rooted tree $T = (N, A)$ with two types of nodes and arcs. 
A node $n \in N$ is either an OR node or an AND node; hence the node set $N$ can be partitioned into the set of OR nodes $N_O$ and the set of AND nodes $N_A$. 
Each arc in $A$ represents either an action or a disambiguation outcome and is not the same as $G$'s edges ($A \neq E$). 
For all $n \in N$, a function $c : A \to \mathbb{R}_{\geq 0}$ assigns the cost to each arc.
Also, for all $n \in N_A$, a function $p : A \to [0, 1]$ assigns a probability to each arc. 
A function $f : N \to \mathbb{R}_{\geq 0}$ is the cost-to-go function if it satisfies the following conditions:
\begin{itemize}
    \item if $n \in N_A$, $f(n) =  \sum_{n' \in N(n)} [p(n, n') \times (f(n') + c(n, n'))]$,
    \item if $n \in N_O$, $f(n) = \min_{n' \in N(n)} [f(n') + c(n, n')]$,
    \item if $n \in N$ is a leaf node, $f(n) = 0$.
\end{itemize}

\begin{algorithm}[t!]
    \caption{The PCCTP-AO* Algorithm}
    \label{algo:pcctp}
    \begin{algorithmic}[1]
    \Require {Graph $G(V, E)$, cost function $c$, heuristic $h$, blocking probability $p$, target set $J$, $k$ stochastic nodes, start node $s$}
    \State $n.a = s$, $n.S = \emptyset$, $n.I = \{A\}^k$
    \State $f(n) = h(n)$; $n$.type = OR; $T.$root = $n$
    \While {$T.$root.status $\neq$ solved}
        \State {$n = $ \Call{SelectNode}{$T.$root}}
        \For {$n' \in \Call{Expand}{n, T}$}
            \State $f(n') = h(n')$
            \If {$\Call{ReachableSet}{J, \; n'.I} \subseteq n'.S$}
            \State $n'.$status = solved
            \EndIf
        \EndFor
        \State \Call{Backprop}{$n, T$}
    \EndWhile
    \If {$T.$root.$f$ == inf}{ \Return No Solution}
    \EndIf
    \State {\Return $T$}

    \Function {Backprop}{$n, T$} \Comment{Update the cost of the parent of $n$ recursively until the root. Same as in \textcite{ctp-guo2019}.
    }
    \While{$n \neq T.$root}
        \If{$n$.type == OR}
            \State $n^* = \text{argmin}_{n' \in N(n)} [f(n') + c(n, n')]$
            \State $f(n) = f(n^*) + c(n, n^*)$
            \If {$n^*.$status == solved}{ $n.$status = solved}
            \EndIf
        \EndIf
        \If{$n.$type == AND}
            \State $f(n) = \sum_{n' \in N(n)} [p(n, n') \times (f(n') + c(n, n'))]$
            \If {$n'.$status == solved $\forall n' \in N(n)$}
            \State $n.$status = solved
            \EndIf
        \EndIf
        \State $n = n.$parent
    \EndWhile
    \EndFunction
    \end{algorithmic}
\end{algorithm}

Now, we can map each node and edge such that the AO tree represents a PCCTP instance.
Specifically, each node $n$ is assigned a label $(n.a, n.S, n.I)$ that represents the state of the robot. 
$n.a$ is the current node, $n.S$ is the set of visited targets, and $n.I$ is the information vector containing the current knowledge of the stochastic edges.
The root node $r$ is an OR node with the label $(s, \emptyset, \mathrm{AA...A})$, representing the starting state of the robot.
An outgoing arc from an OR node $n$ to its successor $n'$ represents an action, which can be either visiting the remaining targets and returning to the start or going to the endpoint of an ambiguous edge via some target nodes along the way. 
An AND node corresponds to the disambiguation event of a stochastic edge, so it has two successors describing both possible outcomes.
Each succeeding node of an OR node is either an AND node or a leaf node.
A leaf node means the robot has visited all reachable target nodes and has returned to the start node.
Each arc $(n, n')$ is assigned a cost $c$, which is the length of travelling from node $n.a$ to node $n'.a$ while visiting the subset of newly visited targets $n'.S \setminus n.S$ along the way.
For all outgoing arcs of an AND node, the function $p$ assigns the traversability probability for the stochastic edge.
The cost of disambiguating that edge is its length.

Once the complete AO tree is constructed, the optimal policy is the collection of nodes and arcs included in the calculation of the cost-to-go from the root of the tree, and the optimal expected cost is $f(r)$.
For example, the optimal action at an OR node $n$ is the arc $(n, n')$ that minimizes the cost-to-go from $n$, while the next action at an AND node depends on the disambiguation outcome.
However, constructing the full AO tree from scratch is not practical since the space complexity is exponential with respect to the number of stochastic edges. 
Instead, we use the heuristic-based AO* algorithm, explained below.

\begin{algorithm}[t!]
    \caption{Algorithm for AO* Expansion}
    \label{algo:expansion}
    \begin{algorithmic}[1]
    \Function {SelectNode}{$n$} \Comment{Find the most promising subtree recursively until reaching a leaf node.}
    \If{$N(n) == $ empty}{ \Return n}{}
    \EndIf
    \State best $= \text{argmin}_{n' \in N(n)} [f(n') + c(n, n')]$
    \State \Return \Call{SelectNode}{best}
    \EndFunction

    \Function {Expand}{$n$, $T$} \Comment{Find the set of succeeding nodes for node $n$ and add them to the tree $T$.} 
    \State $N(n)$ = []
    \If{$n$.type == OR}
        \State $J^R = \Call{ReachableSet}{J, n.I}$
        \State $q$ = Queue(); $q$.append($(n.S, n.a)$)
        \State cost $ = \text{Dictionary}\{(n.S, n.a): 0\}$
        \While{$q$ is not empty} \Comment{Search all paths that end up at an ambiguous edge}
            \State $n = q.$pop()
            \State $A = \{a \mid \Call{hasAmbiguousEdge}{a} \;\; \forall a \in V\}$
            \For {$a' \in (J^R \cup A) \setminus n.S $}
                \State $S' = (n.S \cup \Call{Path}{n.a, a'}) \cap J^R$
                \State $c' = \text{cost}[(n.S, n.a)] + c(n.a, a')$
                \If {$(S', a')$ not in cost or cost$[(S', a')] > c'$}
                \State cost$[(S', a')] = c'$ \Comment{Update the best path and best cost}
                \State $q$.append($(S', a') $)
                \EndIf
            \EndFor
        \EndWhile
        \State costf = Dictionary\{\} \Comment{Find the best route to visit all targets and return to start}
        \For {$(S, a) \in $ costs}
            \State $n' = (S, a, n.I)$;  $N(n)$.append($n'$) \Comment{Add to set of succeeding nodes}
            \If{$S \subseteq J^R$}
            \State costf$[(S, a)]$ = cost$[(S, a)] + c(a, s)$
            \EndIf
        \EndFor
        \State $S^{f}, a^f = \text{argmin}_{S^f, a^f} \text{costf}[(S, a)]$
        \State $n^{f} = (S^{f}, s, n.I)$; $N(n)$.append($n^f$) \Comment{Add to set of succeeding nodes}
    \EndIf
    \If{$n.$type == AND} 
        \For {$e \in \Call{ambiguousEdge}{n.a}$} \Comment{Expand the disambiguation of ambiguous edges}
        \State $n^{T}.I = $ \Call{Unblock}{$n^{T}.I, e$};  $N(n)$.append($n^{T}$)
        \State $n^{U}.I = $  \Call{Block}{$n^{U}.I, e$};  $N(n)$.append($n^{U}$)
        \EndFor
    \EndIf
    \State \Return $N(n)$
    \EndFunction
    \end{algorithmic}
\end{algorithm}

\textbf{PCCTP-AO* Algorithm}
Our PCCTP-AO* algorithm (Algorithm \ref{algo:pcctp} and \ref{algo:expansion}) is largely based on the AO* algorithm \autocite{ao*1971, ao*1978}. 
AO* utilizes an admissible heuristic $h : N \to \mathbb{R}_{\geq 0}$ that underestimates the cost-to-go $f$ to build the AO tree incrementally from the root node until the optimal policy is found.
The algorithm expands the most promising node in the current AO tree based on a heuristic and backpropagates its parent's cost recursively to the root.
This expansion-backpropagation process is repeated until the AO tree includes the optimal policy.

One key difference between AO* and PCCTP-AO* is that the reachability of a target node may depend on the traversability of a set of critical stochastic edges connecting the target to the root.
If a target $j \in J$ is disconnected from the current node $a$ when all the stochastic edges from a particular set are blocked, then this set of edges is critical.
For example, the two stochastic edges in the top-right graph of Fig. \ref{fig:example_graph} are critical because target node 1 would be unreachable if both edges were blocked. 
Thus, a simple heuristic that assumes all ambiguous edges are traversable may overestimate the cost-to-go if skipping unreachable targets reduces the overall cost.

\begin{figure}[t!]
    \centering
    \includegraphics[width=\columnwidth]{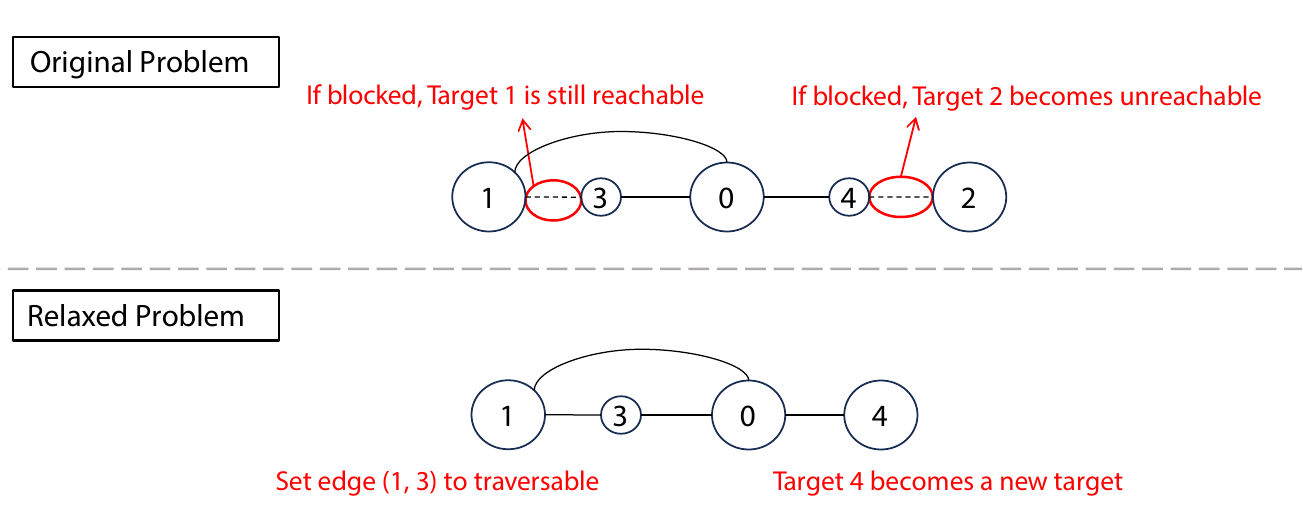}
    \caption{Example of how we relax the original problem graph to calculate the heuristic $h(n)$. At a high level, we construct a relaxed problem by removing all stochastic edges and unreachable nodes from the original graph. Then, the heuristic of the original problem is the cost of the relaxed problem and is always admissible.}
    \label{fig:heuristic}
    \vspace{-0.3cm}
\end{figure}

Alternatively, we can construct the following relaxed problem to calculate the heuristic. 
If a stochastic edge is not critical to any target, we still assume it is traversable.
Otherwise, we remove the potentially unreachable target for the robot and instead disambiguate one of the critical edges of the removed target.
The heuristic is the cost of the best plan that covers all definitively reachable targets and disambiguates one of the critical stochastic edges. 
For example, consider computing the heuristic at starting node 0 in Fig. \ref{fig:heuristic}. 
The goal is to visit both nodes 1 and 2 if they are reachable. 
Node 1 is always reachable; hence we assume it is traversable in the relaxed problem.
Node 2 may be unreachable, so we remove the stochastic edge (4, 2) and ask the boat to visit Node 4 instead in the relaxed problem.
This heuristic is always admissible because the path to disambiguate a critical edge is always a subset of the eventual policy.
We can compute this by constructing an equivalent generalized travelling salesman problem \autocite{set-tsp} and solve it with any optimal TSP solver.

Fig. \ref{fig:ao_tree} shows the result of applying PCCTP-AO* to the example problem in Fig. \ref{fig:example_graph}.
The returned policy (coloured in \textcolor{ForestGreen}{green} nodes) tries to disambiguate the closer stochastic edge $(2, 3)$ to reach target node 1.
Note that the AO* algorithm stops expanding as soon as the lower bound of the cost of the right branch exceeds that of the left branch.
This guarantees the left branch has a lower cost and, thus, is optimal.

\vspace{-0.2cm}
\subsection{Estimating Stochastic Graphs From Satellite Imagery}
\label{sec:graph_estimation}
We will now explain our procedure to estimate the high-level stochastic graph from satellite images.

\begin{figure}[b!]
    \centering
    \includegraphics[width=\columnwidth]{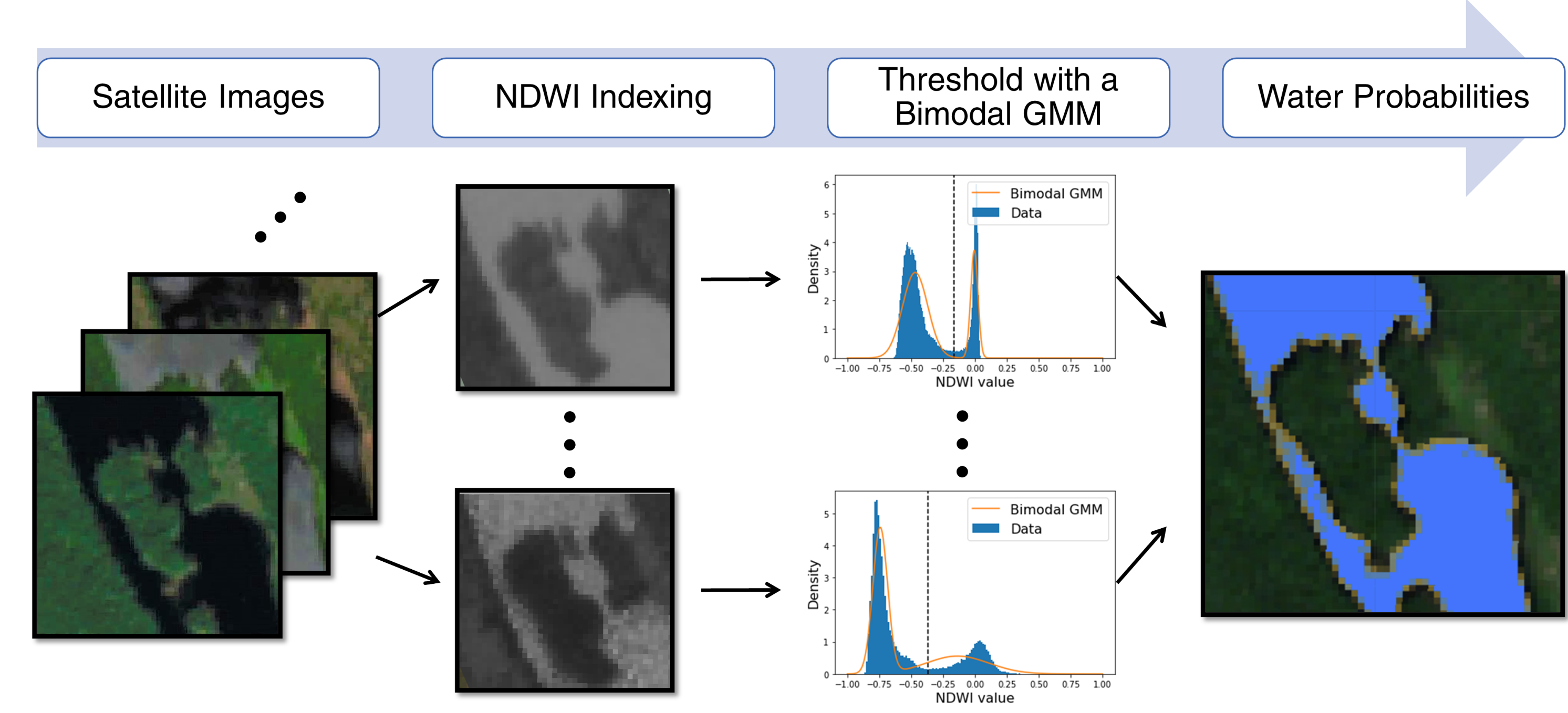}
    \caption{Overview of the water-masking process for deriving water probabilities from satellite imagery. The procedure begins with historical Sentinel-2 satellite images displayed on the left. Water pixels are individually identified in each image by first calculating the NDWI water index and then using a bimodal Gaussian Mixture Model for classification. The results of each classification are averaged to determine the probabilities of water, which are depicted on the right. Pixels with reduced water probabilities are coloured more \textcolor{darkyellow}{yellow}. }
    \label{fig:water-masking}
\end{figure}

\textbf{Water Masking} Our first step is to build a water mask of a water area across a specific period (e.g., 30 days).
We use the \textit{Sentinel-2} Level 2A dataset \autocite{sentinel-2}, which has provided multispectral images at 10 m by 10 m resolution since 2017.
Each geographical location is revisited every five days by a satellite. 
We then select all satellite images in the target spatiotemporal window and filter out the cloudy images using the provided cloud masks.
For each image, we calculate the Normalized Difference Water Index (NDWI) \autocite{ndwi} for every pixel using green and near-infrared bands.
However, the distribution of NDWI values varies significantly across different images over time.
Thus, we separate water from land in each image and aggregate the indices over time.
We then fit a bimodal Gaussian Mixture Model on the histogram of NDWIs to separate water pixels from non-water ones for each image.   
We average all water masks over time to calculate the probabilistic water mask at the target spatiotemporal window.
Each pixel on the final mask represents the probability of water coverage on this 10 m by 10 m area.
If the probability of water for a pixel is greater than 90\%, we treat it as a deterministic water pixel.
We then classify pixels with a probability lower than 90\% but greater than 50\% as stochastic water pixels. 
Finally, we identify the boundary of all deterministic water pixels.
Fig. \ref{fig:water-masking} shows an overview of these steps.

\begin{figure}[t!]
\centering
     \begin{subfigure}[b]{0.43\textwidth}
         \centering
         \includegraphics[width=\textwidth]{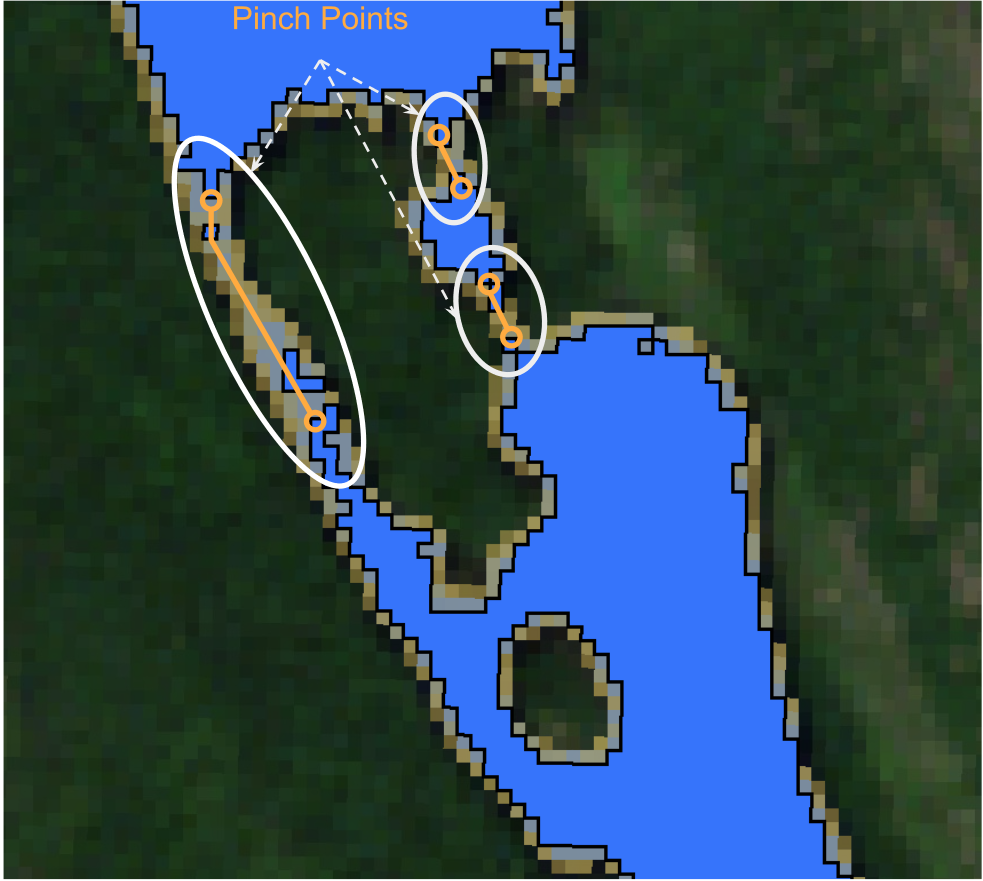}
         \caption{Example of Pinch Points}
         \label{fig:pinch_point}
     \end{subfigure}
     \hfill
     \begin{subfigure}[b]{0.53\textwidth}
         \centering
         \includegraphics[width=\textwidth]{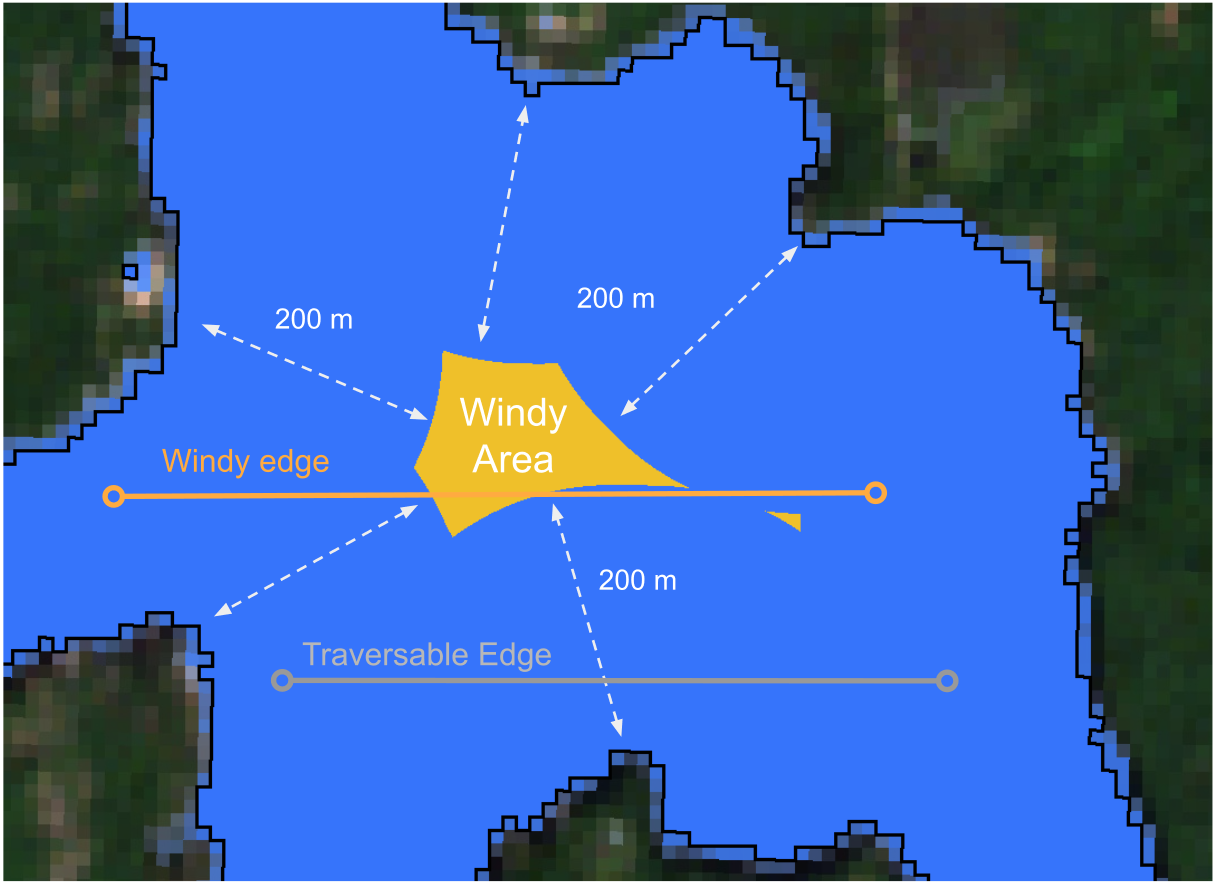}
         \caption{Example of a Windy Edge}
         \label{fig:windy_edges}
     \end{subfigure}
     \hfill
\caption{Satellite images illustrating two types of stochastic edges. Water pixels are marked in blue, with their estimated boundaries in black. The left image demonstrates several pinch points, highlighted in orange, that represent potential paths connecting water pixels from two distinct water bodies that are otherwise far or disconnected. The image on the right visually describes the concept of a windy edge. Any water pixel at least 200 metres from the boundary falls within the \textcolor{darkyellow}{yellow} windy area. If an edge crosses the windy area, then it is classified as a windy edge.}
\end{figure}

\textbf{Stochastic Edge Detection: Pinch Points} We can now identify those stochastic water paths (i.e., narrow straits, pinch points \autocite{Ferguson2004-cf}) that are useful for navigation. 
A pinch point (e.g. Fig. \ref{fig:pinch_point}) is a sequence of stochastic water pixels connecting two parts of topologically far (or distinct) but metrically close water areas.
Essentially, this edge is a shortcut connecting two points on the water boundary that are otherwise far away or disconnected. 
To find all such edges, we iterate over all boundary pixels, test each shortest stochastic water path to nearby boundary pixels, and include those stochastic paths that are shortcuts.
The blocking probability of a stochastic edge is one minus the minimum water probability along the path. 
Since this process will produce many similar stochastic edges around the same narrow passage, we run DBSCAN \autocite{dbscan} and only choose the shortest stochastic edge within each cluster. 

\textbf{Stochastic Edge Detection: Windy Edges} The second type of stochastic edges are those with strong wind. In practice, when an ASV travels on a path far away from the shore, there is a higher chance of running into a strong headwind or wave, making the path difficult to traverse. Hence, we define a water pixel to be a windy area if it is at least 200m away from any points of the water boundary. An edge is then treated as a windy edge if it crosses the windy area at some point and we assign a small probability for the event where the wind blocks the edge. An example of a windy area and an associated windy edge is shown in Fig. \ref{fig:windy_edges}.

\textbf{Path Generation} The next step is to construct the geo-tagged path and calculate all edge costs in the high-level graph.
The nodes in the high-level graph are composed of all sampling targets, endpoints of stochastic edges, and the starting node. 
We run A* \autocite{hart1968formal} on the deterministic water pixels to calculate the shortest path between every pair of nodes except for the stochastic edges found in the previous step.
Since the path generated by A* connects neighbouring pixels, we smooth them by randomized shortcutting \autocite{Geraerts2007-mq}.
Then, we can discard any unnecessary stochastic edges if they do not reduce the distance between a pair of nodes.
For every stochastic edge, we loop over all pairs of nodes and check if setting the edge traversable would reduce the distance between the pair of nodes.
Finally, we check if each deterministic edge is a windy edge and obtain the high-level graph used in PCCTP.

In summary, we estimate water probabilities from historical satellite images with adaptive NDWI indexing and build a stochastic graph connecting all sampling locations and pinch points. The resulting compact graph representing a PCCTP instance can be solved optimally with an AO* heuristic search.

\vspace{-0.2cm}
\section{Simulations}

In this chapter, we will verify the efficacy of our PCCTP planning framework in a large-scale simulation of mission planning on real lakes. The testing dataset and simulation results from this section can be reproduced from our previous work in \textcite{boat2022}.

\vspace{-0.2cm}
\subsection{Testing Dataset}
We evaluate our mission-planning framework on Canadian lakes selected from the \textit{CanVec Series} Ontario dataset \autocite{canvec}. Published by \textit{Natural Resources Canada}, this dataset contains geospatial data of over 1.1 million water bodies in Ontario.
Considering a practical mission length, lakes are filtered such that their bounding boxes are 1-10 km by 1-10 km.
Then, water masks of the resulting 5190 lakes are generated using \textit{Sentinel-2} imagery across 30 days in June 2018-2022 \autocite{sentinel-2}.
We then detect any pinch points on the water masks and randomly sample five different sets of target nodes on each lake, each with a different number of targets. The starting locations are sampled near the shore to mimic real deployment conditions. 


Furthermore, we generate high-level graphs and windy edges from the water mask. 
Graphs with no stochastic edges are removed as well as any instances with more than ten stochastic edges due to long run times.
Ultimately, we evaluate our algorithm on 2217 graph instances, which come from 1052 unique lakes. 

\begin{figure}[b!]
 \begin{subfigure}[b]{0.53\textwidth}
         \centering
         \includegraphics[width=\textwidth]{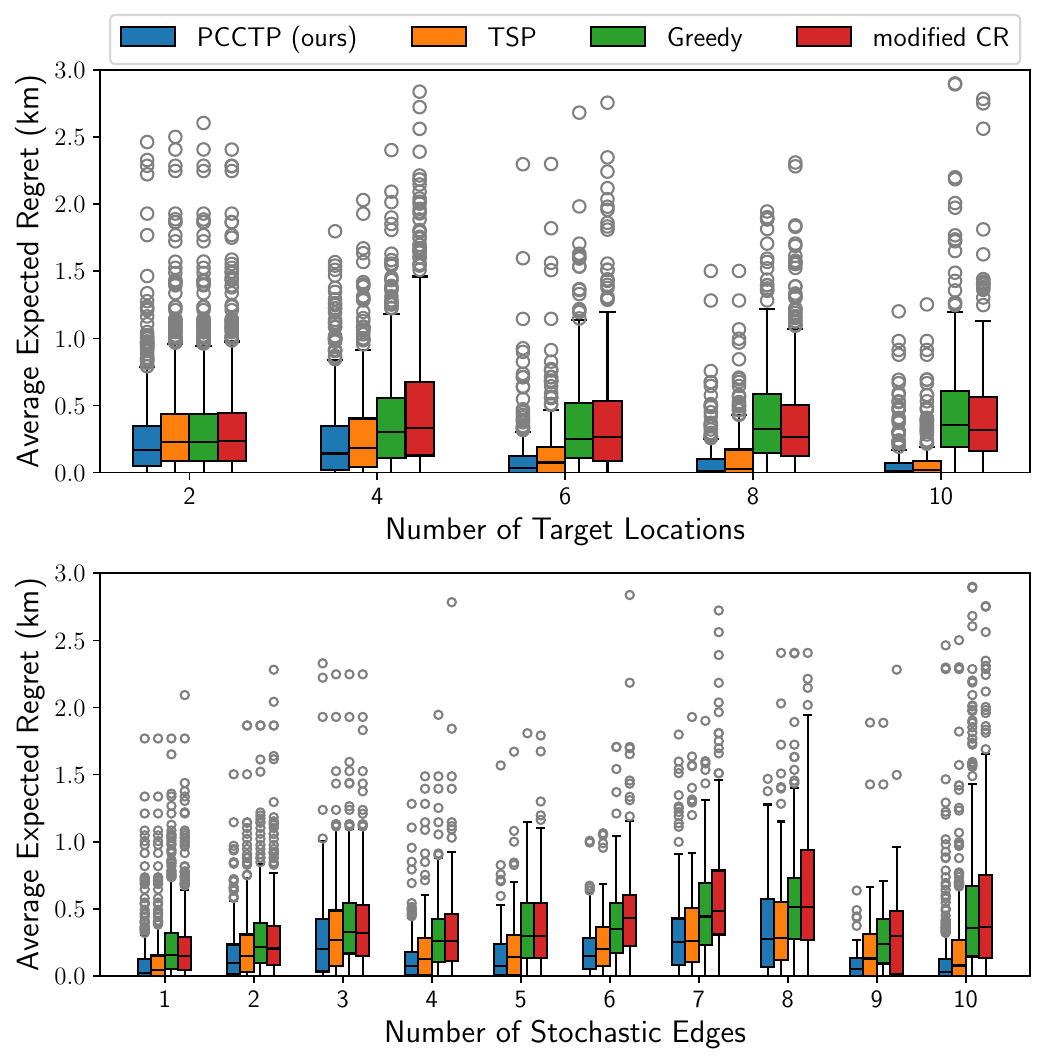}
         \caption{Average expected regret of PCCTP and baselines}
         \label{fig:sim_result}
     \end{subfigure}
     \hfill
     \begin{subfigure}[b]{0.42\textwidth}
         \centering
         \includegraphics[width=\textwidth]{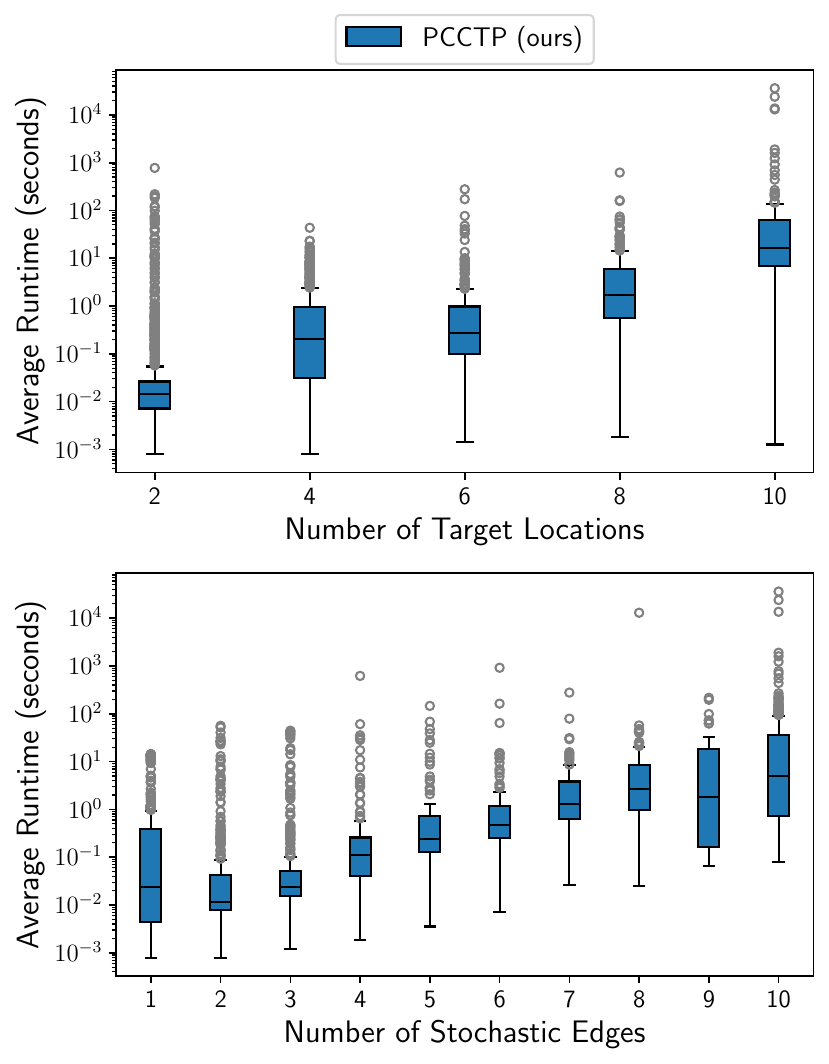}
        \caption{Average offline runtime of PCCTP}
        \label{fig:sim_runtime}
     \end{subfigure}
     \hfill
    \centering
    \caption{Results of PCCTP and baselines in simulation. In (a), the performance of PCCTP is compared against three baselines. Our proposed method achieves the lowest average expected regret and outperforms the next-best baseline by 1.8km in the extreme case. Note that the stochastic edges include both windy edges and pinch points. In (b), only the CPU execution time for PCCTP is shown, since all baselines are online methods.}
    \label{fig:sim-all}
\end{figure}

\subsection{Baseline Planning Algorithms}

The simplest baseline is an online greedy algorithm that always goes to the nearest unvisited target node assuming all ambiguous edges are traversable. 
For a graph with $k$ stochastic edges, we simulate all $2^k$ possible worlds, each with a different traversability permutation, and evaluate our greedy actor on each one.
The greedy actor recomputes a plan at every step and queries the simulator if it encounters a stochastic edge to disambiguate it.
Also, it checks the reachability of every target node upon discovering an untraversable edge and gives up on any unreachable targets.

A more sophisticated baseline is the optimistic TSP algorithm. 
Instead of always going to the nearest target node, it computes the optimal tour to visit all remaining targets assuming all ambiguous edges are traversable.  
Similar to the greedy actor, TSP recomputes a tour at every step and may change its plan after encountering an untraversable edge.
The expected cost is computed via a weighted sum on all $2^k$ possible worlds. 
In contrast to PCCTP, both baselines require onboard computation to update their optimistic plans, whereas PCCTP precomputes a single optimal policy that is executed online.

Lastly, we modify the CR algorithm, originally a method for CCTP \autocite{cctp}, to solve PCCTP.
CR precomputes a cyclic sequence to visit all target nodes using the Christofides algorithm \autocite{Christofides1976WorstCaseAO} and tries to visit all target nodes in multiple cycles while disambiguating stochastic edges.
If a target node turns out to be unreachable, we allow CR to skip this node in its traversal sequence. 

\vspace{-0.2cm}
\subsection{Results}
\vspace{-0.2cm}
Fig. \ref{fig:sim_result} compares our algorithm against all baselines.
To measure the performance across various graphs of different sizes, we use the average expected regret over all graphs. 
The expected regret of a policy $\pi$ for one graph $G$ is defined as
\begin{equation*}
\mathbb{E}_w[\text{Regret} (\pi)] = \sum_{w} [ p(w)(\phi(\pi, w) - \phi(\pi^p, w)) ], \
\end{equation*}
where $\pi^p$ is a privileged planner with knowledge of the states of all stochastic edges, $\phi$ is the cost functional, and  $w$ is a possible state of (the stochastic edges of) the graph.
The cost $\phi(\pi^p, w)$, calculated using a TSP solver for each state, serves as a lower bound to the costs incurred by the policy $\phi(\pi, w)$.
A low expected regret indicates that the policy $\pi$ will find efficient paths to visit all target locations and disambiguate the stochastic edges without prior knowledge of their states.

PCCTP precomputes the optimal policy in about 50 seconds on average in our evaluation, and there is no additional cost online.
Compared to the strongest baseline (TSP), our algorithm saves the robot about 1\%(50m) of travel distance on average and 15\%(1.8km) in the extreme case.
Although the advantage is not statistically significant on average, our planner still offers advantages in many specific scenarios and edge cases, such as those with high blocking probabilities or long stochastic edges. 
The performance of PCCTP may be further enhanced if the estimated blocking probabilities of the stochastic edges are refined based on historical data.

We also find that the performance gap between our algorithms and baselines becomes more significant with more windy edges. 
In fact, if the only type of stochastic edges in our graph is pinch points (i.e., the number of windy edges is 0), the performance gap is almost negligible between PCCTP and the optimistic TSP baseline.
The main reason is that most pinch points only reduce the total trip distance by hundreds of meters on a possible state of the graph.
Pinch points are most likely to be found either on the edges of a lake or as the only water link connecting two water bodies. 
In the first case, these pinch points are unlikely to be a big shortcut.
As for the latter case, if the pinch point is the only path connecting the starting location to a target node, disambiguating this edge has to be part of the policy.
On the other hand, windy edges passing through the centre of a lake are often longer, and the gap between the optimal and suboptimal policy is much more significant.

\textbf{Computational Complexity} The worst-case complexity of our optimal search algorithm is $O(|J|! \times k! \times 2^k)$, which depends on number of stochastic edges $k$ and number of target nodes $|J|$. The complexity is exponential in nature because there are $2^k$ possible states of the stochastic edges, and all possible orders to visit all nodes and disambiguate stochastic edges need to be enumerated without a good heuristic in the worst case. 

In practice, however, our implementation performs efficiently on standard laptop CPUs.  The median runtime of our algorithm, implemented in Python, is less than one second, and 99\% of the instances run under 3 minutes.
Nonetheless, approximately 0.5\% of instances with eight nodes and 4.2\% with ten nodes require more than five minutes to process.
We believe the runtime can be considerably improved by rewriting in a more efficient language, such as C++.
More importantly, we argue that this one-time cost can occur offline before deploying the robot into a water-sampling mission.
Although the worst-case runtime of the AO* algorithm can increase exponentially as the graph increases in size, the number of target locations in each graph cannot grow infinitely for real-world water-sampling missions.
Hence, the runtime of PCCTP is not a concern for practical applications.

\section{Autonomous Navigation System} 

This section will explain our local navigation framework in detail and how the robot can execute the mission-level policy and safely follow its planned trajectory.

\subsection{Stochastic Edge Disambigutaion}
\begin{figure}[t!]
    \centering
    \includegraphics[width=\columnwidth]{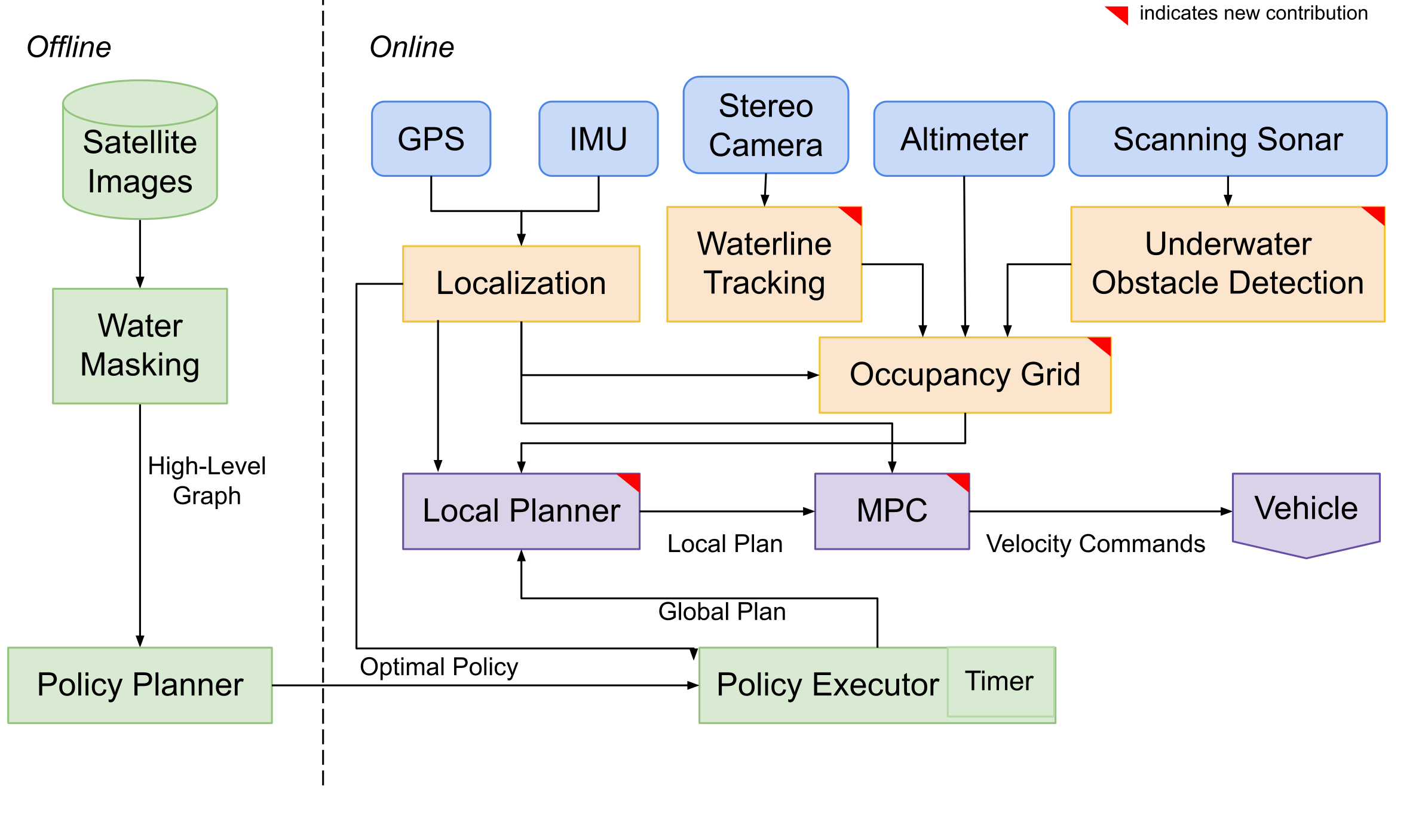}
    \caption{The autonomy modules of our navigation system. Global mission planners are coloured in \textcolor{LimeGreen}{green}, sensors inputs are labelled in \textcolor{SkyBlue}{blue}, localization and local mapping nodes are shaded in \textcolor{Apricot}{orange} while planning and control nodes are in \textcolor{Thistle}{purple}. We have also specifically indicated the modules where we made significant improvements over our previous work \textcite{boat2022}.  }
    \label{fig:autonomy_stack}
    \vspace{-0.2cm}
\end{figure}

One crucial aspect required for fully autonomous policy execution is the capacity to disambiguate stochastic edges.  
Our approach is to build a robust
autonomy framework (Fig. \ref{fig:autonomy_stack}) that relies less on lower-level components such as perception and local planning to execute a policy successfully.
In more general terms, the mission planner precomputes the navigation policies from satellite images given user-designated sampling locations. 
During a mission, the robot will try to follow the global path published by the policy.
Sensor inputs from a stereo camera and sonar scans are processed and filtered via a local occupancy-grid mapper.
The local planner then tries to find a path in the local frame that tracks the global plan and avoids any obstacles detected close to the future path of the robot. 
When the robot is disambiguating a stochastic edge, the policy executor will independently decide the edge's traversability based on the GPS location of the robot and a timer. 
A stochastic edge is deemed traversable if the robot reaches the endpoint of the prescribed path of this edge within the established time limit. 
If it fails to do so, the edge is deemed untraversable. 
There is no explicit traversability check on an ambiguous stochastic edge, such as a classifier or a local map.
The timer allows us to address complications we cannot directly sense, such as heavy prevailing winds or issues with the local planner.
Following this, the executor branches into different policy cases depending on the outcome of the disambiguation.

We made significant improvements to our local navigation framework compared to our previous work in \autocite{boat2022}. Similar to before, the traversability assessment uses a timer and GPS locations to classify an edge's traversability without directly relying on the result of obstacle detections or local mapping. Instead, we made design decision changes to the local planning architecture and improvements to individual modules. Below, we will explain all the important components and highlight any changes we made.

\subsection{Terrain Assessment with Stereo Camera}
An experienced human paddler or navigator can easily estimate the traversability of a lake by visually distinguishing water from untraversable terrains, obstacles, or any dynamic objects. 
In our previous work in \textcite{boat2022}, the video stream collected from the stereo camera is processed geometrically by estimating the water surface from point clouds and clustering point clouds above the water surface as obstacles.
This process was prone to stereo-matching errors due to sun glares and calm water surfaces, and could not detect any obstacles on the water surface such as a shallow rock. 
To address these shortcomings, we use semantic information from RGB video streams and neural stereo disparity maps to estimate traversable waters in front of the robot and identify obstacles.  
We learn a water segmentation network and bundle it with a temporal filter to estimate the waterline in image space and remove outliers.
The estimated waterline is then projected to 3D using the disparity map and used to update the occupancy grid. 
We provide more details in the following sections.

\subsubsection{Water Segmentation Network}
\begin{figure}[b!]
    \centering
    \subfloat[Sub-Optimal Exposure]{
        \includegraphics[width=0.45\columnwidth]{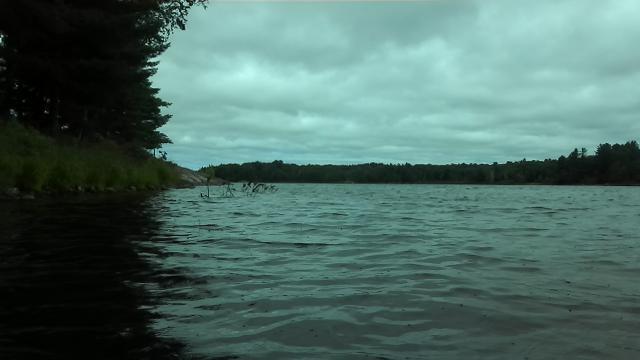}
        \label{fig:seg_contrast}
    }
    \hfill
    \subfloat[Reflections in Still Water]{
        \includegraphics[width=0.45\columnwidth]{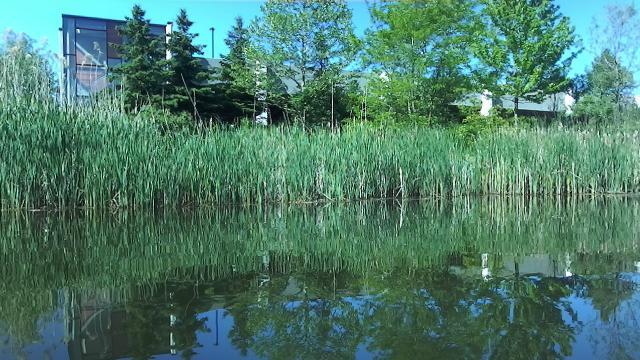}
        \label{fig:seg_reflections}
    }
    \caption{Examples of challenging conditions for semantic segmentation and disparity mapping.}
    \label{fig:seg_examples}
\end{figure}

The most important factor in training a robust neural network for water segmentation is a large and diverse dataset. The characteristics of water's appearance exhibit considerable variation contingent on factors such as wind, reflections, and ambient brightness, as demonstrated in Fig. \ref{fig:seg_examples}. Yet, the stereo camera falters in difficult lighting conditions due to the lack of dynamic range, culminating in inadequately exposed images and the emergence of artifacts such as shadows, lens flare, and noises. 

\begin{figure}[t!]
    \centering
    \includegraphics[width=1\columnwidth]{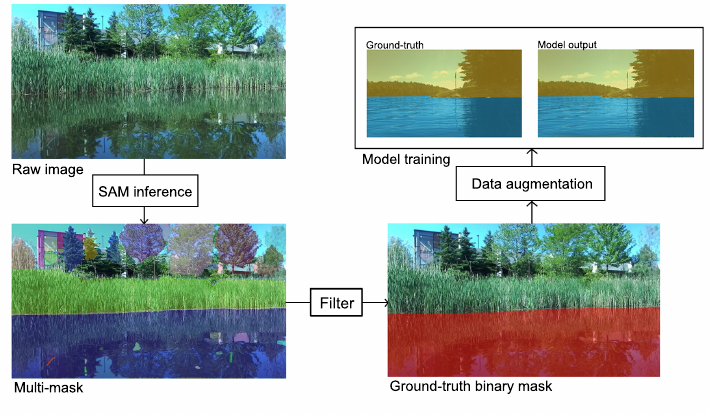}
    \caption{The steps in the automatic process of generating ground-truth labels. Each distinct colour overlay represents a different object as segmented by SAM. The \textcolor{red}{red} region is the final ground-truth water mask after filtering SAM masks, which is used for training our own water segmentation network.}
    \label{fig:seg}
    \vspace{-0.2cm}
\end{figure}

Our image dataset is collected from previous field tests in Nine Mile Lake and a stormwater management pond at the University of Toronto. However, manual annotation of thousands of images is impractical due to its labour and time intensity. Thus, since semantic segmentation is a well-explored research area, we used a pre-trained SAM (Segment Anything Model) to automate the process of creating ground-truth labels \autocite{sam}. SAM will try to segment everything beyond just water, outputting numerous masks of different irrelevant items. While it is not yet capable of classifying labelled regions, because water normally occupies the lower half of the frame and is commonly characterized by substantial area and continuity, we can apply a heuristic that heavily favours these features, scoring regions to distinguish water mask $m_{\rm water}$ from other masks with very high accuracy: 
\[m_{\rm water}=\max_i\left[\frac{A(m_i)}{d(m_i) + 1}\right],\]
where $m_i$ denotes the mask of class $i$ from SAM, $A$ computes the total number of pixels a mask occupies, and $d$ represents the vertical distance, in pixels, of the masked area's centroid from the image's bottom. Then, false positives within the identified mask will be filtered out. With manual checking, we found that this simple approach successfully labelled the entirety of our dataset without failure. An example of this process is shown in Fig. \ref{fig:seg}. Finally, we have a binary mask ready to be fed into training.

\begin{figure}[t!]
\centering
\includegraphics[width=1\columnwidth]{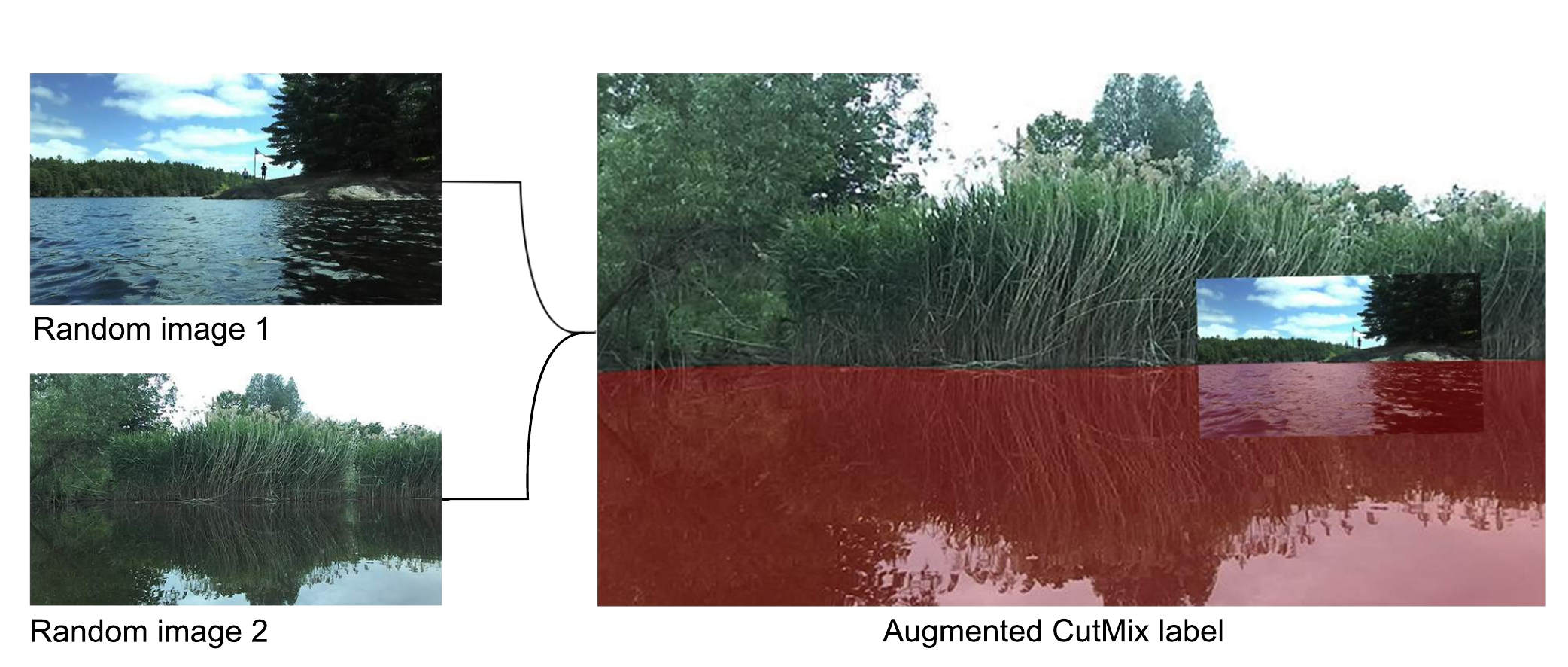}
\caption{Example of CutMix augmented training data. A second image with a random exposure multiplier is randomly resized and placed on top of the original image. The \textcolor{red}{red} region highlights the pixels that are labelled as water.}
\label{fig:cutmix}
\end{figure}

Another important technique to improve the quality of neural networks is data augmentation. Our limited training set cannot match all the possible lighting and environmental conditions that the ASV may encounter in future missions. However, we would like the trained network to be robust against issues such as bad exposure and reflections, which significantly affect segmentation performance. To this end, we use colour-jittering and CutMix \autocite{cutmix} during training and we find that they greatly enhance out-of-distribution performance, yielding superior generalization in challenging weather conditions as in Fig. \ref{fig:seg_examples}. Essentially, regions of one training image are cut and overlaid onto another, as demonstrated in Fig. \ref{fig:cutmix} to encourage the model to learn more diverse and challenging features while also expanding our limited dataset.

\begin{figure}[t!]
    \centering
    \subfloat[Original Images]{
        \includegraphics[width=\columnwidth]{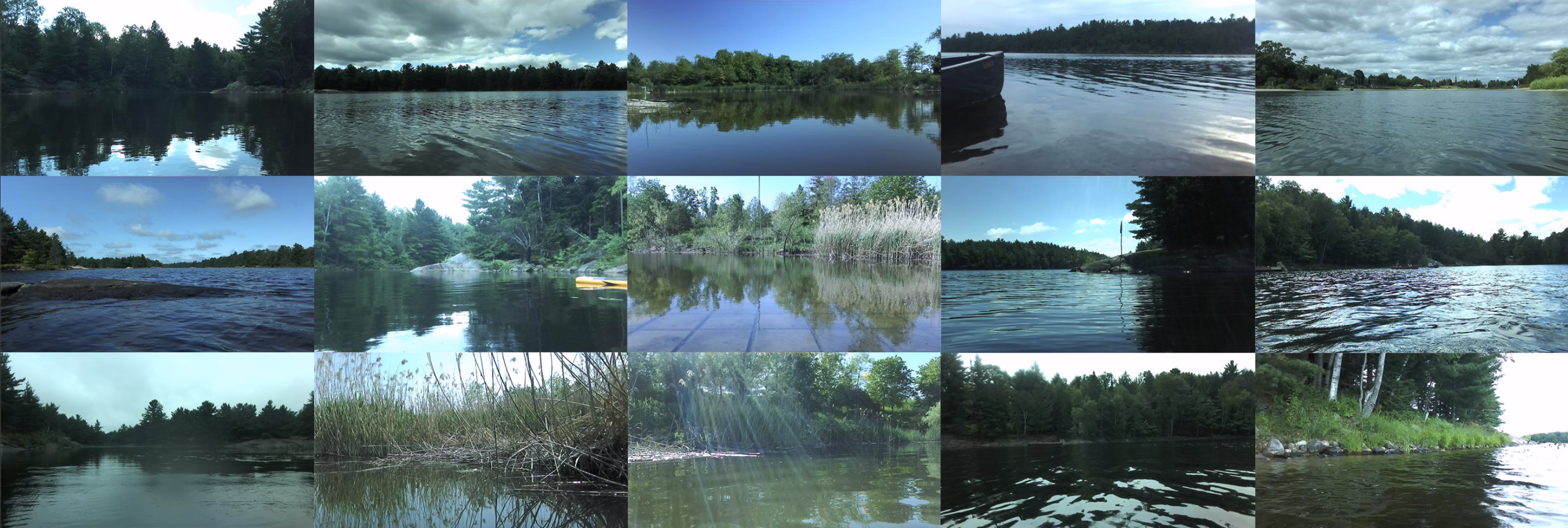}
        \label{fig:tset_ex}
    }
    \hfill
    \subfloat[Predicted Water Masks]{
        \includegraphics[width=\columnwidth]{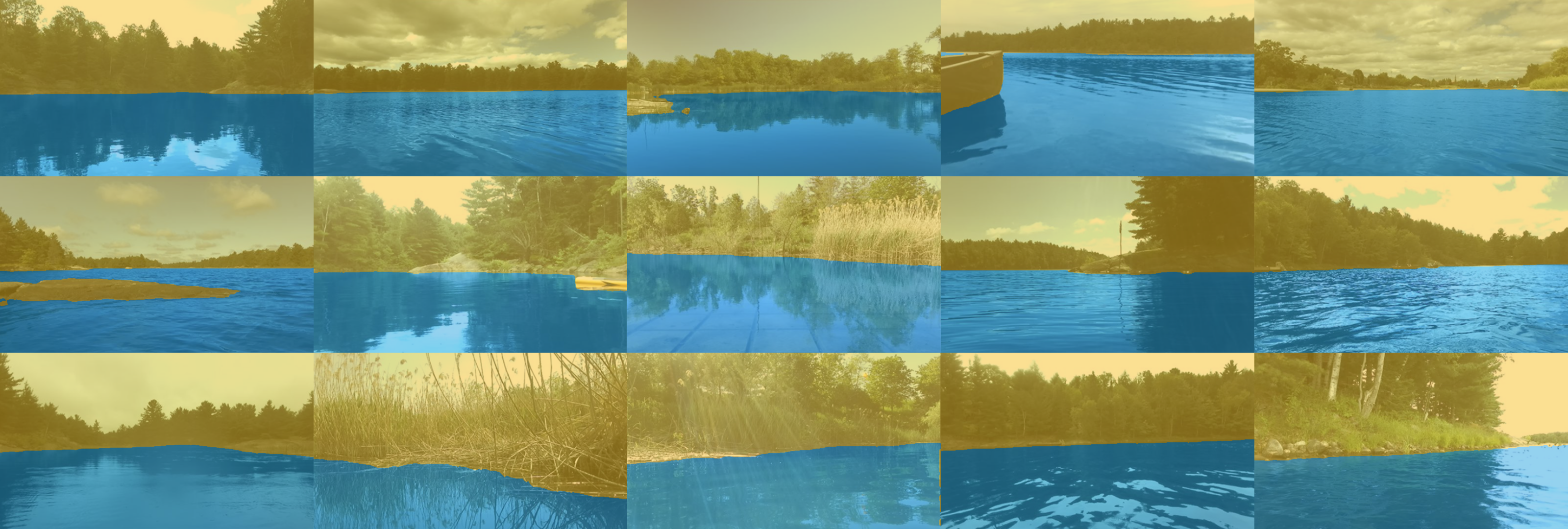}
        \label{fig:tset_pred}
    }
    \caption{Example images and their predicted water mask from our test set. To better assess model capabilities, we hand-picked a diverse set of challenging scenarios, such as reflections, bad exposure, strong glares, aquatic plants, shallow areas, and windy water surfaces. Despite these challenges, our trained water segmentation network reliably produces high-quality water masks.}
    \label{fig:tset}
    \vspace{-0.2cm}
\end{figure}

Our model architecture and pre-trained weights are adopted from the eWaSR maritime obstacle detection network based on the ResNet-18 backbone \autocite{ewasr}. Our training dataset contains 4,000 images while our testing split contains 200 images. Images in the training set are sampled uniformly from video recordings of past experiments, while the images of the test set are held out from challenging scenarios that we manually identify. Figure \ref{fig:tset_ex} displays some examples from the test set. In addition, 10 more labels are generated randomly using CutMix during training for every original labelled image. 

Inspired by the semantic segmentation community \autocite{long2015fully, mastr1325}, we use Intersection over Union (IOU, also known as the Jaccard Index) for water pixels as a metric to evaluate the performance of the trained segmentation network. Let $n_{tp}$ be the number of water pixels with the correct predictions, $n_{p}$  be the total number of pixels classified as water by the neural network, and $n_{g}$ be the total number of pixels labelled as water. The IOU can then be calculated as $n_{tp} / (n_{g} + n_{p} - n_{tp})$. After training, we achieve an average IOU of 0.992 on the 200-image hold-out test set. The results of the predicted water masks after training are shown in Figure. \ref{fig:tset_pred}, and our lightweight yet powerful neural network consistently produces binary masks that accurately and consistently segment water.

\subsubsection{Waterline Estimation and Tracking}

There are several issues associated with the direct use of raw segmentation masks produced by a neural network. 
Firstly, the 2D, per-pixel water labels are not inherently suited for determining traversability in front of the robot. 
Secondly, depth estimations derived from the stereo camera can be severely distorted due to unfavourable conditions such as sun glare or tranquil water reflections. 
In \textcite{boat2022}, the geometry-based approach to estimating water planes from point clouds derived from depth images was highly sensitive to noise and required substantial manual adjustments. 
Lastly, both neural segmentation masks and depth maps can exhibit noise and inconsistency over successive timestamps. 
These issues necessitate avoiding the direct combination of segmentation masks with depth maps to ascertain the existence of a 3D water surface. 
Instead, we filter the segmentation masks both spatially and temporally to approximate a waterline in 2D image space, then project this line into 3D space. 
This projected line then forms the basis for traversability estimates in 3D based on stereo data.

\begin{figure}[t!]
    \centering
    \includegraphics[width=\columnwidth]{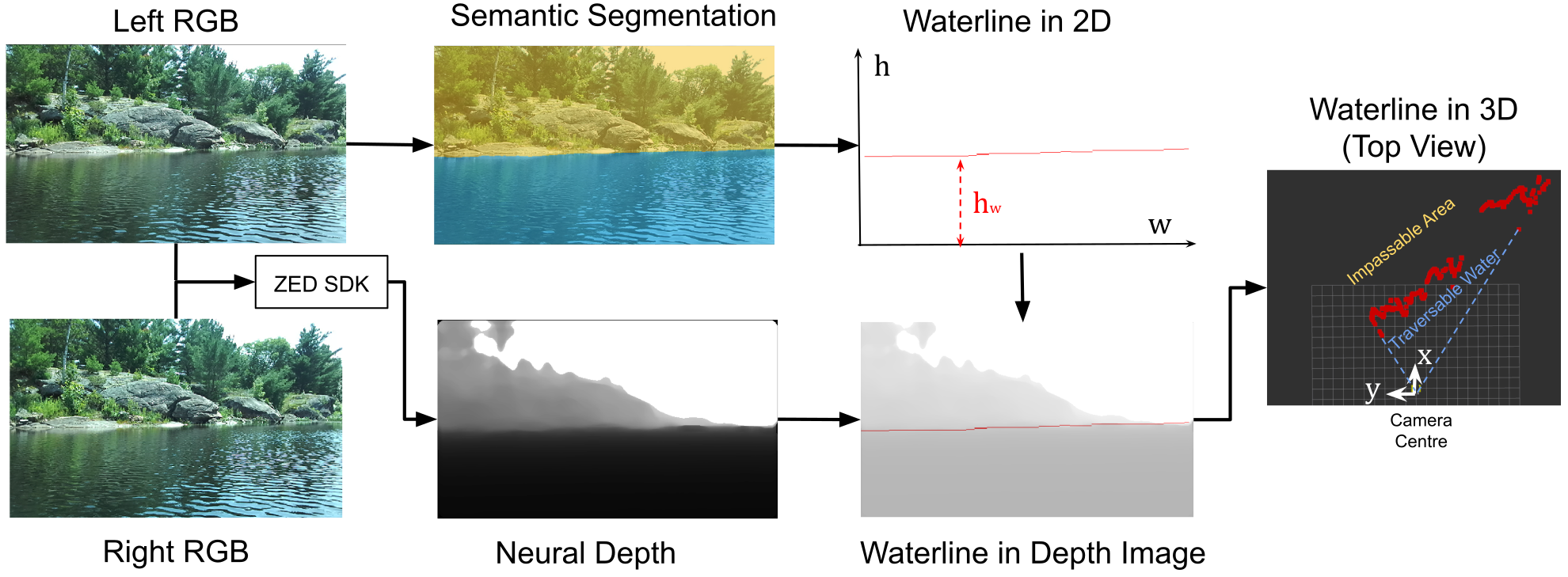}
    \caption{The stereo-based waterline estimation pipeline. A waterline in the 2D image is estimated and tracked using the water segmentation masks. The 2D waterline is mapped to the depth image generated by the ZED SDK and reprojected into the 3D camera frame based on the depth of waterline pixels and camera intrinsics. The final \textcolor{Red}{red} points in top view represent the 3D projection of the estimated waterline and separate the traversable water in front of the robot from the impassable area.}
    \label{fig:stereo-pipeline}
\end{figure}

We approximate the 2D waterline as a vector comprising $n$ elements, where $n$ represents the image's width. Each element serves to indicate the waterline's position for that column. 
The fundamental premise here is that each column contains a clear division between water and everything else – such as the sky, trees, people, shoreline buildings, and other dynamic obstacles. 
Thus, we can presume that only the pixels below the waterline are navigable, while those above are impassable. 
This model works well because water surfaces are typically horizontal when viewed from the first-person perspective of the ASV. 
Therefore, for the purpose of evaluating the robot's forward navigation, we can safely disregard any water pixels higher than the defined waterline in the image space.
The position of the waterline on every column is identified by scanning upwards from the column's bottom until a non-water region is detected using a small moving window. 
If $s$ is the window size, the separation point is the first pixel from the bottom such that the next $s$ pixels above are all non-water.
Usually, the window size is five.

Our filtering process consists of two stages: spatial filtering based on RANSAC \autocite{ransac} and employing a Kalman filter subsequently for temporal tracking of the waterline. 
We design the spatial filtering step to smooth the waterline and remove spatial outliers; to this end, we employ nearest-neighbour interpolation to fit the random samples in each iteration. 
RANSAC uses the squared loss function to compare the interpolated waterline and the raw waterline.
Then, we apply a linear Kalman filter with outlier rejection to track each individual element (column) of the waterline temporally. 
The Kalman filter uses RANSAC-filtered waterline as observations and maintains an estimated waterline as the state. 
Both the state transition matrix and the observation matrix are identities.
We use a chi-squared test to discard outliers, which compares the normalized innovation squared to a predetermined threshold. 
Using both filters, we can eliminate noises in the segmentation mask and mitigate any temporal oscillation or abrupt changes in the predicted water segmentation masks. In practice, we find that the quality of filtering is not very sensitive to the parameters of the RANSAC and the Kalman filter. 

At the end of the filtering process, we have a smoothed 2D waterline with one pixel per column that separates navigable water from everything else. We project this line back to the camera's 3D frame. As shown in Fig. \ref{fig:stereo-pipeline}, we use the depth coordinate of each waterline pixel and calculate their 3D positions with the intrinsics of the stereo camera. In the end, the 3D waterline separates the traversable water in front of our ASV from any obstacles originally above the 2D waterline. If the 2D waterline is at the horizon (i.e., it separates water and the sky), the projected 3D waterline will be very far away or close to infinity, meaning that all fields of view in front of the robot are traversable water.

\begin{figure}[t!]
 \begin{subfigure}[b]{0.5\textwidth}
         \centering
         \includegraphics[width=\textwidth]{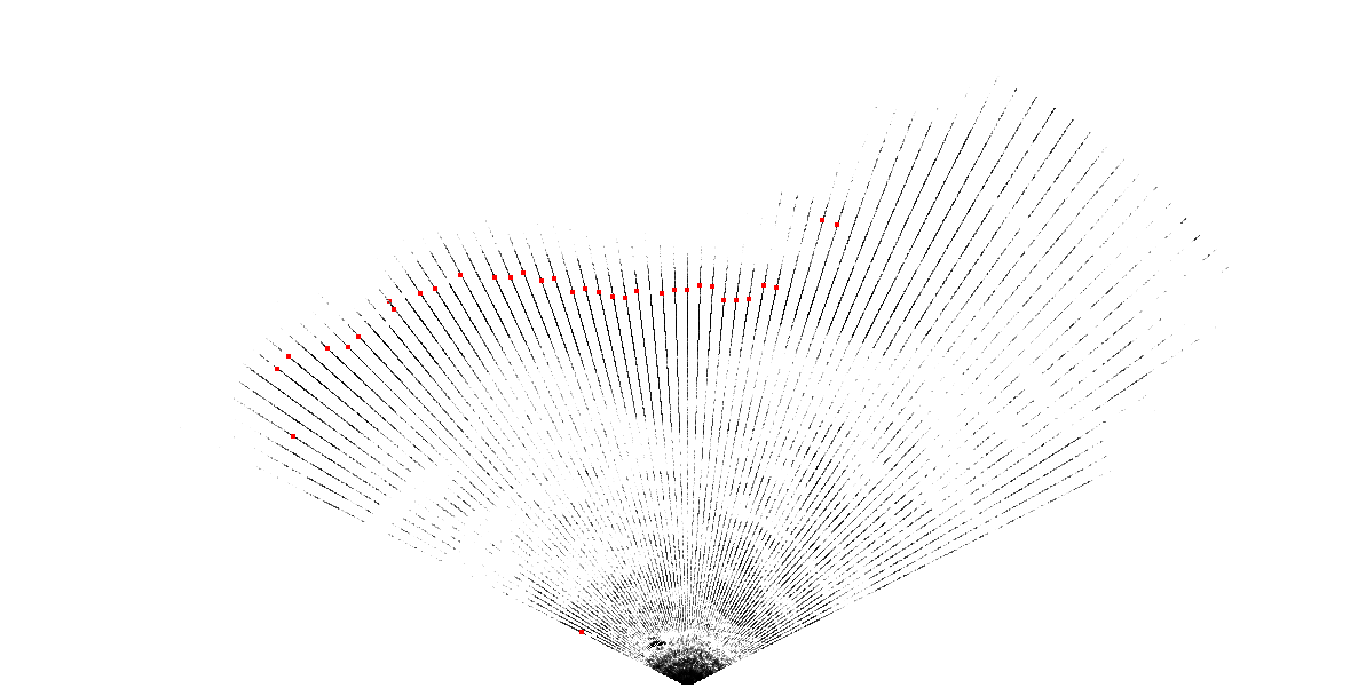}
         \caption{Sonar Scan Bird's Eye View. Detections are \textcolor{Red}{red}.}
         \label{fig:sonar_det_2d}
     \end{subfigure}
     \hfill
     \begin{subfigure}[b]{0.45\textwidth}
         \centering
         \includegraphics[width=\textwidth]{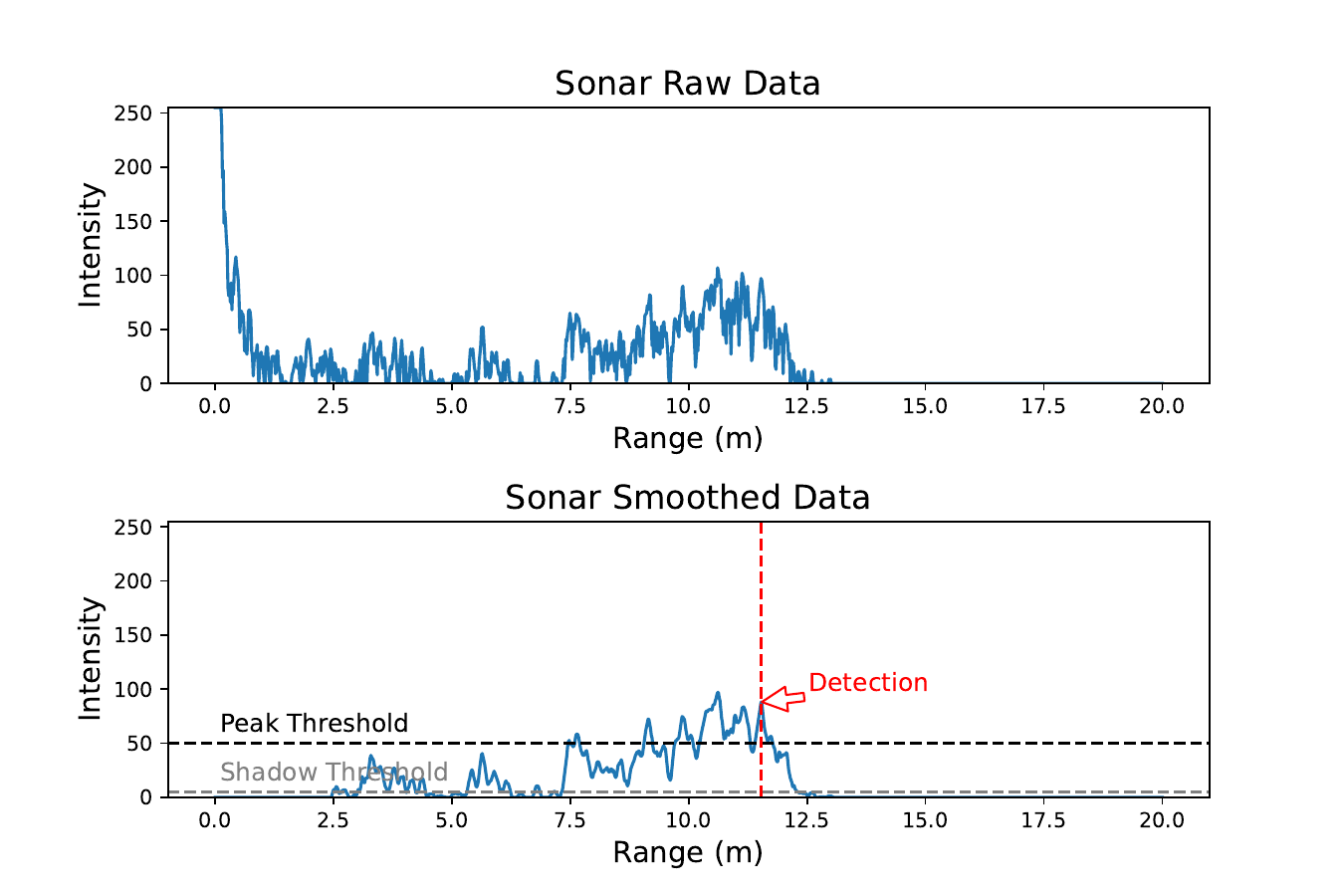}
         \caption{A Single Sonar Scan and its detection}
         \label{fig:sonar_det_single}
     \end{subfigure}
     \hfill
    \centering
    \caption{A sonar scan and obstacle detection result. The scan is taken from the same scene and timestamp as in Fig. \ref{fig:stereo-pipeline}, and the sonar successfully detected the shoreline visible from the bird's eye view on the left. Each sonar scan is processed individually by detecting the first local maxima above a peak threshold on the smoothed data. Consecutive sonar scans are used to filter out noise in the detected obstacles. }
    \label{fig:sonar_det}
\end{figure}

\vspace{-0.2cm}
\subsection{Obstacle Detection with Sonar}
\vspace{-0.2cm}

Sonar is commonly used as a sensor in maritime applications for both ships and submarines. 
A specific type, the Blue Robotics Ping360 mechanical scanning sonar, serves as our primary sensing module underwater. 
It is mounted underwater and operates by emitting an acoustic beam within a forward-facing fan-shaped cone. 
This beam has a consistent width (1°) and height (20°). 
The sonar then records the echoes reflected by objects, with the reflection strength relating directly to the target's density. 
By measuring the return time and factoring in the speed of sound in water, the range of these echoes can be determined. 
The sonar's transducer can also be rotated to control the horizontal angle of the acoustic beam. 
Configured to scan a 120° fan-shaped cone ahead of the boat, the sonar can complete these scans up to a range of 20m in approximately 3.5 seconds.
Additionally, we also have a Ping1D Sonar Echosounder from Blue Robotics that measures water depth. The echosounder is mounted underwater and is bottom-facing.
Each sonar scan yields a one-dimensional vector that corresponds to the reflection's intensity along the preset range.
If an obstacle impedes the path of the acoustic beam, it prevents the beam from passing beyond the obstruction, leading to an acoustic shadow.
This phenomenon facilitates obstacle detection via sonar scanning.

Fig. \ref{fig:sonar_det_2d} illustrates a typical sonar scan cycle that detects obstacles. 
A single sonar scan's raw and processed data with the resulting detected obstacle are shown in Fig. \ref{fig:sonar_det_single}.
The process begins with the removal of noisy reflections within a close range ($<$2.5m) before smoothing the scan using a moving-average filter. 
Following this, all local maxima above a specific peak threshold (50) are detected. 
An obstacle is identified at the first local maxima, where the average intensity post-peak falls below the shadow threshold (5). These thresholds are tuned by hand on data collected from previous field tests in Nine Mile Lake and the stormwater management pond at the University of Toronto. 

A post-processing filter removes detections that do not persist across a minimum of $n$ scans (with $n=2$ in our configuration). 
This is accomplished by calculating the cosine similarity between the current intensity vector and its predecessor. 
If an obstacle is consistently detected $n$ times, and the cosine similarity across these successive intensity vectors exceeds 0.9, along with spatial proximity, this detected obstacle point is included. 
In other words, any detections occurring in isolation, either spatially or temporally, are excluded.
In our previous work~\autocite{boat2022}, sonar was only used for data collection purposes and not for local planning or navigation. Using scanning sonar, we can significantly improve our ability to detect shallow or underwater obstacles even if sonar operates at a much lower frequency than the stereo camera.

\vspace{-0.2cm}
\subsection{Sensor Fusion with Local Occupancy Grid}
\vspace{-0.2cm}

\begin{figure}[t!]
    \centering
    \includegraphics[width=\columnwidth]{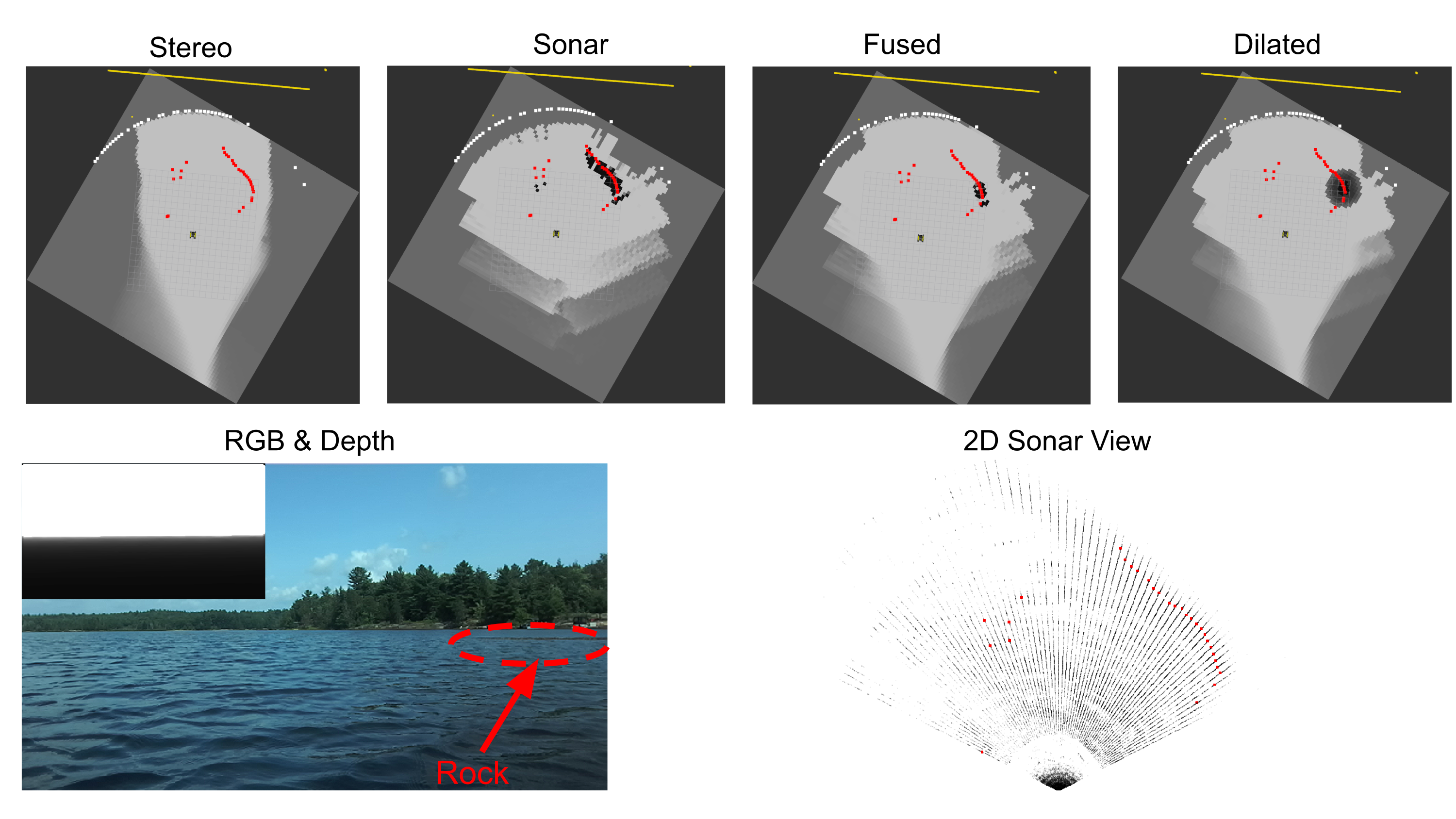}
    \caption{Example of sensor fusion with occupancy map before an extruding rock. \textcolor{darkyellow}{Yellow} line is the waterline estimated by stereo camera, \textcolor{Red}{red} dots indicate underwater obstacles detected by the scanning sonar, and white dots mean that the sonar did not detect an obstacle at that angle. The rock was successfully detected in the final occupancy grid despite being missed by the stereo camera. }
    \label{fig:occgrid}
\end{figure}
Upon receiving detections from the sonar scans and the stereo camera, they are fused into a coherent local representation to facilitate local path planning and robot control. 
We utilize the classic occupancy map \autocite{elfes1989using} for our local mapping representation. 
Unlike a 2D naive cost map used in our previous work \autocite{boat2022}, an occupancy grid maintains a local map in a principled fashion and naturally filters and smoothes sensor measurements temporally and spatially. 
The traversability of each cell is determined by naively summing the separately maintained log-odds ratios for sonar and camera. 
Our occupancy grid is 40m x 40m, with a cell resolution of 0.5m x 0.5m, its centre moving in sync with the robot's odometry updates.
Waterline points, as detected by the stereo camera, are ray-traced in 3D back to the robot, thus lowering the occupied probability of cells within the ray-tracing range. 
Cells containing or adjacent to waterline points have their occupied probabilities increased. 
However, points exceeding a set maximum range do not affect occupied probabilities beyond the maximum range due to the decreasing reliability of depth measurements with increasing range.
The protocol for updating the log-odds ratios for sonar is similar.
Each sonar scan is ray-traced to clear the occupancy grid and marks any cells containing or close to the obstacles. 
The log-odds ratios of existing cells are decayed with incoming measurement updates, enhancing the map’s adaptability to noisy localizations, false positives, and dynamic obstacles. 
Finally, we apply a median filter to the occupancy grid to smooth out and remove outliers. 

A limitation of this system is that the scanning sonar and the stereo camera observe different sections of the environment. 
The sonar may detect underwater obstacles invisible to the camera and vice versa for surface-level objects. 
Fig. \ref{fig:occgrid} provides an example where a shallow rock in the front-right of the ASV is detected by the sonar but missed by the stereo camera. 
Without ample ground-truth data on the marine environment, reconciling discrepancies between these sensors proves challenging. 
Traversability estimation, especially in shallow water, is also complicated due to the potential presence of underwater flora (e.g. Fig. \ref{fig:shallow_water_and_aquatic_plants}) or terrain.
As a solution, we opt for the simplest fusion method: directly summing the log-odds ratio in each cell. 
Additionally, we adjust the occupancy grid dilation based on the echosounder's water depth measurements, increasing the dilation radius when the ASV is in shallower water.
The workflow of this strategy is shown in Fig. \ref{fig:occgrid}.
While this strategy may only present a coarse traversability estimate, it still reliably detects the shoreline despite possible undetected smaller obstacles such as lily pads or weeds. 
The dilation adjustment employed in shallow water allows the ASV to navigate safely, avoiding prevalent aquatic plants near the shore.

\subsection{Local Path Tracking and Control}
\label{sec:local_plan}

\begin{figure}[b!]
\centering
\includegraphics[width=\textwidth]{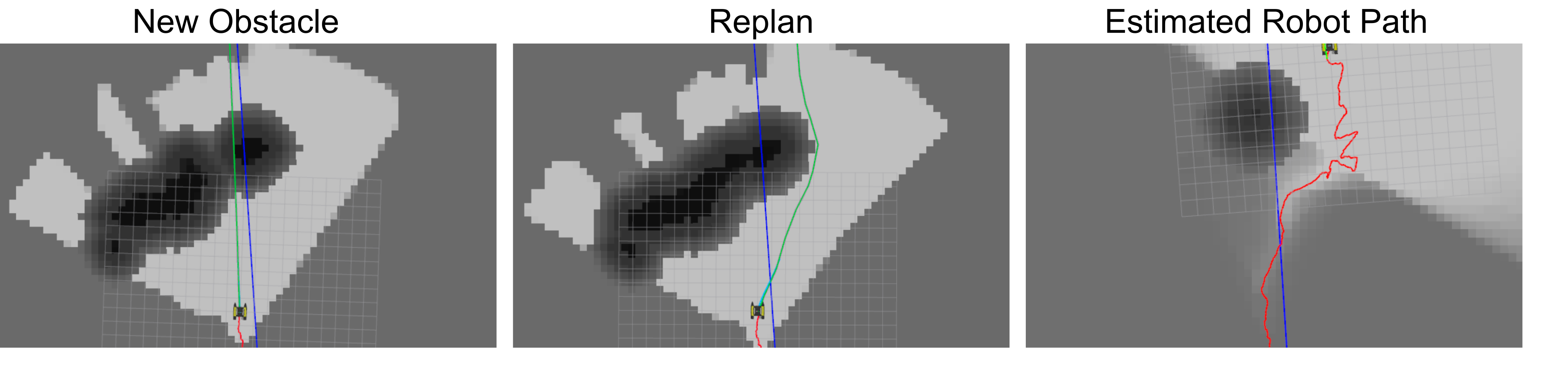}
\caption{Example of the planner replanning around an obstacle and avoiding it. \textcolor{Blue}{Blue} line is the global plan (see Sec. \ref{sec:graph_estimation} for details). \textcolor{ForestGreen}{Green} is the current local plan planned using the local occupancy grid and tries to stay close to the global plan as much as possible (see Sec. \ref{sec:local_plan}). \textcolor{Red}{Red} is the robot's actual trajectory estimated by the GPS. The actual trajectory of the robot is jagged due to both noisy GPS signals, wind, and decaying occupancy map. }
\label{fig:replan}
\vspace{-0.2cm}
\end{figure}

Local path tracking is essential to ensure that the robot adheres to the global mission plan while navigating around obstacles on the local map. The desired controller should run in real-time and work well in obstacle-rich environments. The generated local paths must also be easy for the robot to follow.  In our previous work, the robot's velocities were directly controlled using the Dynamic Window Approach (DWA) \autocite{dwa} to avoid local obstacles. However, DWA only samples a single-step velocity and may fail in cluttered environments with obstacles or cul-de-sacs. Direct tracking of the global path with model predictive control (MPC) is another popular option \autocite{mpc2017ji, mpc2020dong}, but solving the optimization problem can be costly because the obstacle avoidance constraint is nonconvex in general. In this paper, we use an alternative strategy that uses a separate path planner to find a collision-free path that connects to the global path and then tracks the collision-free path with an MPC. We employ a modified version of Lateral BIT*, as proposed by \textcite{sehn2023beaten}, to serve as both our local planner and controller. Our approach provides a stronger guarantee as BIT* is probabilistically complete and asymptotically optimal. 

This optimal sampling-based planner, set within the VT\&R \autocite{vtr2010} framework, follows an arbitrary global path while veering minimally around obstacles.
Lateral BIT* builds upon BIT* \autocite{Gammell_2015} by implementing a weighted Euclidean edge metric in the curvilinear planning domain, with collision checks performed against the occupancy grid in the original Euclidean space.
Samples are pre-seeded along the whole global path in the curvilinear coordinates before random sampling in a fixed-size sampling window around the robot.
The planner operates backward from a singular goal node to the current robot location without selecting any intermediate waypoints.
Lateral BIT* is also an anytime planner and can be adapted for dynamic replanning.
Once an initial solution is found, an MPC tracking controller can track the solution path.
The MPC optimizes the velocity commands in a short horizon to minimize the deviation from the planner solution while enforcing robot kinematic models and acceleration constraints.
Adopted from \textcite{sehn2023beaten}, the MPC solves the following least-squares problem:
$$
\underset{\mathbf{T}, \mathbf{u}}{\text{argmin}} J(\mathbf{T}, \mathbf{u}) = \sum_{k=1}^{K} \ln(\mathbf{T}_{\text{ref}, k}\mathbf{T}_k^{-1})^{\vee^{T}}\mathbf{Q}_k\ln(\mathbf{T}_{\text{ref}, k}\mathbf{T}_k^{-1})^{\vee} + \mathbf{u}_k^T\mathbf{R}_k\mathbf{u}_k 
$$
s.t.
$$
\mathbf{T}_{k+1} = \exp \left( (\mathbf{P}^T\mathbf{u}_k)^{\wedge}h \right)\mathbf{T}_{k}, k=1, 2, ... K
$$
$$
\mathbf{u}_{\text{min}, k} \leq \mathbf{u}_k \leq \mathbf{u}_{\text{max}, k}, k=1, 2, ... K$$

where $\mathbf{T} \in SE(3)$ are poses and $\mathbf{u} = [v \; \omega]^T$ are velocities. The objective function minimizes the pose error between the reference trajectory $\mathbf{T}_{\text{ref}, k}$ and the predicted trajectory $\mathbf{T}_{k}$ while keeping the control effort $\mathbf{u}_k$ minimum. 
The two constraints are the generalized kinematic constraint and actuation limits. 
We tune the cost matrices $\mathbf{Q}$ and $\mathbf{R}$ to balance the cost between different degrees of freedom.
We refer readers to Sec. V of \textcite{sehn2023beaten} for more details.

If a newly detected obstacle obstructs the current best solution path, the planner will truncate its planning tree from the obstacle to the robot, triggering a replan or rewire from the truncated tree to the robot's location.  
Because the resolution of the satellite map is low (10m/cell), our global path could be blocked by large rocks and terrains, especially the pinch points. 
Hence, we adjust the maximum width and length of the sampling window and tune the parameters balancing lateral deviation and the path length. 
If there are no viable paths locally within the sampling window and the planner cannot find a solution after 1 second, the controller will stop the ASV and stabilize it at its current location.

In practice, we cannot directly control the ASV's linear velocity due to the primary source of translational velocity estimates, GPS data, is noisy and unreliable. Consequently, we map the linear velocity commands to motor thrusts through a linear relationship and close the control loop using the MPC tracking controller. Fig. \ref{fig:replan} illustrates an example from the field test where our robot detected obstacles and effectively replanned its trajectory. In the middle image, the lateral BIT* planner finds a smooth path around the obstacle while deviating minimally from the global path. The estimated robot path on the right appears jagged due to significant GPS noise (up to 1m), wind and current influences, and the occupancy grid's time decay, causing the ASV's heading to oscillate and repeatedly rediscover the same obstacle. However, the robot successfully bypasses the obstacle without requiring manual intervention. This highlights the robustness and adaptability of our architecture in dynamic, noisy, and unpredictable environments.

\section{Real World Experiments}
\subsection{Robot}
\begin{figure}[t!]
    \centering
    \begin{subfigure}[b]{1\textwidth}
        \centering
        \includegraphics[width=0.75\linewidth]{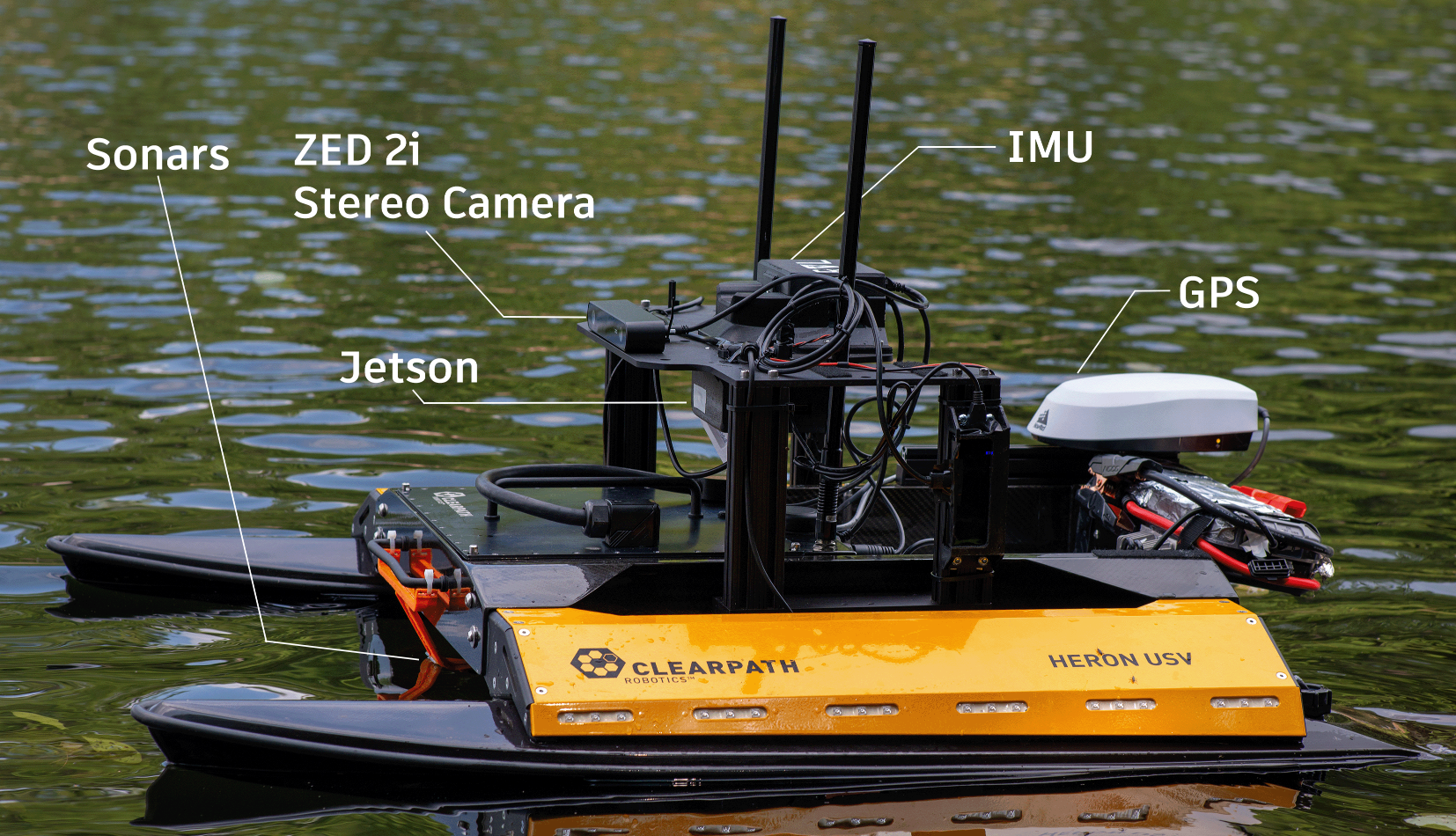}
        \caption{Top View}
        \label{fig:robot_topview}
    \end{subfigure}
    \\
    \begin{subfigure}[b]{0.35\textwidth}
        \includegraphics[width=\linewidth]{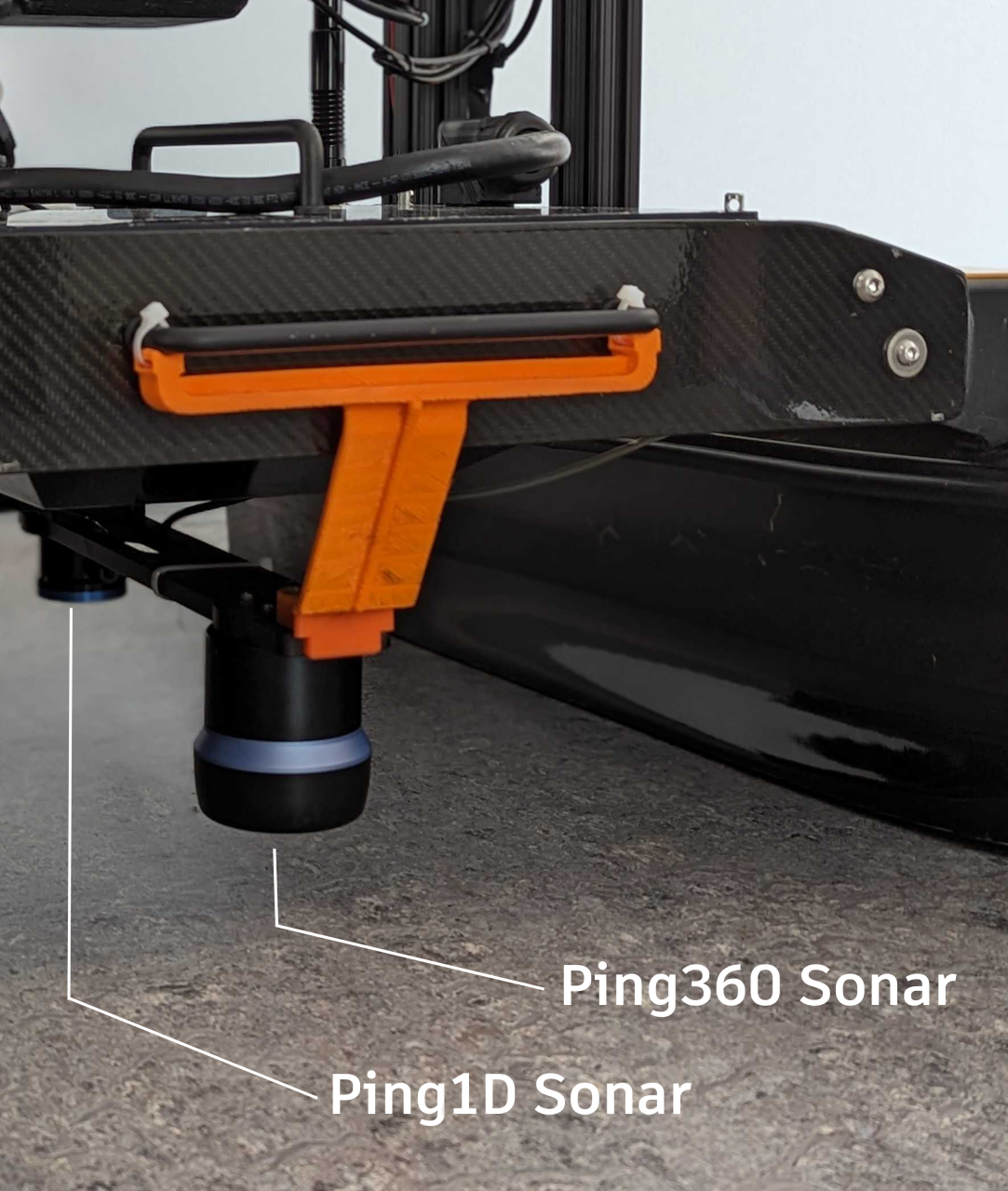}
        \caption{Bottom View}
        \label{fig:robot_botview}
    \end{subfigure}
    \hfill
    \begin{subfigure}[b]{0.6\textwidth}
        \centering
        \includegraphics[width=\linewidth]{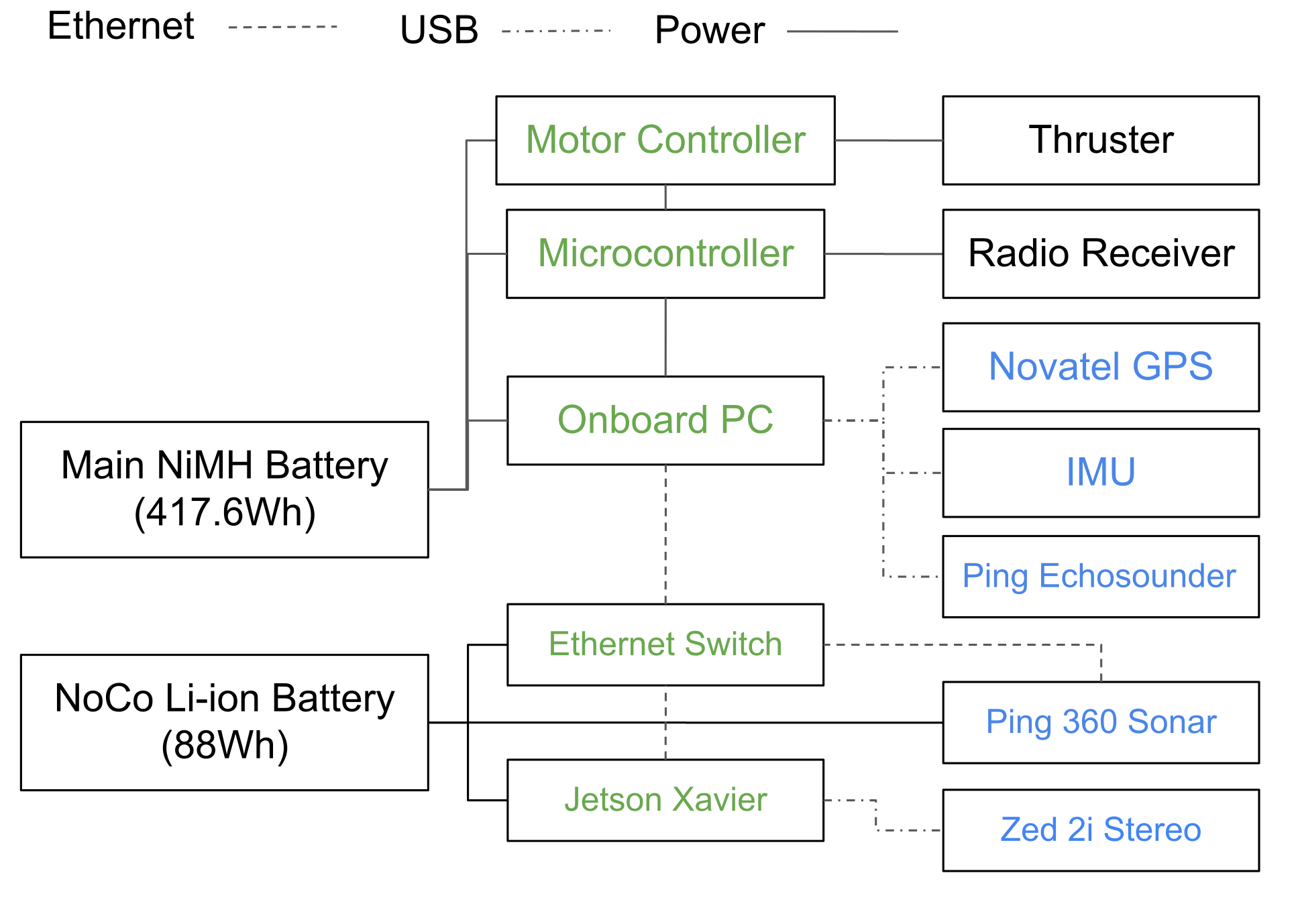}
        \caption{Electrical Diagram}
        \label{fig:electrical}
    \end{subfigure}
    \caption{Our \textit{Clearpath Heron} ASV for monitoring water quality during a field test. The ASV is equipped with various sensors including GPS, IMU, underwater scanning sonar, sonar altimeter, and a stereo camera. It also contains an Nvidia Jetson and an Atom PC for processing the data from these sensors. The locations where the sonars are mounted are depicted in (b), while the power and communication setups are illustrated in (c).}
    \label{fig:robot_combined}
\end{figure}

Our ASV platform, as depicted in Fig. \ref{fig:robot_combined}, consists of a modified \textit{Clearpath Heron} ASV equipped with a GPS, IMU, Zed2i stereo camera, Ping360 scanning sonar, and a Ping1d Sonar Echosounder Altimeter. 
The stereo camera is positioned in a forward-facing configuration and has a maximum depth range of 35 m. 
The Ping360 sonar is configured to perform a 20 m by 125\textdegree\ cone scan in front of the robot every 3.5 seconds, achieving a resolution of 1.8\textdegree. 
All computational tasks are handled by an Nvidia Jetson AGX Xavier and the onboard Intel Atom (E3950 @ 1.60GHz) PC on the Heron. 
The Jetson, stereo camera, and Ping360 sonar are powered by a lithium-ion jump-starter battery with an 88 WH capacity. To maintain the Jetson's power input at 19V, a voltage regulator is employed, allowing the Jetson to operate in its 30W mode. Additionally, a 417.6-Wh NiMH battery pack supplies power to the motors and other electronics. The batteries can support autonomous operations for approximately two hours. 
A schematic of the electrical system is presented in Fig. \ref{fig:electrical}. 
The additional payloads carried by the ASV have a combined mass of roughly 9 kg. 
Although water samplers have not been integrated into our system, they can be easily fitted in the future. 
The maximum speed of our ASV is approximately 1.2 m/s. 
Additionally, we have a remote controller available for manual mode operation, which can be utilized for safety purposes if needed.

\subsection{System Implementation Details} 
Our system's computational load is divided into offline and online processes (Figure \ref{fig:autonomy_stack}). 
Prior to the mission, we precompute the high-level graph and optimal policy, which are loaded onto the onboard PC. 
The online tasks are distributed between two onboard computers: the Atom PC and the Jetson. 
An Ethernet switch connects these computers, the sonar, and Heron's WiFi Radio. 
The GPS and IMU are connected to the Atom PC via USB, while the echo-sounder sonar and the stereo camera are connected to the Jetson via USB.
The switch allows remote SSH access and data transfer between the Atom and the Jetson. 
We use the ROS framework \autocite{quigley2009ros} to implement our autonomy modules in C++ and Python.
To synchronize time between the Jetson and the Atom PC, we employ Chrony for network time protocol setup. 
The Atom PC acts as the ROS master, responsible for vehicle interface, localization, updating the occupancy grids, running the local planner, and MPC controller. 
The Jetson handles resource-intensive tasks such as depth map processing, semantic segmentation, sonar obstacle detection, and data logging. 
Additionally, a ROS node hosting a web visualization page is served on the Atom PC. 
We also provide a Rviz visualizer to display the occupancy grid and outputs of the local planner and MPC. 
During the mission, the web server publishes the robot's locations and policy execution states in real-time on a web page served on the local network, using pre-downloaded satellite maps. 
The web visualization and Rviz can be accessed in the field from a laptop connected to Heron's WiFi. 
The policy executor publishes the global plan to the local planner and starts a timer when navigating a stochastic edge, but importantly does not incur any additional compute cost for planning.
We periodically save the status of policy execution online, enabling easy policy reloading in case of a battery change during testing.

We tune the update rates and resolutions of our sensing, perception, and planning modules based on the computational capacities of our Heron and Jetson systems. Specifically, we aim to maintain a sustainable compute load within the thermal and power limits. We avoid pushing our CPUs and GPUs to their absolute limits because doing so can lead to system unreliability and sudden frame rate drops in the field.

Therefore, we set the ZED stereo camera and the neural depth pipeline to publish at 5Hz with a resolution of 640 x 480. The semantic segmentation network, optimized using Nvidia's TensorRT framework, runs with a latency of less than 50ms. Sonar obstacle detection operates at 20Hz, synchronized with the arrival rate of new sonar scans. The occupancy grid map, with a resolution of 0.5m per cell, updates at 10Hz. The lateral BIT* local planner runs asynchronously in a separate thread, sampling 400 points initially and 150 points in each subsequent batch. The maximum dimensions of the sampling window are 40m in length and 30m in width. The MPC retrieves the planned path and calculates the desired velocity at 10Hz, using a step size of 0.1s and a 40-step lookahead horizon.

\subsection{Testing Site}
\begin{figure}[h!]
    \centering
    \begin{subfigure}[b]{0.98\textwidth}
        \centering
        \includegraphics[width=1\linewidth]{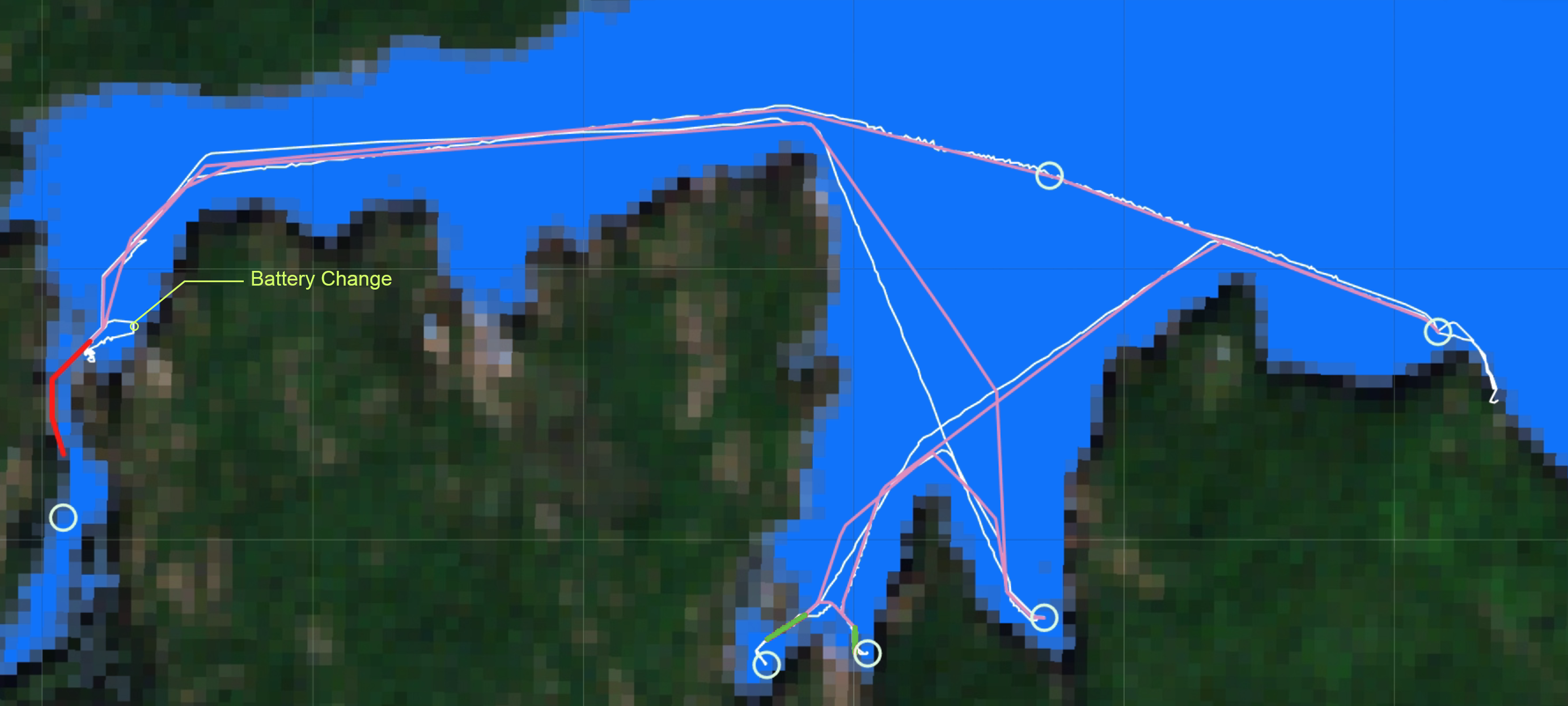}
        \caption{Lower Lake Mission}\label{fig:path_day12}
    \end{subfigure}
    
    \vspace{1em}
    
    \centering
    \begin{subfigure}[b]{0.53\textwidth}
        \centering
        \includegraphics[width=1.08\linewidth]{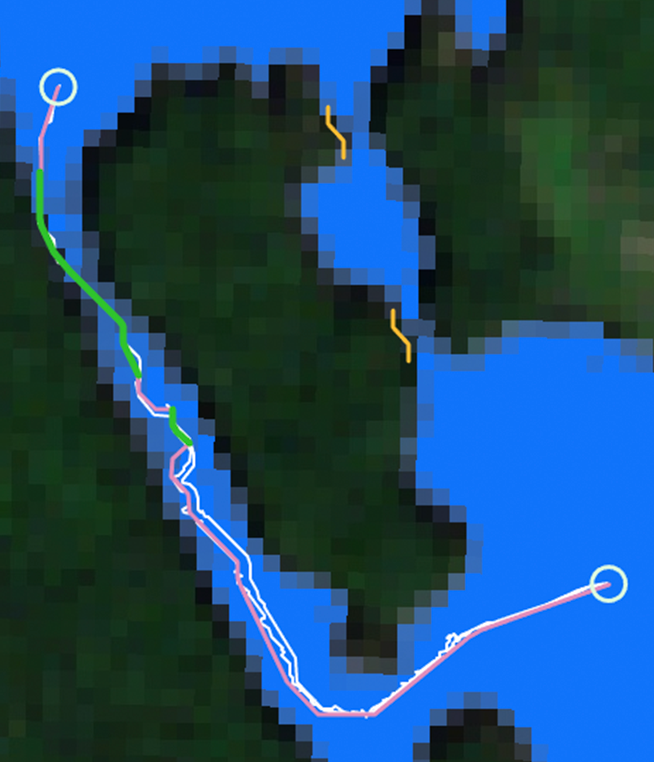}
        \caption{Upper Lake Mission (Short)}\label{fig:path_short1}
    \end{subfigure}
    \quad
    \begin{subfigure}[b]{0.442\textwidth}
        \centering
        \includegraphics[width=0.914\linewidth]{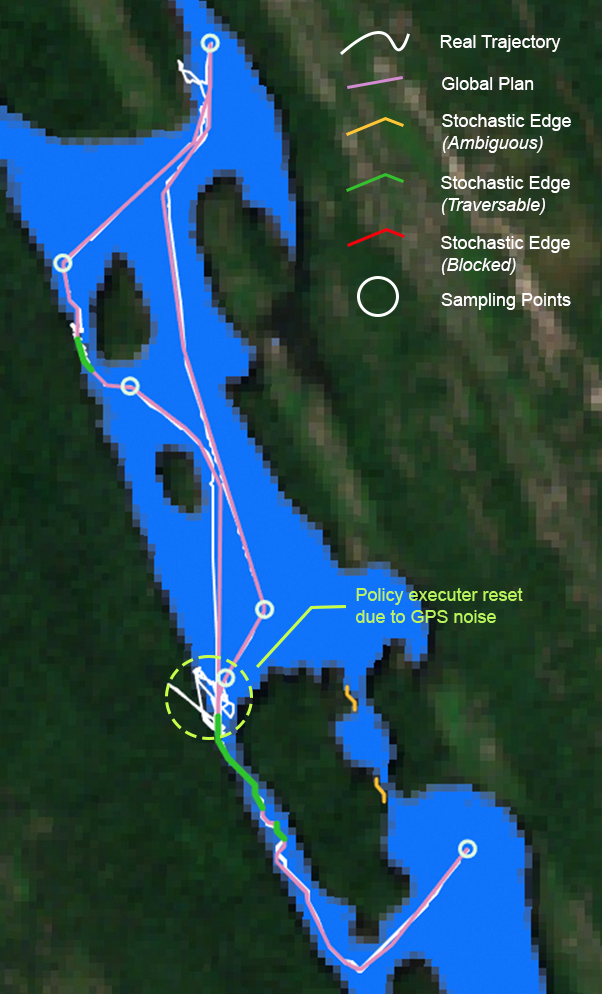}
        \caption{Upper Lake Mission (Long)}\label{fig:path_long1}
    \end{subfigure}
    
    \caption{Representative examples of global plans and trajectories traversed during field experiments. 
    All stochastic edges are labelled in colour.  \textcolor{ForestGreen}{Green} line means the stochastic edge is found traversable, \textcolor{red}{red} means untraversable, and \textcolor{darkyellow}{yellow} means the edge was not explored and remained ambiguous. A battery change in (a) and a manual intervention due to large GPS noises in (c) are also labelled. }
    \label{fig:path_overview}
    \vspace{-0.2cm}
\end{figure}

Our planning algorithm was evaluated at Nine Mile Lake in McDougall, Ontario, Canada. 
Detailed test sites and the three executed missions can be found in Fig. \ref{fig:path_overview}. 
The Lower Lake Mission in Fig. \ref{fig:path_day12} repeats the field test from our prior work \autocite{boat2022}, involving a 3.7 km mission with five sampling points, three of which are only reachable after navigating a stochastic edge. 
The stochastic edge at the bottom-left compels the ASV to manoeuvre through a thin opening amid substantial rocks not discernible in the \textit{Sentinel-2} satellite images. 
Besides repeating the old experiment from our prior work, we added two additional missions in the lake's upper areas.
Also, to assess our local mapping and planning stack's capabilities, an ablation mission was executed to see if the robot could safely navigate the stochastic edge at the bottom-left of the Lower Lake Mission. 
The policy in Upper Lake Mission (Short) was directly generated from the Fig. \ref{fig:example_graph} water mask. 
In fact, the high-level graph in Fig. \ref{fig:example_graph} and policy in Fig. \ref{fig:ao_tree} is a simplified toy version of our testing policy in the Upper Lake Mission (Short). The expected length of the policy is 1.0km long. 
We observed that our NovAtel GPS receiver’s reliability was impaired by large trees on the left stochastic edge in Fig. \ref{fig:path_short1}. 
On the right stochastic edges of the same subfigure, shallow regions, lily pads, and weeds were numerous.
Lastly, we extended this short mission to include three additional sampling sites and another stochastic edge at the lake's farthest point. 
The expected length of this Upper Lake Mission (Long) in Fig. \ref{fig:path_long1} is approximately 3.3 km
Despite having only 5 nodes, the Upper Lake Mission (Long) is still significant due to the complexity introduced by stochastic edges, resulting in 54 contingencies and a policy tree depth of 12. This demonstrates that even small-scale missions require our proposed approach to generate a robust global policy, and the local planner must effectively manage large uncertainties in traversability to execute the mission safely.

\subsection{Results of Mission Planner}
The aim of our field experiments is to test if our autonomy stack can successfully execute a global mission policy correctly and fully autonomously, without any manual interventions.
Results are summarized in Table \ref{tab:result} and we provide the overview and analysis of our results below.

\begin{table}[h]
\caption{Summary of the results of our tests for different policies, including any interventions due to algorithmic failure (excluding battery changes). The first two rows show the results of the Lower Lake Mission from our previous work.}
\begin{center}
\begin{tabular}{ccccc}
\hline
Mission & Sensors Used & Node Visited & \# of Interventions & Appeared In\\
\hline
Lower Lake Mission & Camera Only & 4/5 & 3 & \textcite{boat2022} \\
Lower Lake Mission & Camera Only & 3/5 & 3 & \textcite{boat2022} \\
\hline
Lower Lake Mission & Sonar + Camera & 4/5 & 1 & This Work \\
Lower Lake Mission & Camera Only & 4/5 & 0  & This Work \\
\hline
Upper Lake Mission (Short) & Sonar + Camera & 1/1 & 0  & This Work\\
Upper Lake Mission (Short) & Sonar + Camera & 1/1 & 0 & This Work\\
\makecell{Upper Lake Mission (Short) \\ (Left Edge Blocked)} & Sonar + Camera & 1/1 & 0 & This Work\\
\makecell{Upper Lake Mission (Short) \\ (Left Edge Blocked)} & Sonar + Camera & 0/1 & 0 & This Work\\
\hline
Upper Lake Mission (Long) & Sonar + Camera & 5/5 & 1 & This Work \\
\hline
\end{tabular}
\end{center}
\label{tab:result}

\end{table}

\textbf{Lower Lake Mission} We undertook the lower lake mission twice - first, using both sonar and camera and second, using only the camera. 
The ASV successfully reached 4/5 targeted locations during both trials, with the exception of the bottom-left location. 
This was due to the ASV's inability to autonomously navigate through large rocks within the designated time frame. 
When contrasted with prior experiments noted in \textcite{boat2022}, our trials showed marked improvement, with only a single manual intervention required due to algorithmic failure during the first run, and none during the second. 
The intervention was necessitated by the ASV's collision with a tree trunk (the one in Fig. \ref{fig:extruding_logs}) it failed to identify, resulting in manual manoeuvring to remove the obstruction. 
In both trials, the policy executor deemed the bottom-left stochastic edge untraversable because the local planner did not find a path through large rocks within the time limit.
The ASV was then safely directed back to the last sampling location and starting location.
Moreover, these trials demonstrated a significant improvement in the stability of our navigational autonomy compared to the same field test conducted last year. 
These can be attributed to several factors. 
First, the inclusion of a new semantic segmentation network for the stereo camera allowed the ASV to navigate confidently even in conditions of high sunlight glare or calm water. This was in contrast to the geometric approach in our previous work, which resulted in both numerous false positives and missing obstacles. 
Second, sonar detection capabilities facilitated the identification and avoidance of underwater rocks by the local planner. We were also able to fuse both sonar and stereo camera inputs with a local occupancy grid map. 
Third, through the incorporation of a Model Predictive Control (MPC) tracking controller, the reliance on GPS velocity estimates was removed. 
Lastly, the decision to use an 88Wh battery on the ASV markedly improved the Jetson's battery life, thereby negating the need for battery changes during each mission.
In Table \ref{tab:power}, we show that the Jetson and onboard PC are very power-hungry during one of the testing trials. 
A microcontroller inside our ASV measures the power of the onboard PC, and we use the jetson-stats tool to log the power of the Jetson.
Although the measurement is anecdotal and the exact power consumption can depend on other factors, such as the state of the battery and operating temperatures, the 88WH battery powering the Jetson can certainly last through a two-hour-long experiment.

\begin{table}[t!]
\caption{Average usage and power consumption of our computing devices during a Lower Lake Mission. }
\centering
\begin{tabular}{c|c|cc}
\hline
Device & Heron CPU & Jetson CPU & Jetson GPU \\ 
\hline
Usage(\%) & 75.2 & 61.6 & 89.3 \\
Power(W) & 9.2 & \multicolumn{2}{c}{19.3 (Combined)} \\
\hline
\end{tabular}
\label{tab:power}
\end{table}

\textbf{Upper Lake Mission (Short)}
We performed four successful tests of this new policy on the upper lake to determine if our robot could execute different policy branches and navigate both sides of the central island, which were visibly passable based on aerial observations. The expected length is 1.0km. 
The success criteria were defined as either safely traversing the stochastic edges on either side within the assigned time limit or safely returning to the starting point without collisions.
Initially, we executed the policy twice without modifications.
As depicted in Fig. \ref{fig:path_short1}, the policy guided the ASV to navigate and return along the left stochastic edge, which had a lower expected cost than the right edge. 
For the subsequent two trials, we deliberately triggered an early timeout to block the left edge in the policy, forcing the ASV to navigate the right edge.

Throughout the four trials, the ASV executed the mission-level policy fully autonomously, except for a battery change. 
Navigating the left side was straightforward despite occasional GPS signal disruptions. 
On the right side, the ASV successfully reached the target area once. 
However, in the second attempt, it travelled too slowly in a shallow area with many aquatic plants (see Fig. \ref{fig:shallow_water_and_aquatic_plants}) and eventually reached the time limit, rendering the right stochastic edge untraversable. 
Despite this, the ASV safely returned to the starting node.
Importantly, we considered a trial a success despite the ASV not reaching the designated target if the overall policy was executed autonomously.
No collisions occurred during any trial.

\textbf{Upper Lake Mission (Long)}
We expanded the previous policy to a more extensive mission, covering a larger area of Nine Mile Lake's upper parts with the same starting point as the shorter mission. The expected length is significantly longer at 3.3km. 
First, the boat navigated the stochastic edges on the island's left side to reach the sample point, and it returned using the same path. 
Despite this, significantly deteriorated GPS signals were observed at the edge's end, preventing the mission-level policy executor from detecting the completion of the edge traversal due to GPS solution noise. 
Consequently, a manual restart of the policy executor was necessary.
Thereafter, the ASV proceeded upward to the next sample point before making a left turn to go through a shortcut pinch point, visiting two more sample points. 
Following a brief stop for battery replacement, the ASV completed the remaining mission.

In evaluation, our local perception pipeline performed commendably in this area despite having never previously collected data here. 
In particular, the synergy of sonar obstacle detection and the stereo camera's semantic waterline estimation showed high reliability in close-range shoreline and obstacle detection with very minimal false positives. Along with the previous four trials for the shorter mission, we demonstrate that our autonomous navigation architecture is effective not only in familiar environments but also in previously unseen conditions.

\begin{figure}[t!]
    \centering
    \begin{subfigure}{0.465\textwidth}
        \includegraphics[width=\textwidth]{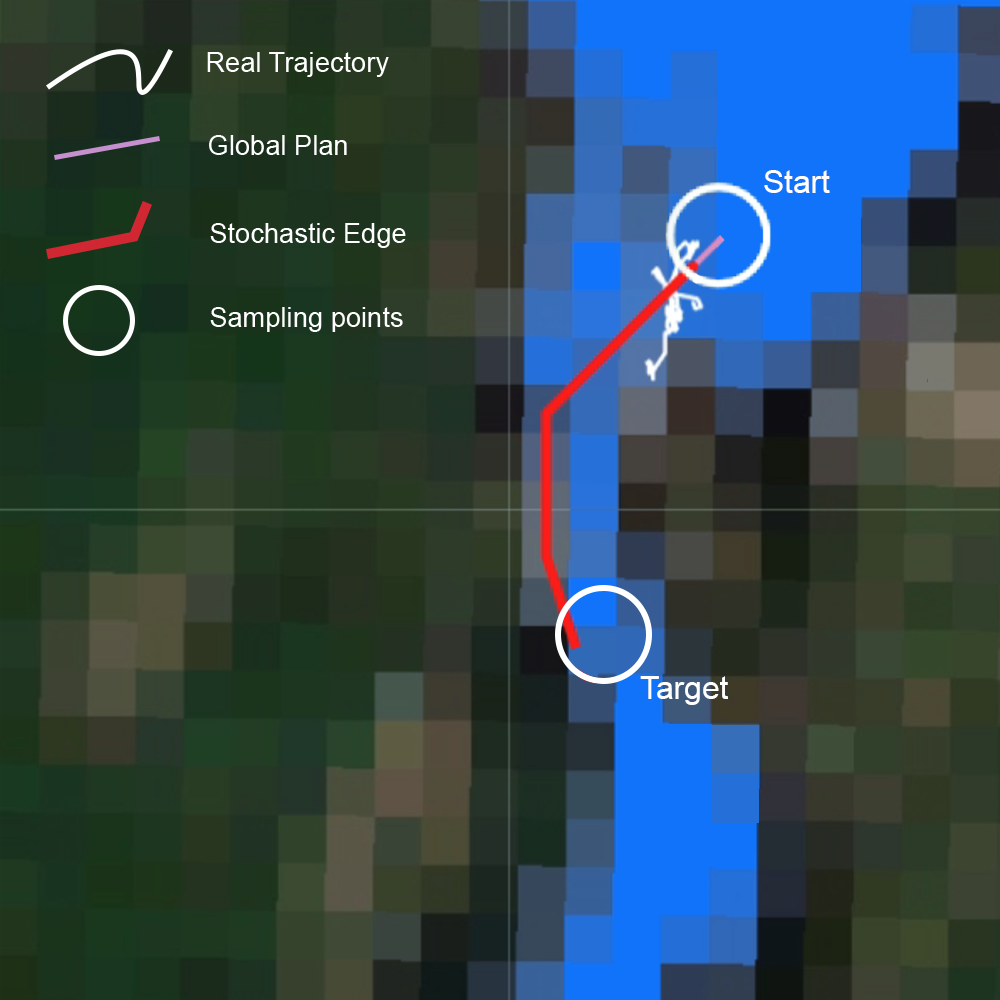}
        \caption{Autonomous attempt (forward)}
    \end{subfigure}
    \hfill
    \begin{subfigure}{0.465\textwidth}
        \includegraphics[width=\textwidth]{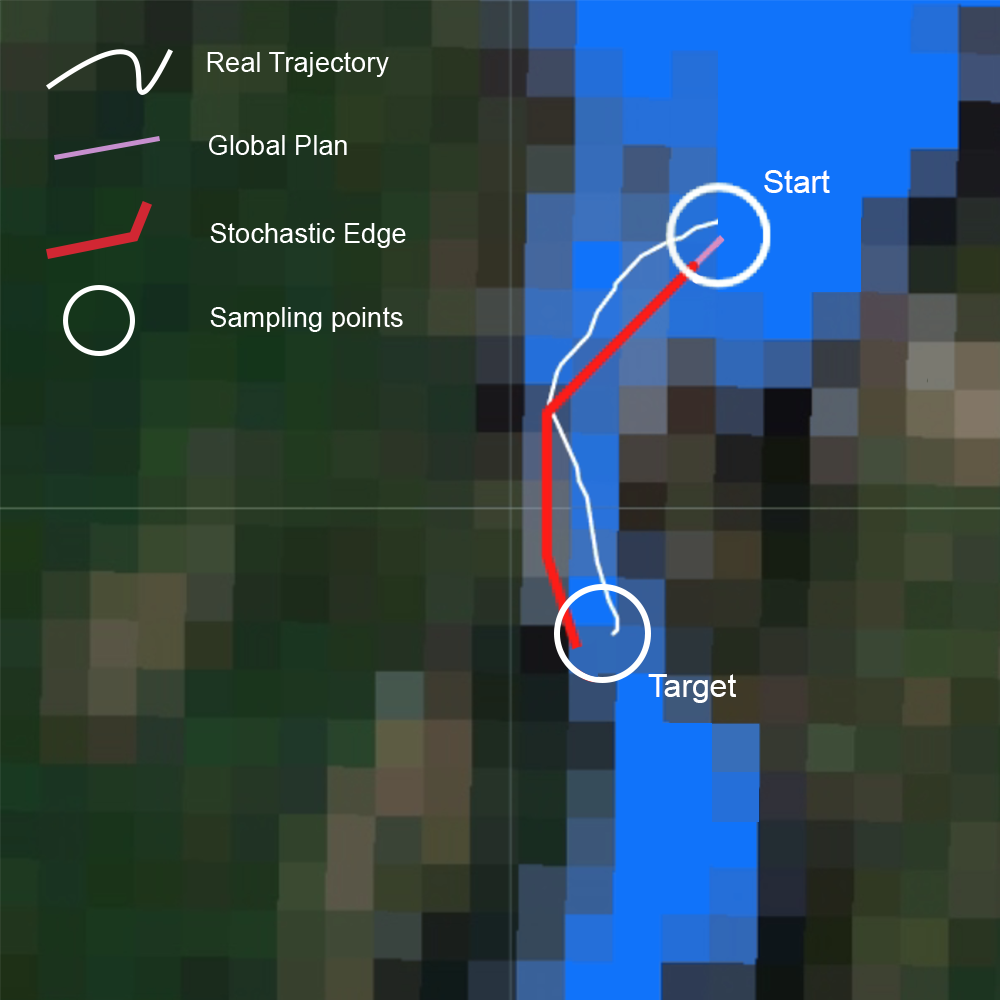}
        \caption{Manual attempt (forward)}
    \end{subfigure}
        \begin{subfigure}{0.465\textwidth}
        \includegraphics[width=\textwidth]{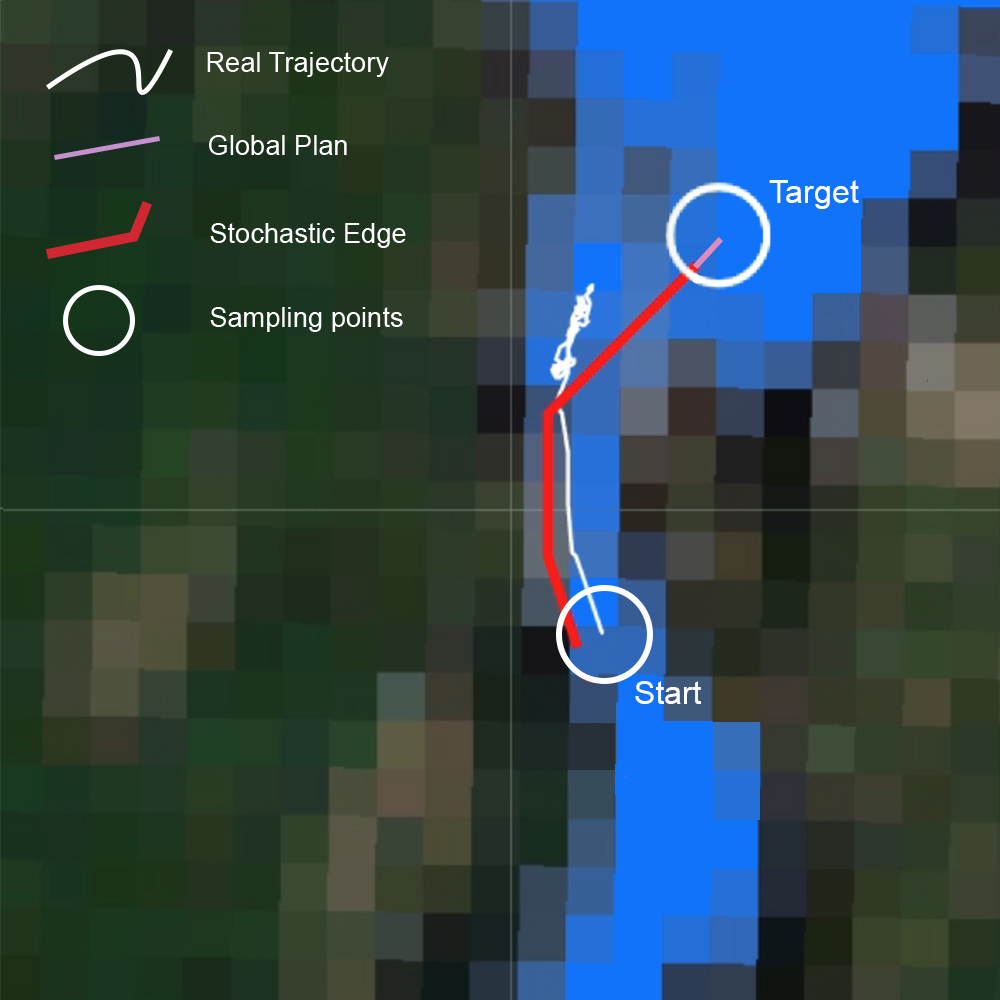}
        \caption{Autonomous attempt (back)}
    \end{subfigure}
    \hfill
    \begin{subfigure}{0.465\textwidth}
        \includegraphics[width=\textwidth]{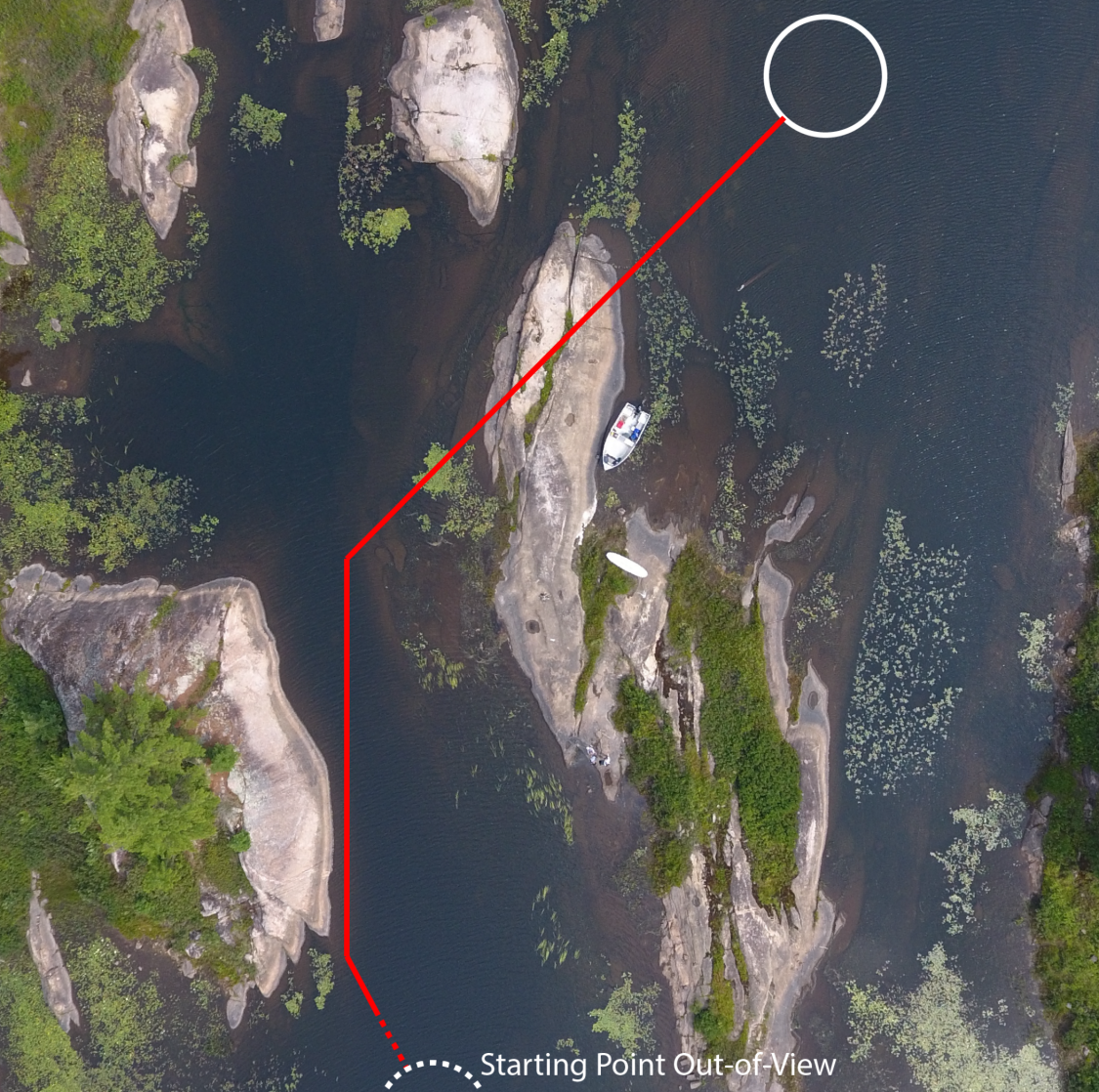}
        \caption{Aerial view (hand-sketched path from global plan)}
    \end{subfigure}
    \caption{Comparison of the global plan, manual traversal, and autonomous navigation through the stochastic edge. The global plan, calculated from coarse satellite images, is blocked by a rock. In (b), the ASV was able to pass the narrow opening under manual teleoperation. However, the ASV was unable to identify the opening in the local occupancy grid in autonomous mode (see. Fig. \ref{fig:occ_map_rock}), so it searched for an opening in place until the time limit and returned to the start. }
    \label{fig:trajectory_stochastic_edge}
    \vspace{-0.2cm}

\end{figure}

\begin{figure}[t!]
    \centering
    \begin{subfigure}{0.45\textwidth}
        \includegraphics[width=\textwidth]{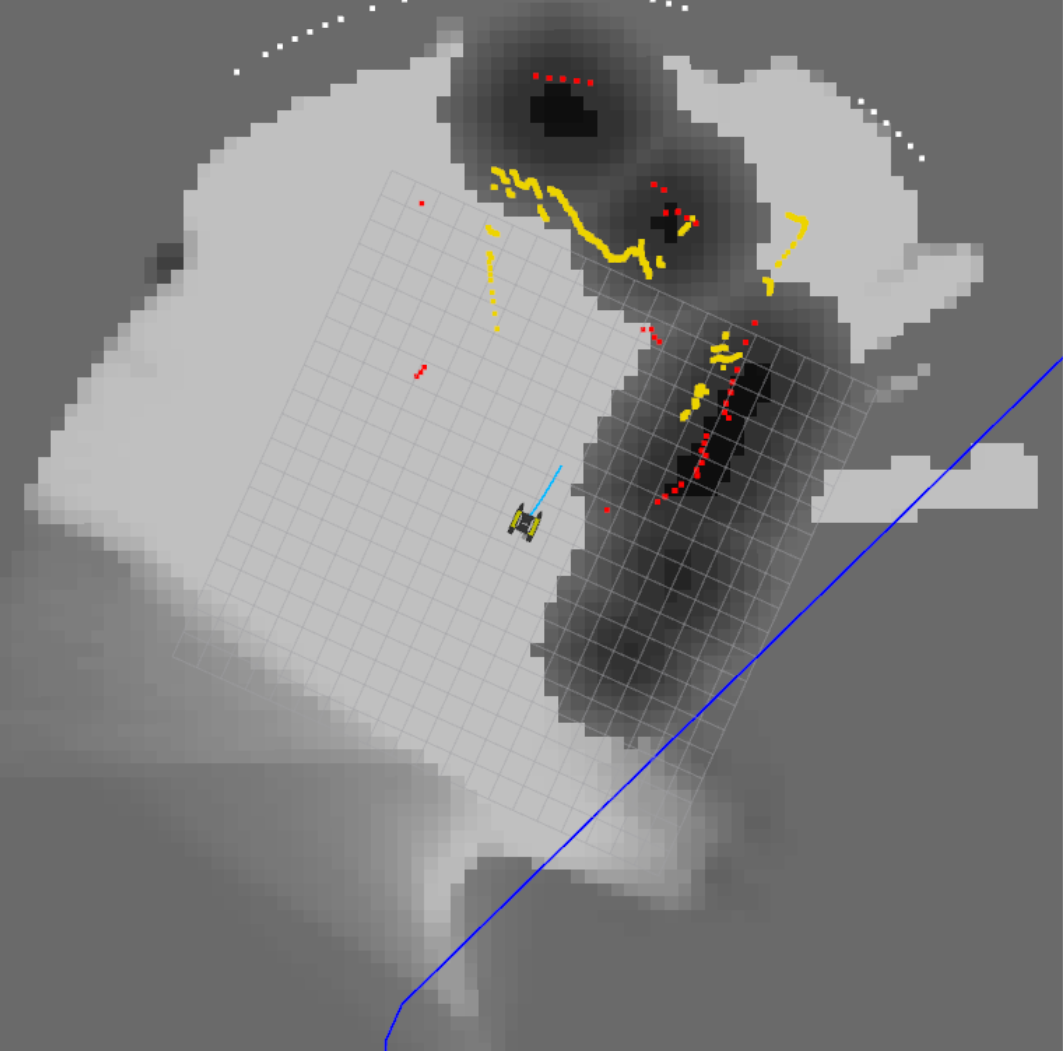}
        \caption{Occupancy Grid Map (ASV Start from South)}
    \end{subfigure}
    \hfill
    \begin{subfigure}{0.48\textwidth}
        \includegraphics[width=\textwidth]{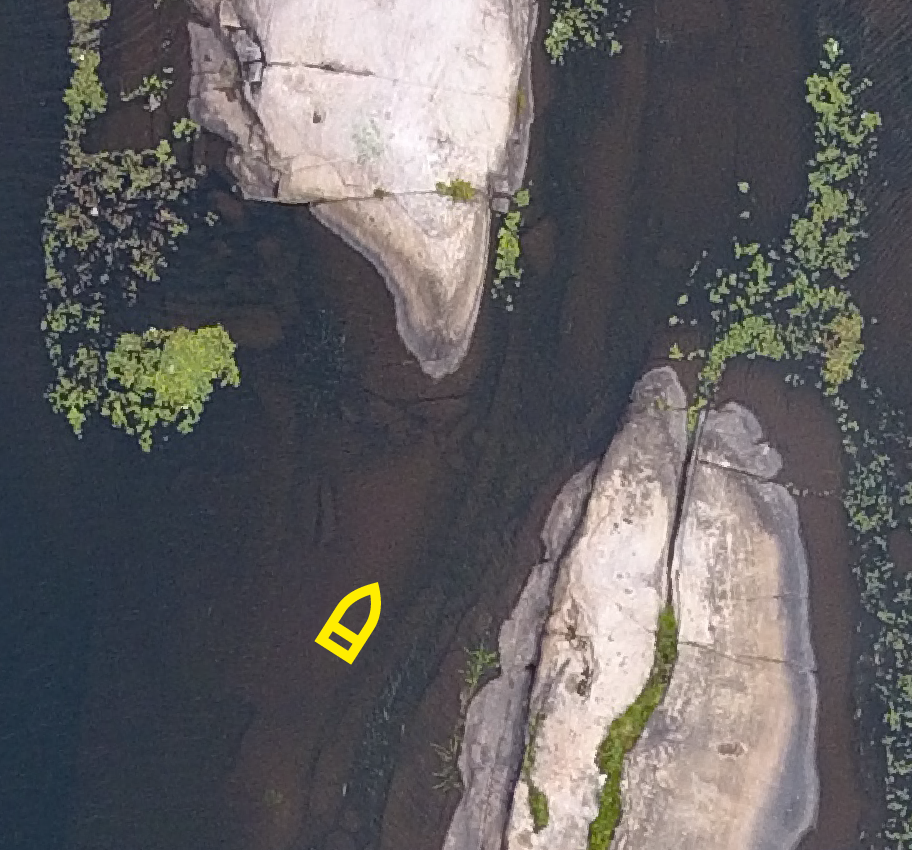}
        \caption{Aerial View of the Scene in (a)}
    \end{subfigure}
    \begin{subfigure}{0.45\textwidth}
        \includegraphics[width=\textwidth]{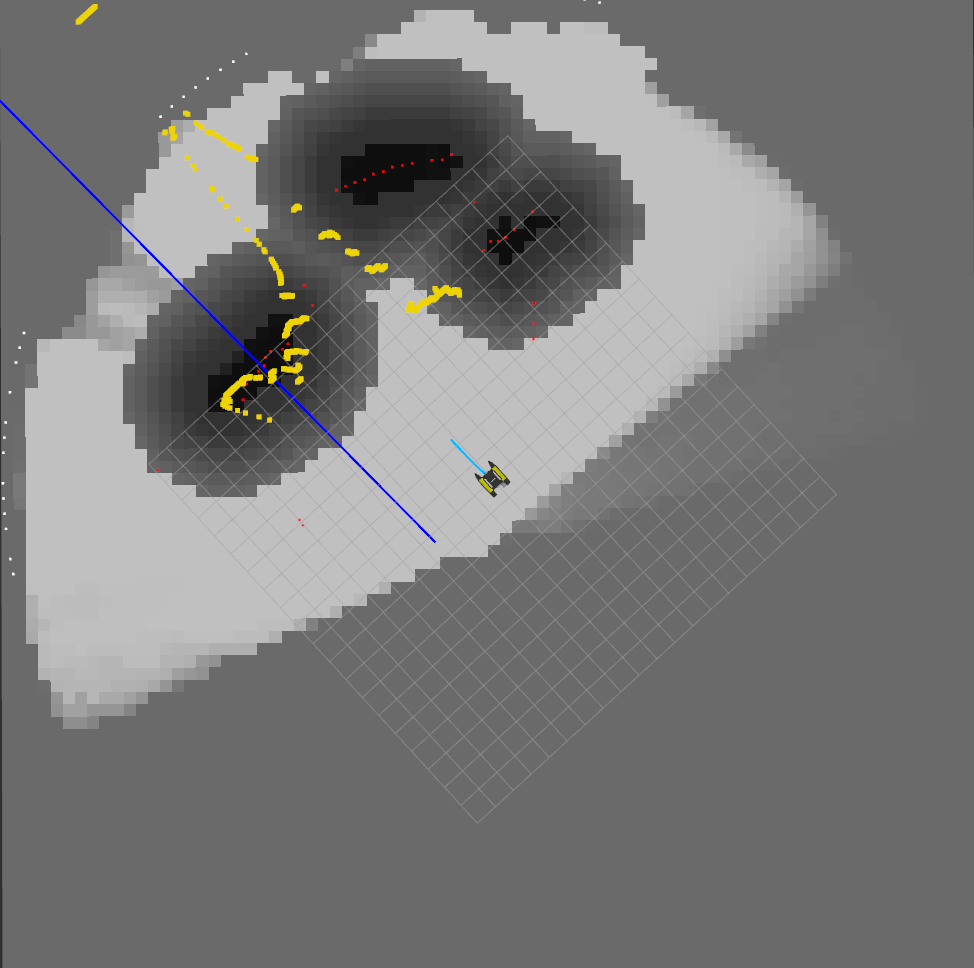}
        \caption{Occupancy Grid Map (ASV Start from North)}
    \end{subfigure}
    \hfill
    \begin{subfigure}{0.48\textwidth}
        \includegraphics[width=\textwidth]{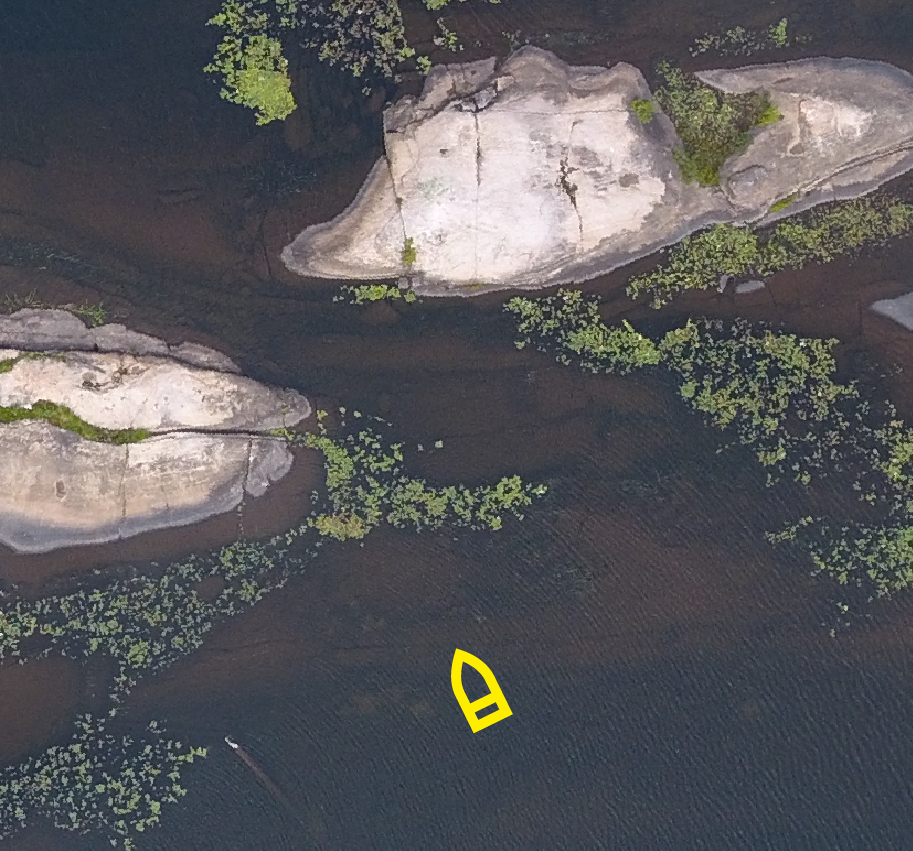}
        \caption{Aerial View of the Scene in (c)}
    \end{subfigure}
    \caption{Comparison between the robot's occupancy grid maps and aerial image. \textcolor{darkyellow}{Yellow} dots are waterline estimated in 3D. \textcolor{Red}{Red} dots are obstacles detected by sonar. Boat symbols are added to (b) and (d) for context. The global plan (\textcolor{blue}{blue} line) is blocked by rocks, so the ASV needs to detour through the narrow opening. However, the passage is blocked on the occupancy grid due to our inaccurate detection, localization, and excessive dilation.}
    \label{fig:occ_map_rock}
    \vspace{-0.2cm}
\end{figure}

\subsection{Isolated Testing of Local Planner}
A main contribution of our current work is the new perception and local planner modules that can safely disambiguate stochastic edges and navigate safely and autonomously in obstacle and terrain-rich waterways without high-resolution prior maps. 
To verify this, we tested the local planner on a stochastic edge ten times with the exact same parameter, five times each in either direction.
Success was demonstrated by either reaching the stochastic edge's other endpoint within a set time frame or returning to the starting point upon timeout of the policy executor. 
Without intervention, the ASV accomplished this 70\% of the time. However, in three instances, it collided with or became trapped by obstacles, such as rocks and a tree trunk. 

The global path extracted from the \textit{Sentinel-2} Image was interrupted by a large rock, with only two narrow openings between the rocks, manually traversable, as demonstrated in Fig. \ref{fig:trajectory_stochastic_edge} (b). One of the narrow openings is visible from the aerial view in Fig. \ref{fig:trajectory_stochastic_edge} (d). 
Our ASV can detect these rocks; however, the over-aggressive dilation parameter obstructs the local planner from charting a path through the central passageway (see Fig. \ref{fig:occ_map_rock}). 
There is another wider opening on top of the visible narrow opening, but it is over 30m away from the nearest point on the global path and thus exceeds the maximum corridor width of our local planner's curvilinear space.

Relying exclusively on GPS/IMU for location and a local occupancy grid centred around the ASV poses considerable challenges in this terrain, due to imprecision in localizing obstacles relative to the robot and issues controlling tight turns and precise path tracking, escalating the collision risk in confined spaces. 
In order to mitigate noise and path plan conservatively, occupancy values were decayed over time, and substantial dilation was applied around occupied cells. 
As such, the ASV would not construct and finetune a consistent local map, but would instead overlook previously encountered obstacles. 
Consequently, the local planner oscillates between two temporarily obstacle-free paths in the occupancy grid, while the ASV stops and unsuccessfully searches for a traversable path locally until the timer limit is reached, as shown in Fig. \ref{fig:trajectory_stochastic_edge} (a) and Fig. \ref{fig:trajectory_stochastic_edge} (c).

Another key reason for the low quality of the occupancy grid is the difficulty of fusing sonar and stereo camera measurements, especially at longer ranges. 
Since sensor fusion occurs solely within the occupancy map, both sensors need to detect an obstacle simultaneously at the same location in the map for accurate fusion.
This can be challenging due to a variety of reasons. 
For instance, depth measurements produced by the stereo camera tend to be noisier over a larger range. 
Our camera is not capable of detecting underwater obstacles detected by sonar. 
Additionally, our system lacks effective uncertainty measures for updating sonar and stereo observations within the occupancy map, especially when the two sources provide conflicting data.
For example, the ASV simply did not detect the tree trunk.
Thus, our sensor fusion mechanism proves effective only over shorter ranges where the sonar and camera are more likely to align. If it is possible to extend the range of our perception modules, the ASV could formulate more optimized navigation paths, preventing collisions with obstacles such as rocks.

\section{Lessons Learned}
\vspace{-0.4cm}

In this section, we outline insights garnered from our field tests, emphasizing successful design aspects related to field-tested ASV navigation systems and suggesting potential improvements for future iterations. 

\textbf{Timer} Primarily, we found that using a timer to disambiguate stochastic edges was simple, robust, and practical.
Integration of a timer within our ROS-based system was easy and could accommodate unexpected hindrances such as strong winds, making stochastic edges difficult to traverse. 
This allowed for uninterrupted policy execution even when the local planner failed to identify viable paths through a traversable stochastic edge. 
Essentially, the inclusion of a timer fostered independence between the execution of our mission-level policy and the selection of local planners, enabling the ASV to conduct water sampling missions irrespective of local planner errors.

\textbf{Localization} A critical limitation of our system lies in the absence of precise GPS localization. 
Our system necessitates a seamless integration of local mapping with broader satellite maps to facilitate accurate navigation in complex scenarios, such as those illustrated in Fig. \ref{fig:trajectory_stochastic_edge}. 
A GPS alternative, such as SLAM, would introduce redundancy, bolstering navigation robustness when GPS signals become compromised due to obstructions, interferences, or adverse weather conditions. 
Furthermore, minimizing localization noise could enhance speed and steering control, enabling the ASV to operate more swiftly and smoothly.

\textbf{Occupancy Grid} As demonstrated in the previous section, our occupancy grid map also struggles with sensor fusion - particularly over long ranges where sonar and stereo camera measurements can contradict. 
These inconsistencies necessitated the introduction of a time-decay factor and significant dilation around obstacles. As a result, we observed a `drunken sailor' phenomenon, wherein the ASV constantly navigates within a confined space without any real progress. 
We think that semantic SLAM integration with the stereo camera could ameliorate local occupancy map issues.
If SLAM can provide a locally consistent and metrically accurate map of higher quality, the decay factor in the occupancy grid becomes unnecessary and the planner will not oscillate. 
While SLAM is impractical in open water due to the absence of stationary features near the robot, it becomes viable in densely obstacle-populated scenes such as pinch points or shorelines. 
Localizing the robot against semantic-based local features could lead to more accurate localization and, furthermore, improve obstacle-relative pose estimation and traversability assessment.
As we can store and grow the map as the robot explores unknown areas, the planner can also work with a static occupancy grid and avoid any oscillation.
Furthermore, we also recommend better exploration strategies to build local maps and search for traversable paths rather than fixing the planning domain size around the precomputed global path from inaccurate satellite images. 

As the map can be expanded when the robot explores unknown areas, the planner can work with a fixed occupancy grid to avoid oscillation. Additionally, more effective local map building and traversable path searching strategies might provide better solutions than confining the planning domain size around inaccurate satellite images' precomputed global path.

\textbf{Evasive Manoeuvres} Our system currently lacks evasive manoeuvres. Despite collisions with obstacles, the robot could feasibly retreat and navigate back to unobstructed waters. However, our local planner often fails to detect forward obstacles, continuing to chart a forward path after collisions. Both the stereo camera and sonar have minimum range limitations, resulting in undetected proximate obstacles. 
We could introduce the timer mechanism to prompt evasive manoeuvres. For instance, if the ASV remains stationary despite forward movement instructions from the planner and controller, it should back up and reset its local planner to circumnavigate the same area. While the ASV may struggle to self-extricate from a beach or shallow rock without human assistance, evasive manoeuvres could facilitate the avoidance of obstacles such as tree trunks or aquatic plants.

\textbf{Sonar} The incorporation of sonar in our system entails both advantages and drawbacks. Positively, it enabled the detection and circumvention of underwater obstacles, beyond the stereo camera's capabilities. Conversely, the sonar's slow scanning rate (3 seconds per scan) restricts it from being the solitary onboard perception sensor. Additionally, our heuristic-based obstacle detection method fails to recognize minor obstacles, such as lilypads or weeds. While the sonar effectively gauges obstacle distances from the ASV, it cannot determine the depth of underwater obstacles since it scans horizontally. This depth ambiguity complicates traversability estimation, which relies on exact water and underwater obstacle depth knowledge. Moreover, merging sonar with the stereo camera proves challenging due to their observing different world sections.

\textbf{System Integration} 
While designing autonomy algorithms with general marine navigation in mind, we recognize that the integration process was tailored to our particular ASV platform and test scenarios. The primary objective of system tuning is to optimize performance metrics such as speed, accuracy, tracking error, and reliability within the bounds of certain constraints, including latency, computational usage, and sensor capabilities. For each autonomy module, we identify key parameters that significantly impact performance. For instance, in the occupancy grid map, grid resolution, smoothing and dilation models, and measurement weights are crucial. Obstacle detection with sonar is governed by the peak threshold and the size of the smoothing window. The runtime of the Lateral BIT* planner is affected by the batch size and sampling window size. The controller's tracking performance depends on the controller cost terms and lookahead horizon. Notably, the accuracy of stereo waterline estimation is not very sensitive to the smoothing parameters but is primarily dependent on the quality and volume of the training data.

Initially, we manually tuned these parameters using previously collected datasets or in simulation to establish a baseline. Subsequently, we deploy all autonomy algorithms on the real robot, testing each component individually again. Here, we utilize ROS software tools such as Rviz and the dynamic reconfigure package to evaluate each module's performance and adjust. Often, it is necessary to reduce the update rate and resolution of the algorithms to prevent performance degradation due to latency or computational constraints. We continuously record and assess our ASV’s performance under varying conditions, refining them as needed before conducting field experiments with a fixed set of parameters.

Our autonomy algorithms demonstrated commendable field performance, but many potential improvements from a system engineering standpoint still remain. An immediate goal is enhancing our software's efficiency to decrease computational load and power consumption on both the Atom PC and the Jetson. For instance, running semantic SLAM alongside the existing stack would require additional power and considerable software optimization to avoid straining our computers further. 
Aside from optimizing power use, improvements to efficiency, reliability, and usability could be advantageous, particularly for nontechnical users. Our Rviz and web interface user displays contain critical monitoring and debugging information but demand extensive navigation system familiarity. 
Our data logging pipeline consumes substantial storage space (about 1GB/min), imposing both storage and time cost burdens for copying and analysis. Booting up the GPS in the field was another challenge due to prolonged wait times for adequate satellite acquisition for autonomous navigation. 
In terms of future hardware, vegetation-proof boat hulls and propellers should be considered given the increased drag and potential damage to the propeller blades from aquatic plant interference. Furthermore, electronic connectors capable of withstanding transportation-induced vibrations and cables that shield connections from interference would enhance overall system robustness.

\textbf{PCCTP Formulation} Currently, we have found PCCTP to be a robust framework for enabling longer-term autonomous environmental monitoring tasks. Our policy is designed to be resilient against environmental uncertainties, ensuring that the robot can complete its mission both safely and efficiently. In the Nine Mile Lake experiments, the ASV detected many obstacles that were missing or not clearly mapped in the satellite imagery. This demonstrates the importance of using a global mission policy when deploying autonomous robots in unfamiliar and remote marine environments, where unforeseen obstacles absent in the satellite images can halt the execution of a single task plan. By characterizing only pinch points and windy edges as stochastic edges, the problem becomes tractable to solve optimally, effectively capturing the uncertainties visible across different satellite images. In field tests, we found that edges assumed to be ``traversable" were indeed always traversable. Furthermore, we could manually inspect and verify all possible global paths in the policy before field deployment since we found the optimal policy offline. This approach made it easy to understand the ASV's high-level objective during our field test, particularly when radio communication with the robot became unreliable. 

However, we have identified several potential enhancements to our problem formulation after concluding our field tests. First, the high-level policy could still result in a deadlock if a ``traversable edge" becomes untraversable due to unexpected factors. This did not occur in our testing, but one way to mitigate this is to add a small blocking probability to these edges, However, the scalability of the algorithm may need to be improved to efficiently plan for additional stochastic edges. Second, the blocking probabilities of stochastic edges may be correlated in a real environment. For example, wind and water levels can simultaneously affect the traversability of all stochastic edges, and the same aquatic plants may proliferate across nearby stochastic edges. These factors could be modelled by using a joint distribution with covariance for the probabilities of all stochastic edges, or by using a Bayesian model with latent variables to represent common environmental factors. Third, power is often a significant constraint for fully autonomous execution, requiring the robot to return to the base for charging periodically. In PCCTP, we can address this by imposing a distance limit on the planner, ensuring that the robot must return to the start or designated charging locations. Lastly, PCCTP is a one-shot planning algorithm and does not replan online for new robot tasks. An interesting extension for PCCTP would be for lifelong water monitoring missions, where target locations can be added online, and the robot can replan its policy with its next targets as a new starting location. In this setting, the traversability estimates may also be updated online during the lifelong execution.

In addition to these changes to the formulation, we envision ways to expand the PCCTP to incorporate scientific heuristics. A straightforward extension could involve using multispectral satellite bands, such as MODIS \autocite{Wu2009-nx}, to analyze water quality in the target area and automatically select target locations. If the robot is equipped with an online-capable water quality sensor like the YSI sonde, further opportunities arise. For instance, with a predefined set of target locations or scanning patterns, the ASV could optimize its policy to maximize both information gain and expected cost under uncertain traversability conditions. Our problem formulation would then need to be extended to a multi-objective framework, employing either a weighted sum of objectives or Pareto-based approaches \autocite{Chen2021-vl,Salzman2023-dr}. If the ASV is tasked with repeatedly patrolling the same area, an online approach might be more suitable, allowing the robot to continually update its model of traversability and the scientific value of the target area. However, efficiently solving these optimization problems remains a challenge.

\textbf{Field Logistics} Our field logistics proved successful largely due to employing a motorboat, facilitating rapid transportation of the robot, personnel, and supplies to remote testing locations on the lake. During trials, staying in close proximity to the robot or flying a drone for tracking was straightforward using a motorboat. In case of forgotten crucial equipment, swift return trips to the base camp for recovery were possible. Our field tests, spanning three days, were completed as planned, despite limited time and battery life.

\vspace{-0.3cm}
\section{Conclusions}
For a robot to be effective in real-world environments, it must adapt to variations and uncertainties stemming from natural or human factors, despite potential mismatches between the real world and the planning model. Therefore, a robust mission planning framework for long-term autonomy should possess several key qualities: resilience to allow continuous operation without failure, adaptability to incorporate uncertainties specific to the task and environment, and efficiency in meeting critical performance metrics such as time, throughput, and energy cost.

With these criteria in mind, we have proposed a framework for planning mission-level autonomous navigation policies offline using satellite images. Our mission planner treats the uncertainty in these images as stochastic edges and formulates a solution to the Partial Covering Canadian Traveller Problem (PCCTP) on a high-level graph. We introduce PCCTP-AO*, an optimal, informed-search-based method capable of finding a policy with the minimum expected cost. Tested on thousands of simulated graphs derived from real Canadian lakes, our approach demonstrates significant reductions in travel distance—ranging from 1\% (50m) to 15\% (1.8km).

We then developed a GPS-, vision-, and sonar-enabled ASV navigation system to execute these preplanned policies. We proposed a conceptually simple yet robust timer-based approach to disambiguate stochastic edges. Local mapping modules integrate a neurally estimated waterline from the stereo camera with underwater obstacles detected by sonar, while the local motion planner ensures obstacle avoidance in adherence to the precomputed global path. Our ASV navigation system has successfully executed three different km-scale missions a total of seven times in environments with unmapped obstacles, requiring only two interventions in total. Additionally, we achieved a 70\% success rate in an isolated test of our local planner.

Our findings highlight that while the system performs robustly, traversability assessment and localization continue to be bottlenecks for local mapping and motion planning. We hope that the lessons learned from this development process will foster future advances in long-term autonomy algorithms and ASV environmental monitoring systems.

\subsubsection*{Acknowledgments}
We would like to acknowledge the Natural Sciences and Engineering Research Council of Canada (NSERC) for supporting this research.

\printbibliography 

\end{document}